\documentclass{article}
\usepackage{graphicx}
\usepackage{subcaption}
\usepackage{amsmath}
\usepackage{amssymb}
\usepackage{geometry}
\usepackage{hyperref}
\usepackage{xcolor}
\usepackage[mathlines]{lineno}

\usepackage{titlesec}
\date{}

\usepackage[sort&compress, numbers, round]{natbib}
\usepackage{url}

\title{Deep learning for predicting the occurrence of tipping points}

\newcommand{\affilmark}[1]{\textsuperscript{#1}}

\author{
	Chengzuo Zhuge\affilmark{1,2} \and
	Jiawei Li\affilmark{2,3} \and
	Wei Chen\affilmark{2,3,4}*
}

\date{
	\begin{flushleft}
		\small
		\affilmark{1}School of Mathematical Sciences, Beihang University, Beijing, 100191, China\\
		\affilmark{2}Key Laboratory of Mathematics, Informatics and Behavioral Semantics (LMIB), Beihang University, Beijing, 100191, China\\
		\affilmark{3}School of Artificial Intelligence, Beihang University, 100191, Beijing, China\\
		\affilmark{4}Zhongguancun Laboratory, 100194, Beijing, China\\[2ex]
		* chwei@buaa.edu.cn
	\end{flushleft}
}

\begin{document}
	
	\maketitle
	
    \section{Abstract}
    \fontsize{11.2pt}{15.85pt}\selectfont Tipping points occur in many real-world systems, at which the system shifts suddenly from one state to another. The ability to predict the occurrence of tipping points from time series data remains an outstanding challenge and a major interest in a broad range of research fields. Particularly, the widely used methods based on bifurcation theory are neither reliable in prediction accuracy nor applicable for irregularly-sampled time series which are commonly observed from real-world systems. Here we address this challenge by developing a deep learning algorithm for predicting the occurrence of tipping points in untrained systems, by exploiting information about normal forms. Our algorithm not only outperforms traditional methods for regularly-sampled model time series but also achieves accurate predictions for irregularly-sampled model time series and empirical time series. Our ability to predict tipping points for complex systems paves the way for mitigation risks, prevention of catastrophic failures, and restoration of degraded systems, with broad applications in social science, engineering, and biology.
	
	\section{Introduction}
	Many real-world systems, ranging from biological systems to climate systems and financial systems, can experience sudden shifts between states at critical thresholds which are so-called tipping points, for example, the epileptic seizures\cite{RN111}, abrupt shifts in ocean circulation\cite{RN75,RN110} and systemic market crashes\cite{RN118}. Accurately predicting tipping points before they occur has many important applications, including making strategies to prevent disease outbreak, avoiding disasters induced by climate change and designing robust financial systems. However, due to the complexity of various real-world systems, it is a challenge task to invent an effective tool to predict the occurrence of tipping points\cite{RN24,RN56}.
	
	A classical mathematical tool to understand tipping points is bifurcation theory which focuses on how dynamical systems undergo sudden qualitative changes as a parameter crosses a threshold\cite{RN66}. There are basically two classes of methods based on bifurcation theory to deal with tipping points prediction. The first class of methods is using lag-1 autocorrelation which is based on ``critical slowing down'', a paramount clue of whether a tipping point is approached\cite{RN24,RN53}. Such methods are widely used in various complex systems, including degenerate fingerprinting\cite{RN21}, BB method\cite{RN16} and ROSA\cite{RN39}. The second class of methods is using dynamical eigenvalue (DEV)\cite{RN58} which is based on Takens embedding theorem\cite{RN125}. Yet, the success of the existing methods based on bifurcation theory are impaired by two fundamental limitations. First, these methods are not applicable for irregularly-sampled time series data which are commonly observed from various scientific fields, such as geoscientific measurements\cite{RN103,RN101}, medical observations\cite{RN100} and biological systems\cite{RN102}. Second, the performance of existing methods based on bifurcation theory in prediction accuracy is affected by several approximations used in these methods. One such approximation is that only the first-order term of the dynamical systems is considered in these methods while the impact of the higher-order terms is ignored. However, the higher-order terms become significant for prediction when a tipping point is approached\cite{RN18} (Supplementary information Note S1). Besides, in the approximation of fast-slow systems\cite{RN3}, the delay on the bifurcation-tipping due to the changing rate of the bifurcation parameter is ignored\cite{RN25,RN97} (Supplementary information Note S2).
	
	Recent studies show that some machine learning based techniques have been used as effective early warning signals of tipping points by learning generic features of bifurcation\cite{RN18,RN108,RN199,RN28}. However, they can not be used for predicting where the tipping points occur. A machine learning framework based on reservoir computing has been developed for tipping points prediction\cite{RN29,RN200,RN201}. This algorithm requires time series sampled from all interacting variables of study system for training and prediction. However, this information is often not available for real-world systems. Therefore, it is often not feasible to use this machine learning framework for predicting tipping points of real-world systems.
	
	To circumvent above limitations of existing methods for tipping points prediction, we develop a deep learning (DL) algorithm based on 2D CNN-LSTM architecture. Based on the embedding theorem for irregular sampling\cite{RN205} (Supplementary information Note S3), this DL algorithm only requires the time series of the state and that of the bifurcation parameter from a single variable of a system (Fig. 1). The 2D CNN layer uses convolution kernels with length $d$ to extract features of the system reconstructed by a $d$-dimensional delay embedding. These convolutional kernels move along the time series, extracting features from each segment of the state time series and the bifurcation parameter time series, thereby generating sequences of feature representations. Then the LSTM layer is used to identify long-term dependencies from the sequences of features of the reconstructed system for tipping points prediction. We assume that the DL algorithm can detect features that emerge in time series prior to a tipping point, such as the features of the recovery rate in normal forms, which are associated with the occurrence of tipping points. We apply the DL algorithm to predict the occurrence of tipping points in systems it was not trained on. Our DL algorithm is applicable to both regularly-sampled and irregularly-sampled time series. We first test this DL algorithm on regularly-sampled time series generated by the models from ecology\cite{RN85,RN86,RN84} and climatology\cite{RN87,RN88,RN115,RN64}. The DL algorithm outperforms traditional methods in prediction accuracy for regularly-sampled model time series. We further validate our DL algorithm on irregularly-sampled time series generated by the same models and two other models from neuroscience with hysteresis phenomena\cite{RN105,RN107}. Finally, we validate our DL algorithm in irregularly-sampled empirical data from microbiology\cite{RN37} and thermoacoustics\cite{RN25}. In this work, we show that our DL algorithm is effective in dealing with dynamical systems exhibiting fold, Hopf and transcritical bifurcation. We anticipate that our DL algorithm may also be applicable to dynamical systems exhibiting other bifurcation types.
	\begin{figure}[htbp]
	\centering
		\includegraphics[width=0.9\linewidth]{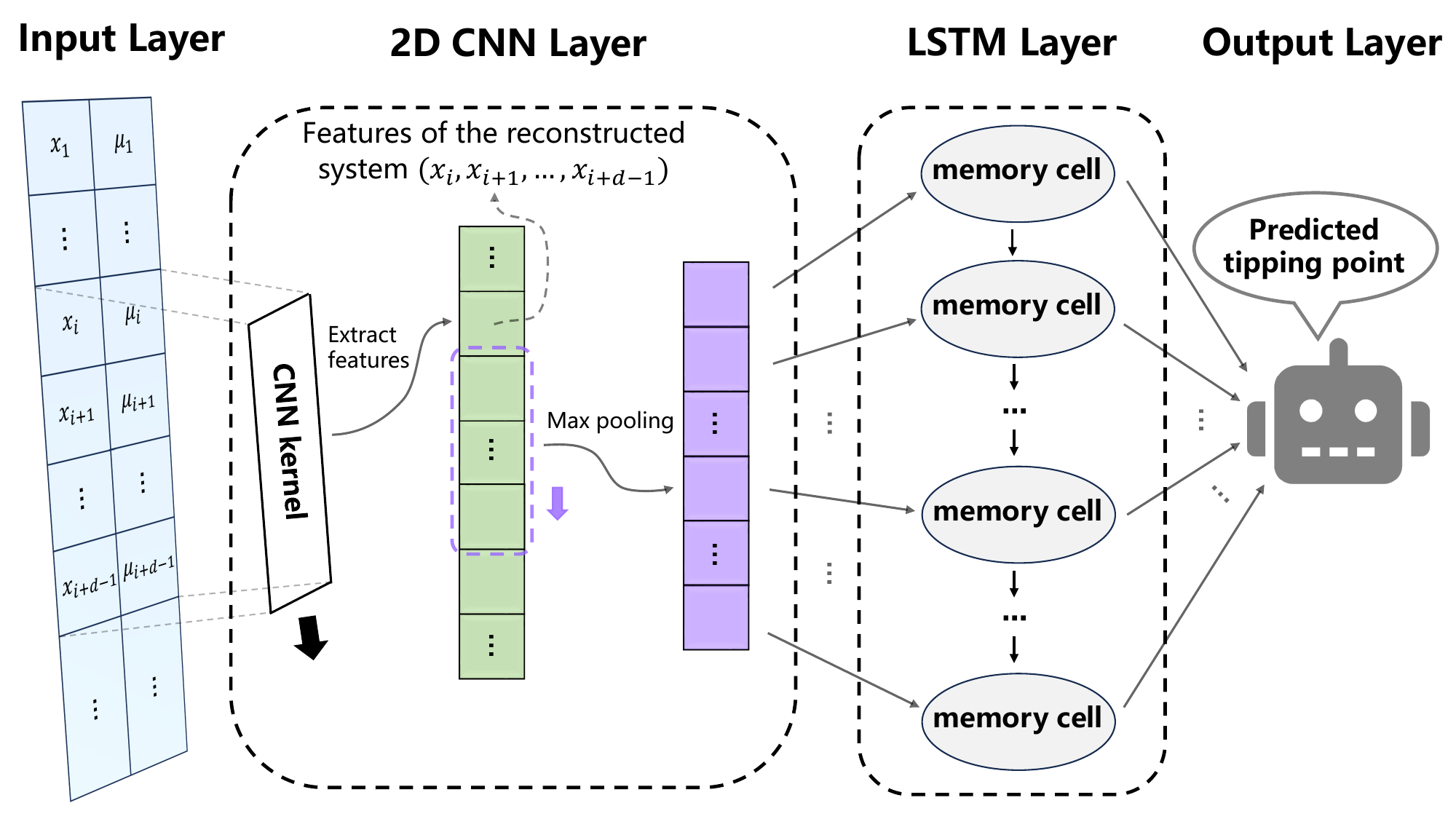}
        \caption*{\textbf{Fig. 1.}\fontsize{10pt}{12pt}\selectfont{
		The 2D CNN-LSTM architecture. The 2D convolutional kernel processes state series and bifurcation parameter series into one-dimensional features series of the reconstructed system $(x_i,x_{i+1},\dots,x_{i+d-1})$. This features series is then subjected to local max pooling, followed by analysis using an LSTM layer. Finally, the features processed by the LSTM are mapped to the predicted tipping point.}}
	\end{figure}
	
	\section{Results}
	\subsection{Bifurcation theory}
	The bifurcation theory is a classical and widely used mathematical tool to understand tipping points. According to the center manifold theorem\cite{RN66}, as a high-dimensional dynamical system approaches a bifurcation, its dynamics converges to a lower-dimensional space which exhibits dynamics topologically equivalent to those of the normal form of that bifurcation\cite{RN66}. Here we focused on codimension-one bifurcations in continuous-time dynamical systems, including the fold, Hopf, and transcritical bifurcation. Many tipping points of complex systems from the nature and the society are initiated by these three types of bifurcation\cite{RN18,RN63}. The normal forms of fold, Hopf, and transcritical bifurcation are shown in Fig. 2 A-C respectively.
	
	Suppose $dx/dt=f(x,\mu(t))$ has quasi-static attractor $x^*(\mu)$ where $\mu$ is the bifurcation parameter. Then the recovery rate is defined as the maximal real part of the eigenvalues of the Jacobian matrix when $x=x^*(\mu)$ in an $n$-dimensional dynamical system\cite{RN43,RN73}
	\begin{equation}
		\lambda = max(Re(eigvals(
		\begin{bmatrix}
			\frac{\partial f_1}{\partial x_1} & \frac{\partial f_1}{\partial x_2} & \cdots & \frac{\partial f_1}{\partial x_n} \\
			\frac{\partial f_2}{\partial x_1} & \frac{\partial f_2}{\partial x_2} & \cdots & \frac{\partial f_2}{\partial x_n} \\
			\vdots & \vdots & \ddots & \vdots \\
			\frac{\partial f_n}{\partial x_1} & \frac{\partial f_n}{\partial x_2} & \cdots & \frac{\partial f_n}{\partial x_n}
		\end{bmatrix}_{x=x^*(\mu)}))), \notag
	\end{equation}
	where $eigvals$ refers to the operation of computing eigenvalue and $Re$ refers to the operation of taking the real part. Note that when the recovery rate of a system shifts from negative value to positive value, the bifurcation occurs\cite{RN66}. In other words, if we know the relation between the recovery rate of a system and the bifurcation parameter, we can easily identify the tipping point for this system. It is also worth noting that the relation between the recovery rate and the bifurcation parameter is the same for the systems of the same bifurcation type in normal forms, as shown in Fig. 2 D-F respectively. We assume that our DL algorithm can detect features of the recovery rate for the system by training it on a training set generated from a sufficiently diverse library of possible dynamical systems with fold, Hopf and transcritical bifurcation. Then the DL algorithm can be used to predict the tipping points where the recovery rate becomes zero based on the detected relation between the recovery rate and the bifurcation parameter of the system. Note that we only focus on dynamical systems exhibiting fold, Hopf and transcritical bifurcation in this paper. In order to predict tipping points of other bifurcation types, we can expand the training library by including simulated data exhibiting those dynamics.

    \clearpage
	\begin{figure}[htbp]
	\centering
		\includegraphics[width=1\linewidth]{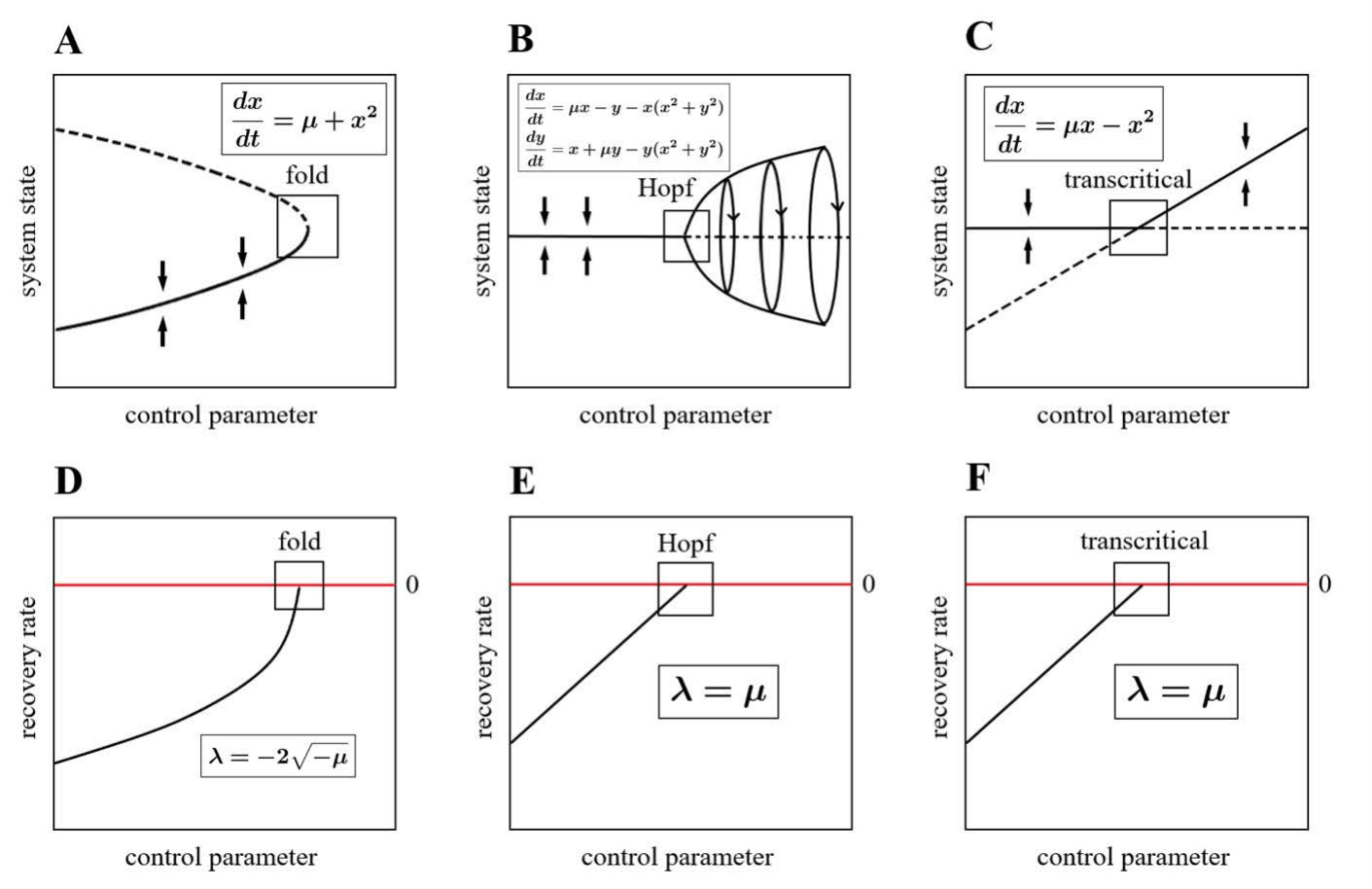}
        \caption*{\textbf{Fig. 2.}\fontsize{10pt}{12pt}\selectfont{
	(A-C) The normal forms of fold, (supercritical) Hopf, and transcritical bifurcation where bifurcation occurs at $\mu=0$ (hollow square). Their mathematical expressions are inside the boxes within the corresponding figures. (D-F) The recovery rate $\lambda$ of the normal forms for the fold, (supercritical) Hopf, and transcritical bifurcation as a function of the control parameter $\mu$, their function expressions are inside the boxes within the corresponding figures. Bifurcation occurs when the recovery rate $\lambda$ reaches zero (hollow square).}}
	\end{figure}

	\subsection{Performance of DL algorithm on model time series}
	We applied the DL model on regularly-sampled time series generated from three ecological models with white noise and three climate models with red noise. We generated test data for each model by changing the bifurcation parameter with eleven different initial values and five different changing rates. We generated 10 test time series for each initial value and changing rate of the bifurcation parameter (50 test time series for each initial value). To evaluate the performance of DL algorithm, we measured the relative error\cite{RN29} of tipping points prediction for each of the 50 test time series associated with every initial value of the bifurcation parameter. The performance of DL algorithm is evaluated by the mean relative error of tipping points prediction, which is averaged over the 50 measurements of the relative error of tipping points prediction for each initial value of the bifurcation parameter. We designed an ablation study by dropping out the 2D CNN layer, with LSTM serving as a competing algorithm. We first compared the results of DL algorithm with those of degenerate fingerprinting\cite{RN21}, DEV\cite{RN58} and LSTM in three ecological models with white noise which exhibit fold, Hopf and transcritical bifurcations respectively. Then we compared the results of DL algorithm with those of BB method\cite{RN16}, DEV\cite{RN58} and LSTM in three climate models with red noise which exhibit fold, Hopf and transcritical bifurcations respectively. As Fig. 3 shows, the DL algorithm outperforms the other competing algorithms for all initial values of each model. Moreover, the DL algorithm exhibits smaller fluctuations of relative error of tipping points prediction than the other competing algorithms for each initial value and exhibits smaller fluctuations of the mean relative error of tipping points prediction than the other competing algorithms across different initial values. Our results suggest that the DL algorithm is more robust against different initial values of the bifurcation parameter than competing algorithms.

    We further tested the DL model on irregularly-sampled time series from the above six models. It is worth noting that the above competing algorithms are not applicable for irregularly-sampled time series. Therefore, we used linear interpolation to transform these irregularly-sampled time series into equidistant data. This allows us to use the competing algorithms of degenerate fingerprinting, BB method and DEV for detecting early warning signals on reconstructed model time series\cite{RN74}. LSTM is also applied as a competing algorithm. From Fig. 4, we find that DL algorithm outperforms the other competing algorithms for all initial values of each model. Moreover, our results suggest that the DL algorithm is more robust against different initial values of the bifurcation parameter than competing algorithms.

    \begin{figure}[htbp]
	\centering
	\includegraphics[width=1\linewidth]{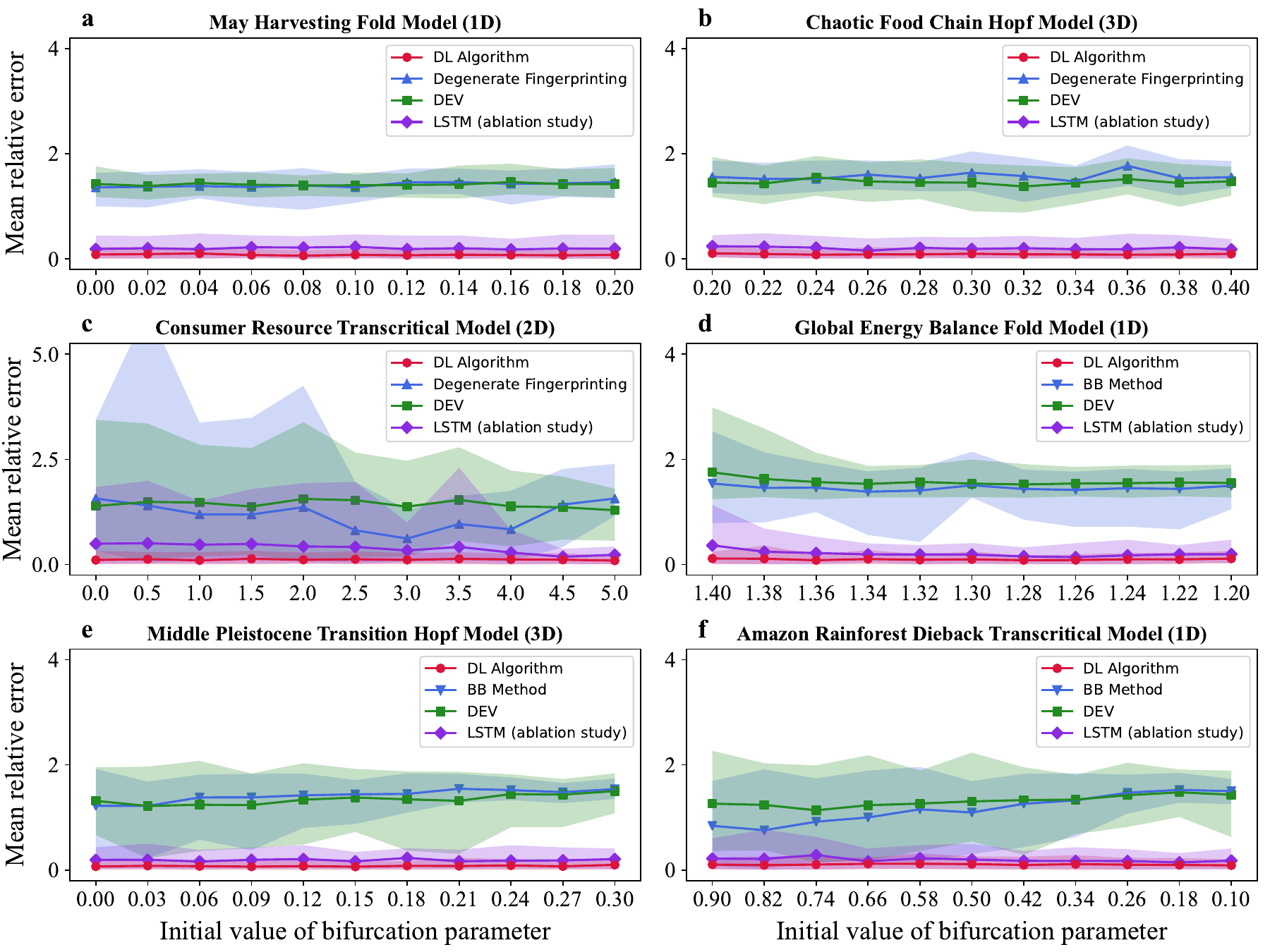}
	\caption*{\textbf{Fig. 3.}\fontsize{10pt}{12pt}\selectfont{
      We tested the DL model on regularly-sampled model time series with eleven initial values of bifurcation parameter. Here the mean relative error of tipping points prediction is plotted as line graphs against the initial value of the parameter. The area covered by the polyline represents the 90\% confidence interval for the relative error of tipping points prediction. (a-c) We compared the DL algorithm (red lines) with degenerate fingerprinting (blue lines), DEV (green lines) and LSTM (purple lines) on three ecological model time series with white noise. These model time series undergo fold, Hopf, and transcritical bifurcation, respectively. (d-f) The DL algorithm (red lines) is compared with BB method (blue lines), DEV (green lines) and LSTM (purple lines) on three climate model time series with red noise. These model time series undergo fold, Hopf, and transcritical bifurcation, respectively.}}
    \end{figure}

    In the ablation study, our DL algorithm achieves a mean relative error about 9\% in predicting tipping points on these model time series, while LSTM exhibits a mean relative error about 23\%, as shown in Supplementary information Fig. S1 and Fig. S2. These results suggest that the accuracy of predicting tipping points is higher when the LSTM layer receives sequences of features from the reconstructed system, rather than sequences of a single variable of the system. We also plotted the line graphs of the mean relative error of tipping points prediction against the distance between the final value of bifurcation parameter time series and the value of the tipping point, as shown in Supplementary information Fig. S3 and Fig. S4. Our results suggest that the DL algorithm is more robust against different distance to the tipping point than competing algorithms. Moreover, we plotted the line graphs of the mean relative error of tipping points prediction against the initial value of the parameter in five different changing rates of the bifurcation parameter, as shown in Supplementary information Fig. S5 and Fig. S6. Our results suggest that the DL algorithm is robust against various changing rates of bifurcation parameter.
    
    \begin{figure}[htbp]
	\centering
	\includegraphics[width=1\linewidth]{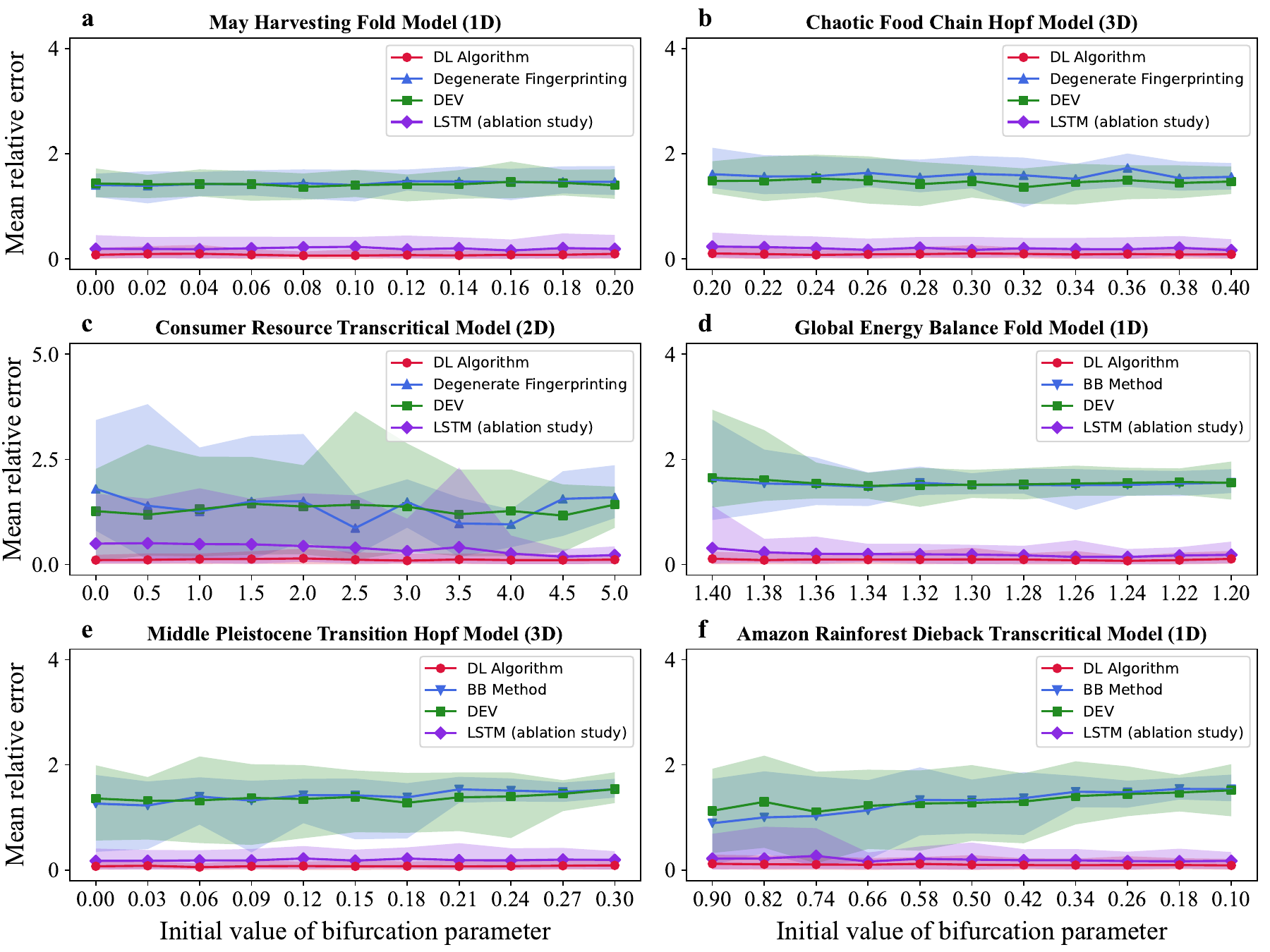}
	\caption*{\textbf{Fig. 4.}\fontsize{10pt}{12pt}\selectfont{
       We tested the DL model on irregularly-sampled model time series with eleven initial values of bifurcation parameter. Here the mean relative error of tipping points prediction is plotted as line graphs against the initial value of the parameter. The area covered by the polyline represents 90\% confidence interval for the relative error of tipping points prediction. (a-c) We compared the DL algorithm (red lines) with degenerate fingerprinting (blue lines), DEV (green lines) and LSTM (purple lines) on three ecological model time series with white noise. (d-f) The DL algorithm (red lines) is compared with BB method (blue lines), DEV (green lines) and LSTM (purple lines) on three climate model time series with red noise. We used linear interpolation to transform these irregularly-sampled time series into equidistant data so that they are suitable for degenerate fingerprinting, BB method and DEV.}}
    \end{figure}
    
    We then tested the DL model on irregularly-sampled time series from the ascending arousal system with a fold/fold hysteresis loop. We conducted irregular sampling of 400 points from twenty-two initial values of the bifurcation parameter, which are $0.1,0.2,\dots,1.1$ (bifurcation parameter increases from these initial values) and $1.9,1.8,\dots,0.9$ (bifurcation parameter decreases from these initial values). We applied the DL model on these irregularly-sampled time series and the mean relative errors of tipping points prediction are 3.12\% and 3.31\% respectively, for the increasing and decreasing bifurcation parameter cases, as shown in Fig. 5 (a). Similarly, we tested the DL model on irregularly-sampled time series from the Sprott B system with a Hopf/Hopf-hysteresis bursting. We irregularly sampled 400 points from twenty-two initial values of the bifurcation parameter, which are $\pi,1.04\pi,\dots,1.4\pi$ (bifurcation parameter increases from these initial values) and $2\pi,1.96\pi,\dots,1.6\pi$ (bifurcation parameter decreases from these initial values). We applied the DL model on these irregularly-sampled time series and the mean relative errors of tipping points prediction are 2.82\% and 2.54\% respectively, for the increasing and decreasing bifurcation parameter cases, as shown in Fig. 5 (b). Our results suggest that the DL algorithm is effective for tipping point prediction in theoretical models with hysteresis phenomenon.
    
   \begin{figure}[htbp]
	\centering
	\includegraphics[width=1\linewidth]{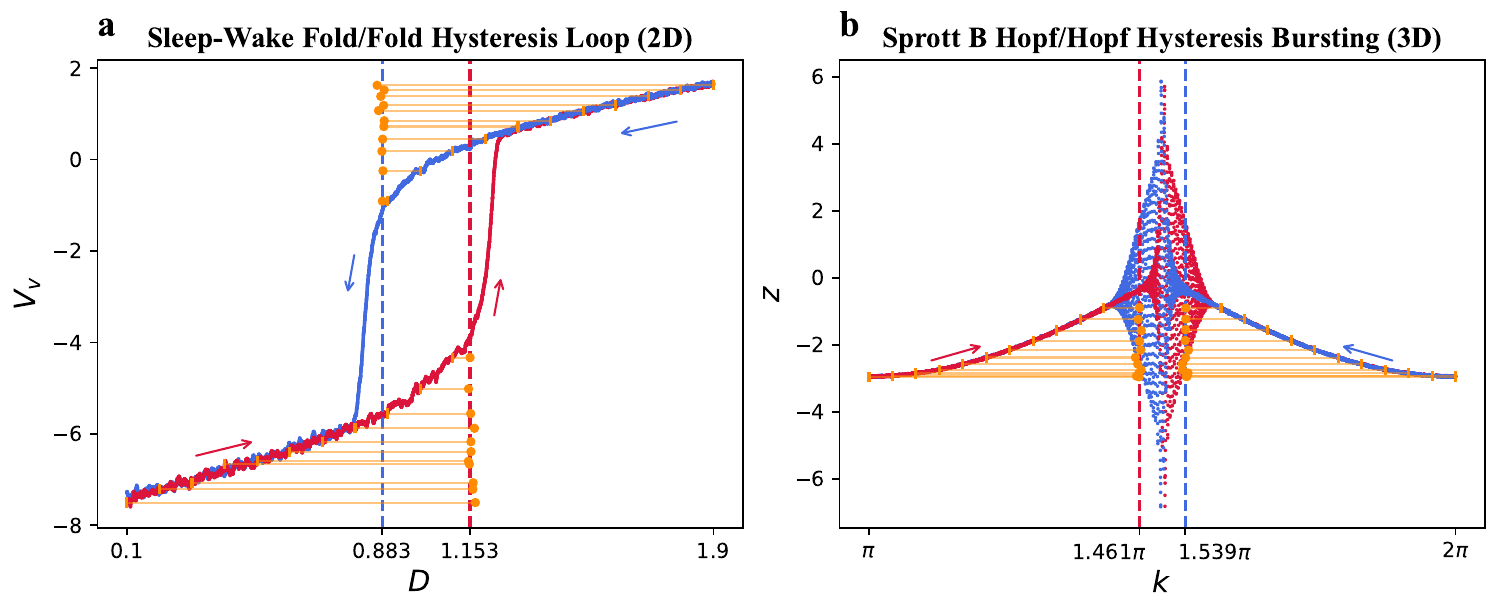}
	\caption*{\textbf{Fig. 5.}\fontsize{10pt}{12pt}\selectfont{
		The performance of the DL algorithm in predicting transitions of irregularly-sampled time series generated by theoretical models exhibiting hysteresis. The red and blue curves illustrate the time series forced by increasing and decreasing bifurcation parameter respectively. The orange dots denote the DL predictions, and the orange short vertical lines denote the initial points of the time series data used for prediction. We connect them with orange lines. (a) In the case of the sleep-wake system, a fold/fold hysteresis loop is observed at $D=1.153$ (red dashed line) and $D=0.883$ (blue dashed line) respectively. The predicted tipping points are between $[1.148,1.169]$ and $[0.8679,0.8898]$ respectively.  (b) For the Sprott B system, a Hopf/Hopf hysteresis bursting occurs at $k=1.461\pi$ (red dashed line) and $k=1.539\pi$ (blue dashed line) respectively. The predicted tipping points are between $[1.454\pi,1.464\pi]$ and $[1.534\pi,1.545\pi]$ respectively.
	}}
  \end{figure}
  
   \subsection{Performance of DL algorithm on empirical time series} 
   We tested the DL model on two empirical examples, including a cyanobacteria microcosm experiment under light stress\cite{RN37} and a physical experiment of voice oscillation during heat\cite{RN25}. For each empirical system, we applied the DL model to several irregularly-sampled time series. Each time series includes 400 data points with different initial values of the control parameter. Then we used linear interpolation to transform these irregularly-sampled records to time series with equidistant data. This allows us to use the competing algorithms of degenerate fingerprinting, BB method and DEV for detecting early warning signals on reconstructed empirical records\cite{RN74}. Note that these two empirical examples from ecology and physics are subject to white noise. Therefore, we applied degenerate fingerprinting to these empirical data and did not use the BB method for comparison. We compared the results of the DL algorithm with two competing algorithms, which are degenerate fingerprinting and DEV.
   
   In the cyanobacteria microcosm experiment under light stress, the photo-inhibition drives a cyanobacterial population, measured by light attenuation coefficient, towards a tipping point with fold bifurcation when a critical threshold of light irradiance is approached\cite{RN37}. We sampled seven time series with different initial values of light irradiance which are 477, 517, 557, 903, 944, 985 and 1025 $\rm \mu$mol photons $\rm m^{-2}s^{-1}$. Since the data between 879 and 903 $\rm \mu$mol photons $\rm m^{-2}s^{-1}$ is missing, the time series is sampled with initial values of 477, 517, 557 $\rm \mu$mol photons $\rm m^{-2}s^{-1}$ respectively and the same final value of 879 $\rm \mu$mol photons $\rm m^{-2}s^{-1}$. The mean relative error of the prediction with our DL algorithm is 3.82\%, where the ground truth of the tipping point is 1091 $\rm \mu$mol photons $\rm m^{-2}s^{-1}$. As shown in Fig. 6 A1-A3, the DL algorithm outperforms the other competing algorithms in all cases. These results suggest the robustness of the DL algorithm in tipping points prediction against different initial values of light irradiance.
   
   The second empirical data we analyzed is the thermoacoustic system. The state of the thermoacoustic system is measured by acoustic pressure. The thermoacoustic system undergoes a Hopf bifurcation from a non-oscillatory to an oscillatory state with increasing voltage\cite{RN28,RN91,RN90,RN76}. It has also been demonstrated that the transition to high amplitude limit cycle oscillations occurs later for faster changing rate of voltage, which can be explained by rate-delayed tipping\cite{RN25,RN97}. We sampled time series with nine initial values of voltage which are $0,0.2,\dots,1.6$ V under voltage changing rates of 20 mV/s and 40 mV/s, and with ten initial values of voltage which are $0,0.2,\dots,1.8$ V under a voltage changing rate of 60 mV/s. We then applied a change-point detection algorithm\cite{RN207} to detect tipping points of these empirical time series under voltage changing rates of 20 mV/s, 40 mV/s and 60 mV/s, which are 1.72 V, 1.76 V and 1.87 V respectively. As shown in Fig. 6 B1-B3, we find that the DL algorithm outperforms the other competing algorithms in predicting the tipping points in most cases under the voltage changing rate of 40 mV/s, with mean relative error of 4.31\% by our DL algorithm. We further test these algorithms for other two changing rates of voltage as shown in Supplementary information Fig. S7, which suggests the DL algorithms also shows best performance among all these algorithms. These results suggest that the DL algorithm is robust against various initial values and changing rages of voltage.
  
	\section{Discussion}
	Predicting the occurrence of tipping points based on time series is a challenging problem. In this work, we develop a deep learning algorithm that exploits information of normal form from the training systems for tipping points prediction. This algorithm can deal with both regularly-sampled and irregularly-sampled time series by using the time series sampled from a single variable of a system. Our results show that the DL algorithm not only outperforms traditional methods in the accuracy of tipping points prediction for regularly-sampled time series, but also achieve accurate prediction for irregularly-sampled time series. Our results pave the way to make effective strategies to prevent and prepare tipping points from various real-world systems\cite{RN77}.
	
	There are three major advantages of our work. First, traditional methods for tipping points prediction can only deal with regularly-sampled time series\cite{RN21,RN16,RN39,RN58}. However, the DL algorithm used in this paper can deal with both regularly-sampled time series and irregularly-sampled time series. This is because the convolution kernels in the 2D CNN layer can extract features of the reconstructed system from segments of irregularly-sampled time series (Supplementary information Note S3), which is used for tipping points prediction in the LSTM layer. Second, most existing methods\cite{RN21,RN204,RN29,RN200,RN201} require information of multiple interacting variables in the study system for predicting tipping points. However, based on the embedding theorem for irregular sampling\cite{RN205}, the DL algorithm only requires the time series sampled from a single variable of the study system. Third, the effect of rate-delayed tipping has been ignored in previous methods for tipping points prediction\cite{RN21,RN16,RN39,RN58}. This is due to the limitations of the theory of fast-slow systems used in previous methods. However, we consider the effect of rate-delayed tipping in our work and label the data by using the fact that the quasi-static attractor loses stability at the occurrence of tipping points in our training set (see Methods for details). The DL algorithm trained with such labels is robust against different changing rates of bifurcation parameter (Supplementary information Figs. S5-S7).
	
	The embedding theorem for irregular sampling\cite{RN205} requires that the dimension $d$ of the system reconstructed by delay embedding must be larger than twice the dimension $m$ of the study system (Supplementary information Note S3). Therefore, the length of convolution kernels used in the CNN layer of our DL algorithm is required to be larger than $2m$. However, due to the time-varying nonstationarity of dynamical system approaching a bifurcation, the features extracted from shorter convolutional kernel should contain much more dynamical information of the system, compared to those extracted from longer convolutional kernel\cite{RN197}. Therefore, our DL algorithm may not perform well in predicting tipping points in high-dimensional systems. Another limitation of our DL algorithm is that it requires the existence of tipping points as prior knowledge. If there is no tipping point in the system, the output of the DL algorithm is meaningless. Therefore, several methods such as DEV\cite{RN58} or some DL classifiers\cite{RN18,RN108,RN199,RN28} should be used to determine whether a system is approaching a tipping point first. Then our DL algorithm can be applied to predict the specific location of a tipping point if it exists.

    \begin{figure}[htbp]
	\centering
	\includegraphics[width=1.0\linewidth]{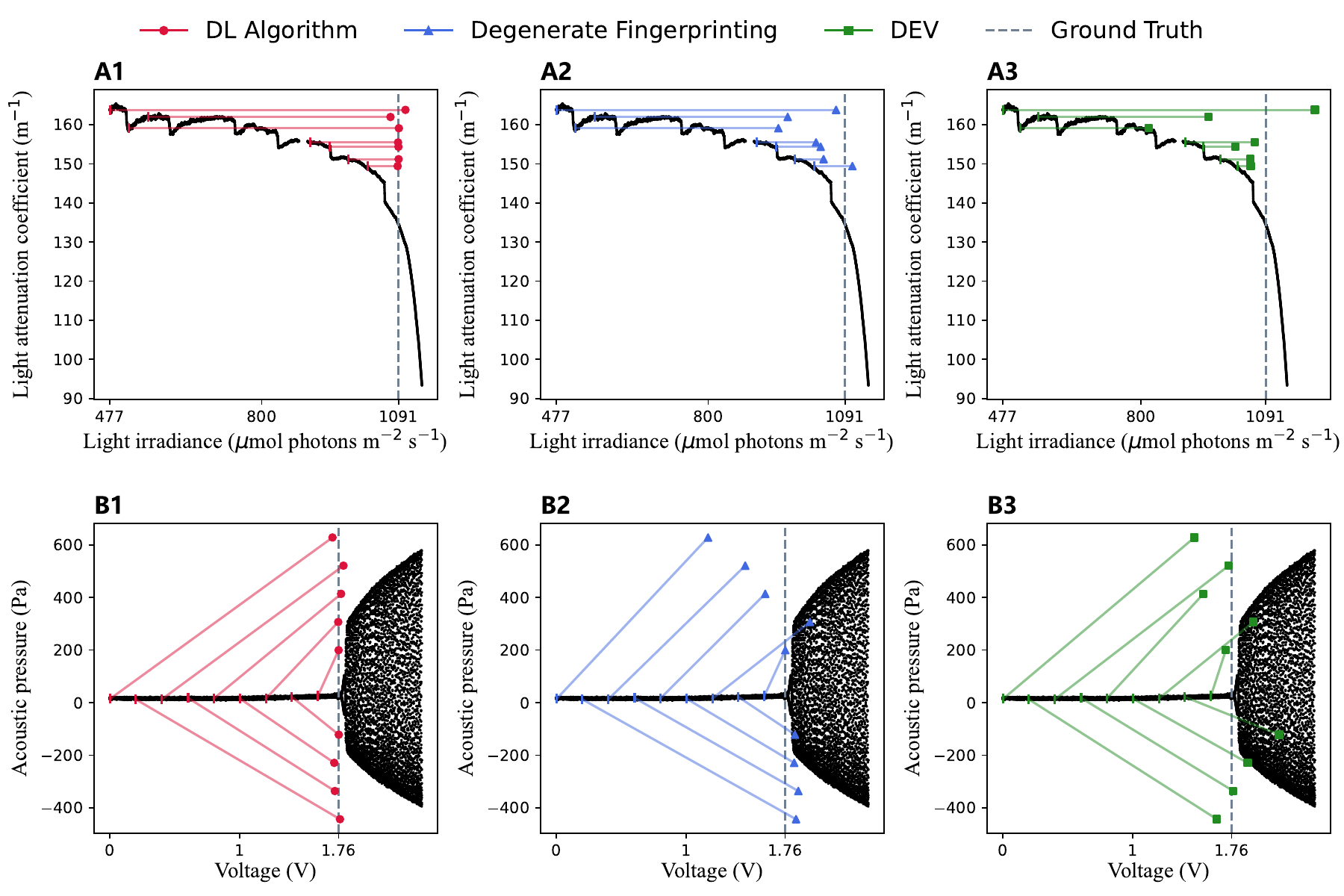}
	\caption*{\textbf{Fig. 6.}\fontsize{10pt}{12pt}\selectfont{
		The performance of DL algorithm in predicting tipping points on irregularly-sampled empirical time series. The dots denote the DL predictions, and the short vertical lines denote the initial points of the time series data used for prediction. We connect them with lines. (A1-A3) Cyanobacterial population undergoes a fold bifurcation, which is measured by light attenuation coefficient. The fold bifurcation occurs at 1091 $\rm \mu$mol photons $\rm m^{-2}s^{-1}$, while the predictions by our DL algorithm are between 1074 $\rm \mu$mol photons $\rm m^{-2}s^{-1}$ and 1106 $\rm \mu$mol photons $\rm m^{-2}s^{-1}$. (B1-B3) The thermoacoustic system undergoes a Hopf bifurcation under the 40 mV/s changing rate of the control parameter voltage. The Hopf bifurcation occurs at 1.76 V, while the predictions by our DL algorithm are between 1.71 V and 1.79 V. We compared the performance of the DL algorithm (red) with degenerate fingerprinting (blue) and DEV (green). We used linear interpolation to transform these irregularly-sampled time series into equidistant data so that they are suitable for degenerate fingerprinting and DEV.
	}}
    \end{figure}
	
	To analyze whether the DL algorithm uses the recovery rate in the normal form for prediction, we designed two control experiments. First, we trained three DL models on three datasets, each consisting solely of time series with fold, Hopf, and transcritical bifurcation, respectively. Then we applied these three DL models on irregularly-sampled model time series from six theoretical models. We found that the DL model trained on dataset containing the features of a bifurcation perform better in predicting tipping points of test time series with that bifurcation, compared to another two trained on datasets without the corresponding normal form features (Supplementary information Fig. S8). These results suggest that the DL algorithm can extract the features of normal form. Second, we trained two DL models on two datasets, each consisting solely of time series with supercritical and subcritical pitchfork bifurcation, respectively (Supplementary information Note S7). Then we applied these two DL models on irregularly-sampled model time series from a model with supercritical pitchfork, where both models achieved similar performance in predicting the tipping points (Supplementary information Fig. S9). The normal forms of supercritical and subcritical pitchfork bifurcations exhibit the same relation between the recovery rate and the bifurcation parameter, and only differ in the cubic term. Thus, the DL model trained on time series with subcritical pitchfork bifurcation can be used to predict tipping points of time series with supercritical pitchfork bifurcation, which suggest that the DL algorithm can extract the features of the recovery rate in normal form.
	
	Our work raises several problems worthy of future pursuit. First, we have studied the situation of local codimension-one bifurcations in continuous-time dynamical systems in this paper. For future work, we can focus on tipping points prediction in discrete-time dynamical systems, such as period-doubling bifurcation\cite{RN66} which arise naturally in physiology\cite{RN128,RN130} and ecology\cite{RN129}. It would be interesting to develop a method that can deal with both regularly-sampled and irregularly-sampled time series for systems with period-doubling bifurcation. Second, it would be interesting to investigate tipping points prediction in systems with other types of bifurcation, such as codimension-two bifurcation and global bifurcation\cite{RN66}. Yet, tipping points prediction in systems with codimension-two bifurcation and global bifurcation are more challenging than that in systems with codimension-one bifurcation. Third, one can develop interpretable machine learning models for tipping points prediction by combing dynamical system theory with neural networks\cite{RN119,RN120,RN121,RN122,RN123}, which offer an avenue for making safe and reliable high-stakes decisions for policy makers\cite{RN124}.
	
	\section{Methods}
	\subsection{Generation of training data for the DL algorithm}
	We construct two-dimensional dynamical systems of the following form which are used to generate the training data based on simulation\cite{RN18}
	\begin{equation}
		\begin{aligned}
			\frac{dx}{dt}=\sum\limits_{i=1}^{10}a_ip_i(x,y)  \\
			\frac{dy}{dt}=\sum\limits_{i=1}^{10}b_ip_i(x,y),
		\end{aligned}
		\tag{1}
	\end{equation}
	where $x$ and $y$ are state variables, $p(x,y)$ is a vector containing all polynomials in $x$ and $y$ from zero up to third order
	\begin{equation}
		p(x,y)=(1,x,y, x^2,xy,y^2,x^3,x^2y,xy^2,y^3), \notag
	\end{equation}
	$p_i(x,y)$ is the $i$-th component of $p(x,y)$. $a_i$ and $b_i$ are parameters randomly drawn from standard normal distribution, and then, half of these parameters are selected at random and set to zero. The parameters for the cubic terms are set to the negative of their absolute values to encourage models with bounded solutions. Since our training data is required to contain representation of all possible dynamics that may occur in the study time series data, we generate models with different sets of parameter values in Eqs. 1 until a required number of each type of bifurcation (fold, Hopf, or transcritical) has been discovered. We add white or red noise to these models to simulate training data for the DL algorithm.
    
    After a model is generated, we perform numerical simulation of this model with 10,000 time steps from a randomly drawn initial condition and test whether the system converges to an equilibrium point. The odeint function from the Python package Scipy\cite{RN92} is applied in the numerical simulation with a step size of 0.01. The criteria that we used to determine the convergence is that the maximum difference between the final 10 points in numerical simulation is less than $10^{-8}$. The convergence is required in order to search for bifurcations which occur at non-hyperbolic equilibria\cite{RN66}. The models that do not converge are discarded. For the models that converge, we apply AUTO-07P\cite{RN93} to identify bifurcations along the equilibrium branch by either increasing or decreasing each nonzero parameter. For each bifurcation identified, we first set the initial condition with the value of the equilibrium of the model and a burn-in period of 100 units of time. Then we run simulations of the model with noise and obtain the quasi-static attractor time series with noise and bifurcation parameter time series which are used for training. We also run simulations of the same model without noise and obtain the corresponding quasi-static attractor time series and bifurcation parameter time series which are used for calculating the recovery rate $\lambda$ to locate the bifurcation point where $\lambda$ changes from negative value to positive value,
    \begin{equation}
    	\lambda=max(Re(eigvals(
    	\begin{bmatrix}
    		\frac{\partial (\sum\limits_{i=1}^{10}a_ip_i(x,y))}{\partial x} & \frac{\partial (\sum\limits_{i=1}^{10}a_ip_i(x,y))}{\partial y} \\
    		\frac{\partial (\sum\limits_{i=1}^{10}b_ip_i(x,y))}{\partial x} & \frac{\partial (\sum\limits_{i=1}^{10}b_ip_i(x,y))}{\partial y}
    	\end{bmatrix}_{(x,y)=(x^*,y^*)}))). \tag{2}
    \end{equation}
    The simulations of the models with noise are based on the Euler Maruyama method\cite{RN98} with a step size of 0.01 and the simulations of the models without noise are based on Euler method\cite{RN99} with the same step size.
    
    For each model where a bifurcation is identified, we perform five independent simulations of the model with varying bifurcation parameter from its initial value to a terminal value beyond the bifurcation point given by AUTO-07P. In these five simulations, we set the changing rate of the bifurcation parameter to remain constant within each simulation where the changing rate of the bifurcation parameter is the change in the bifurcation parameter per unit time with a step size of 0.01, but vary across different simulations. The ratio of each changing rate of the bifurcation parameter to the smallest changing rate of the bifurcation parameter in these five simulations is drawn from $1,2,\dots,10$. The nonzero changing rate of the bifurcation parameter can lead to the delay in the occurrence of tipping points, which is called rate-delayed tipping\cite{RN25,RN97}. Due to rate-delayed tipping, the theoretical bifurcation point given by AUTO-07P is not the accurate location of the tipping point. Therefore, we identify the location of the tipping point by the recovery rate (Eq. 2) and use this identified tipping point as the training label for our supervised deep learning training process (Supplementary information Note S2). If no tipping point is detected in a simulation, this simulated data will be discarded. Otherwise, we randomly and irregularly sample $l_s$ points from quasi-static attractor time series with noise which is simulated by the variable $x$ of the model and corresponding bifurcation parameter time series between the initial value and the tipping point, where $l_s$ is drawn from $Uniform(505,1000)$. Then we select the first 500 points as training data. The obtained training data can make DL algorithm to learn the features of recovery rate from time series data with different changing rates of parameter and varying distances between the end of the time series and the tipping points.
	
    Here we train deep learning algorithm on datasets with white and red noise. The white noise in simulation has the amplitude drawn from a triangular distribution centered at 0.01 with upper and lower bounds 0.0125 and 0.0075 respectively while the red noise is modeled by AR(1) process in which the lag-1 autoregressive coefficient is between -1 and 1. It is worth noting that the simulations shift to new regime before the bifurcation point due to the noise\cite{RN67,RN27}. Since the location of this premature transition is stochastical, we still use the location where the quasi-static attractor without noise loses stability as the training label for DL algorithm even it is not accurate. We mathematically illustrate how slight perturbation causes bifurcation to occur earlier and its stochasticity in Supplementary information Note S6.
	\subsection{DL algorithm architecture and training}
	In this paper we use the 2D CNN-LSTM DL algorithm\cite{RN42,RN1}. The 2D CNN-LSTM architecture is shown in Fig. 1. We use a train/validation/test split of 0.95/0.04/0.01 for the DL algorithm, and MSE as the loss function to minimize the difference between the real tipping points (training labels) and the predictions. The DL algorithm is trained on 300,000 instances for 200 epoches. We train ten neural networks and report the performance of the DL algorithm averaged over them. The hyperparameters of the DL model are tuned through random search, and we present the optimal hyperparameters of the DL model in Supplementary information Table S1.
	 
	Before training our DL algorithm, there are several preprocessing matters\cite{RN19}. The quasi-static attractor time series with noise are detrended using Lowess smoothing with a span of 0.2 to obtain the residual time series used for training\cite{RN206}. Besides, in order to obtain the robustness of DL algorithm on time series of various lengths, we zero out the left side of the residual time series and corresponding bifurcation parameter time series, with the length drawn from $Uniform(0,250)$. Due to the zeroing process, the DL algorithm can deal with time series even if it is a little short. After the zeroing process, each residual time series is normalized by dividing each time series data point by the average absolute value of the residuals across the non-zero part of the time series. In addition, each bifurcation parameter time series is also normalized: each data point in the time series minus the initial value of the bifurcation parameter time series, and then divided by the distance between the initial and final values of the bifurcation parameter time series, thereby mapping the bifurcation parameter time series to the interval of $[0,1]$. The corresponding label, i.e., the tipping point, is normalized following the same procedure as the normalization of the data point in bifurcation parameter time series. Thus the relative distance of tipping point to the time series in the training set for the DL algorithm (calculated as the tipping points minus the final value of the bifurcation parameter time series, divided by the distance between the initial and final values of the bifurcation parameter time series) is between 0.01 and 2. In other words, the labels in the training set, after normalization, will fall within the interval of $[1.01,3]$.
	\subsection{Theoretical models used for testing}
	The simulation of theoretical models with noise is based on the Euler Maruyama method\cite{RN98} with a step size of 0.01 ($\delta t=0.01$ units of time) unless otherwise stated. The amplitude of white noise $\sigma$ is 0.01 while the red noise is modeled by AR(1) process in which the lag-1 autoregressive coefficient $\phi$ is drawn from $Uniform(-1,1)$. We set five different changing rates of the bifurcation parameter in simulations of each theoretical model. The bifurcation point of each simulation is the location where the recovery rate changes from negative value to positive value. Then we sample from these simulations to obtain regularly-sampled model time series and irregularly-sampled model time series for testing. The lengths of these model time series range from 250 to 500, and the relative distances of tipping points to these time series range from 0.01 to 2.
	
	We apply the relative error $\varepsilon$ of tipping points prediction to evaluate the performance of the DL algorithm, which is defined as
    \begin{equation}
		\varepsilon = \frac{\left |\widehat{\mu}_c-\mu_c \right |}{\left |\mu_{end}-\mu_c \right |}, \notag
	\end{equation}
	where $\widehat{\mu}_c$ is the predicted tipping point, $\mu_c$ is the real tipping point, $\mu_{end}$ is the value of bifurcation parameter for last point of the test time series.
	 
	 \subsubsection{Theoretical models with white noise}
	 To test the fold bifurcation with white noise, we use the May’s harvesting model\cite{RN85}. This is given by
	 \begin{equation}
	 	\frac{dx}{dt} = rx(1-\frac{x}{k})-\frac{hx^2}{s^2+x^2}+ \sigma \xi(t), \notag
	 \end{equation}
	 where $x$ represents biomass of some population, $k$ is its carrying capacity, $h$ is the harvesting rate, $s$ characterizes the nonlinear relationship between harvesting output and current biomass, $r$ is the intrinsic per capita growth rate of the population, and $\xi(t)$ is a Gaussian white noise process. We use parameter values $r=1$, $k=1$, $s=0.1$ and $h$ increases at rates of $1\times 10^{-5},2\times 10^{-5},\dots,5\times 10^{-5}$. We generate model time series by this equation from eleven initial values of $h$, which are $0,0.02,0.04,\dots,0.18,0.2$. In this configuration, the fold bifurcation occurs in $h \in [0.2713,0.2926]$.
	 
	 To test the Hopf bifurcation with white noise, we use a three-species chaotic food chain model\cite{RN86}. This is given by
	 \begin{equation}
	 	\begin{aligned}
	 	&\frac{dr}{dt} = r(1-\frac{r}{k})-\frac{x_cy_ccr}{r+r_0} + \sigma \xi_r(t),  \notag \\
	 	&\frac{dc}{dt} = x_cc(\frac{y_cr}{r+r_0}-1) - \frac{x_py_ppc}{c+c_0} + \sigma \xi_c(t), \notag \\
	 	&\frac{dp}{dt} = x_pp(\frac{y_pc}{c+c_0}-1) + \sigma \xi_p(t),  \notag
	 	\end{aligned}
	 \end{equation}
	 where $r$, $c$, and $p$ represent the population densities of the resource, consumer, and predator species, respectively. $k$ characterizes the environmental capacity of the resource species. $x_c$, $y_c$, $x_p$, $y_p$, $r_0$, and $c_0$ are other parameters in the system. $\xi_r(t)$, $\xi_c(t)$ and $\xi_p(t)$ are independent Gaussian white noise processes. For our simulations, we use parameter values $x_c = 0.4$, $y_c=2.009$, $x_p=0.08$, $y_p=2.876$, $r_0=0.16129$, $c_0=0.5$ and $k$ increases at rates of $1\times 10^{-5},2\times 10^{-5},\dots,5\times 10^{-5}$. We generate model time series by these equations from eleven initial values of $k$, which are $0.20,0.22,0.24,\dots,0.38,0.40$. In this configuration, the Hopf bifurcation occurs in $k \in [0.4656,0.4839]$.
	 
	 To test the transcritical bifurcation with white noise, we use the Rozenzweig-MacArthur consumer-resource model\cite{RN84}. This is given by
     \begin{equation}
		\begin{aligned}
	 	&\frac{dx}{dt} = gx(1-\frac{x}{k}) - \frac{axy}{1+ahx} + \sigma \xi_x(t),  \notag \\
        &\frac{dy}{dt} = \frac{eaxy}{1+ahx} - my + \sigma \xi_y(t),  \notag
		\end{aligned}
	\end{equation}
	 where $x$ and $y$ represent the population densities of the resource, consumer, respectively. $g$ is the intrinsic per capita growth rate of $x$, $k$ is its carrying capacity, $a$ is the attack rate of $y$, $e$ is the conversion factor, $h$ is the handling time, $m$ is the per capita consumer mortality rate, $\xi_x(t)$ and $\xi_y(t)$ are independent Gaussian white noise processes. We fix the parameter values $r=4$, $k=1.7$, $e=0.5$, $h= 0.15$, $m=2$ and $a$ increases at rates of $1\times 10^{-4},2\times 10^{-4},\dots,5\times 10^{-4}$. We generate model time series by these equations from eleven initial values of $a$, which are $0,0.5,1.0,\dots,4.5,5.0$. In this configuration, the transcritical bifurcation in $a \in [5.8819,5.8823]$.
	 
	 \subsubsection{Theoretical models with red noise}
	 To test the fold bifurcation with red noise, we use a climate model describing temperature of an ocean on a spherical planet subjected to radiative heating\cite{RN88,RN87} which can simulate a transition from a greenhouse to an icehouse Earth\cite{RN74}. The model is given by
	 \begin{equation}
	 \begin{aligned}
	 	&\frac{dT}{dt} = \frac{-e\rho T^4 + \frac{1}{4}uI_0(1-a_p)}{c} + \eta(t), \notag \\
         &\text{with}\quad a_p = a-bT, \notag
	 	\end{aligned}
	 \end{equation}
	 where $T$ represents the average temperature of ocean, $u$ is relative intensity of solar radiation, $e$ is effective emissivity, $I_0$ is solar irradiance, $c$ is a constant thermal inertia, and $a_p$ is the planetary albedo. $a$ and $b$ are parameters which define a linear feedback between ice and albedo variability and temperature. $\eta(t)$ is a red noise process with lag-1 autoregressive coefficient $\phi$. We use $e=0.69$, $I_0=71944000$, $c=10^8$, $a=2.8$, $b=0.009$, $\rho=0.03$, $\delta t=1$ time unit and $u$ decreases at rates of $5\times 10^{-7},6\times 10^{-7},\dots,9\times 10^{-7}$. We generate model time series by this equation from eleven initial values of $u$, which are $1.4,1.38,1.36,\dots,1.22,1.2$. In this configuration, the fold bifurcation occurs in $u \in [0.9596,0.9631]$.
	 
	 To test the Hopf bifurcation with red noise, we use the middle Pleistocene transition system\cite{RN115}, explaining the dynamics of glacial cycles, which is given by
	 \begin{equation}
	 \begin{aligned}
	 	 &\frac{dx}{dt} = -x-y + \eta_x(t), \notag \\
         &\frac{dy}{dt} = -pz+uy+sz^2-yz^2 + \eta_y(t), \notag \\
         &\frac{dz}{dt} = -q(x+z) + \eta_z(t), \notag
	 	\end{aligned}
    \end{equation}
    where $x$, $y$ and $z$ represent the global ice volume, the atmospheric greenhouse gas concentration and the deep ocean temperature, respectively. All variables are rescaled to dimensionless form, $p=1$, $q=1.2$, $s=0.8$ are parameters, and $u$ increases at rates of $1\times 10^{-5},2\times 10^{-5},\dots,5\times 10^{-5}$. $\eta_x(t)$, $\eta_y(t)$ and $\eta_z(t)$ are independent red noise processes with lag-1 autoregressive coefficient $\phi$. We generate model time series by these equations from eleven initial values of $u$, which are $0,0.03,0.06,\dots,0.27,0.3$. In this configuration, the Hopf bifurcation occurs in $u \in [0.3546,0.3547]$.
	 
	 To test the transcritical bifurcation with red noise, we use the simplified version of TRIFFID dynamic global vegetation model\cite{RN64,RN23}. It can be used to simulate the dieback of the Amazon rainforest\cite{RN23}, which is mainly driven by climate change\cite{RN36}. The model is given by
	 \begin{equation}
	 	\begin{aligned}
	 	&\frac{dV}{dt} = PV^*(1-V) - GV + \eta(t), \notag
	 	\end{aligned}
	 \end{equation}
	 where $V$ represents the broadleaf fraction, $G$ is a disturbance coefficient (0.004/year) and $V^*$ is either the value of $V$ or 0.1 if $V$ falls below 0.1. $\eta(t)$ is a red noise process with lag-1 autoregressive coefficient $\phi$. $P$ is the productivity, in dimensionless area fraction units, it decreases at rates of $1\times 10^{-5},2\times 10^{-5},\dots,5\times 10^{-5}$. We generate model time series by this equation from eleven initial values of $P$, which are $0.90,0.82,0.74,\dots,0.18,0.10$. In this configuration, the transcritical bifurcation occurs in $P \in [-0.0061,-0.0046]$.
	 \subsubsection{Theoretical models with hysteresis phenomenon}
     We also test our DL model on irregularly-sampled model time series generated by two theoretical neuroscience models with white noise. These systems exhibit hysteresis, which suggests that when the bifurcation parameter changes in opposite directions, these system will undergoes sudden shifts at different bifurcation points. The ascending arousal system is modeled in terms of the neuronal populations and their interactions\cite{RN105}, which is given by
	 \begin{equation}
	 	\begin{aligned}
          &\tau_v\frac{dV_v}{dt} = -V_v + v_{vm}Q_m + D +\sigma \xi_v(t), \notag \\
          &\tau_m\frac{dV_m}{dt} = -V_m + v_{ma}Q_a + v_{mv}Q_v +\sigma \xi_m(t). \notag
	 	\end{aligned}
	 \end{equation}
	 Each population $j=v,a,m$, where $v$ is ventrolateral preoptic area, $a$ is acetylcholine, and $m$ is monoamines. Each population $j$ has a mean cell body potential $V_j$ relative to resting and a mean firing rate $Q_j$. The relation of $Q_j$ to $V_j$ is described by $Q_j=Q_{max}/[1+exp(-(V_j-\theta)/\sigma)]$, where $Q_{max}$ is the maximum possible rate, equals 100$\rm s^{-1}$. Besides, $V_a$ is constant, $\tau_j$ is the decay time for the neuromodulator expressed by group $j$, the $v_{jk}$ weights the input from population $k$ to $j$, $\xi_v(t)$ and $\xi_m(t)$ are independent Gaussian white noise processes. The model parameters are consistent with physiological and behavioral measures, $\theta=10$mV, $\sigma=3$mV, $v_{ma}Q_a=1$mV, $v_{vm}=v_{mv}=-1.9$mVs and $\tau_m=\tau_v=10$s. In our simulations within the interval of $[0.1,1.9]$, we force the parameter $D$ to increase from 0.1 or decrease from 1.9 at a rate of $1/7200$mV per unit time. In this configuration, a fold/fold-hysteresis loop occurs at $D=1.153$ (in the increasing direction) and $D=0.883$ (in the decreasing direction) respectively.
	  
	 Another system is Sprott B system\cite{RN106} with a single excitation, which can be used for researching bursting oscillation in neuroscience\cite{RN107}. The model is given by
	 \begin{equation}
	  	\begin{aligned}
	  		&\frac{dx}{dt}=a(y-x) + \sigma \xi_x(t), \notag \\
	  		&\frac{dy}{dt}=xz +\beta \cos(k) + \sigma \xi_y(t), \notag \\
	  		&\frac{dz}{dt}=b - xy + \sigma \xi_z(t), \notag
	  	\end{aligned}
	 \end{equation}
	 where $a=8$, $b=2.89$ and $\beta=5$. $\xi_x(t)$, $\xi_y(t)$ and $\xi_z(t)$ are independent Gaussian white noise processes. We force the parameter $k$ to increase from $\pi$ or decrease from $2\pi$ at a rate of $\pi \times 10^{-3}$ within the interval of $[\pi,2\pi]$. A Hopf/Hopf-hysteresis bursting occurs at $k=1.461\pi$ (in the increasing direction) and $k=1.539\pi$ (in the decreasing direction) respectively in this configuration.
	 \subsection{Empirical systems used for testing}
	 In this work, we use two empirical examples in the fields of microbiology and thermoacoustics to evaluate the performance of the DL algorithm.
	 
	 In the first example, photo-inhibition drives cyanobacterial population to a fold bifurcation when a critical light level is exceeded\cite{RN37}. In this system, bifurcation parameter light irradiance starts at 477 $\rm \mu$mol photons $\rm m^{-2}s^{-1}$ and is increased in steps of 23 $\rm \mu$mol photons $\rm m^{-2}s^{-1}$ each day. The time series includes 7,784 data points spanning overall 28.86 days with time interval equal to 0.0035 day (5 min).
	 
	 The second example is a thermoacoustic system where positive feedback between the unsteady heat release rate fluctuations and the acoustic field in the confinement can result in a transition to high amplitude limit cycle oscillations\cite{RN76}. Induja et al.\cite{RN25} conduct experiments in a thermoacoustic system exhibiting Hopf bifurcation. They pass several constant mass flow rates of air through a horizontal Rijke tube which consists of an electrically heated wire mesh in a rectangular duct and increase the voltage applied across the wire mesh to attain the transition to thermoacoustic instability. In such case, they observe that the occurrence of Hopf bifurcation depends on the changing rate of control parameter. We select three experimental sets for our study with the bifurcation parameter of voltage increasing at rates of 20 mV/s, 40 mV/s, and 60 mV/s. The voltage is ranging from 0 V to 2.4 V. The corresponding time series data consists of 1,200,000, 600,000, and 400,000 data points, respectively.
	 
	\subsection{Comparison of DL algorithm with competing algorithms}
	For detecting early warning signals (EWS) in quasi-static attractor time series with white noise, H. Held and T. Kleinen\cite{RN21} develop a technique called degenerate fingerprinting using PCA to approximate the critical mode, then estimating the lag-1 autoregressive coefficient of critical mode as an indicator (Supplementary information Note S5). Boettner and Boers\cite{RN33,RN16} propose an unbiased estimate of the lag-1 autoregressive coefficient for time series with red noise which we refer to as the BB method (Supplementary information Note S5). Florian Grziwotz et al.\cite{RN58} introduce a EWS robust to limited level of noise, named dynamical eigenvalue (DEV), which is rooted in bifurcation theory of dynamical systems to estimate the dominant eigenvalue of the system (Supplementary information Note S5). After obtaining EWS through these three approaches, linear regression or nonlinear fit between EWS and bifurcation parameters is used to make an extrapolation to anticipate the occurrence of tipping points. In our comparative experiment, we employ linear regression or fit a quadratic curve for extrapolation, and then we select the estimate with the better performance.
	 
	 It is worth noting that the required information of detecting early warning signals with degenerate fingerprinting is different from that with our DL algorithm. The degenerate fingerprinting requires information from all variables of the study system to approximate the critical mode, whereas our DL algorithm only requires data from one variable of the study system. Similarly, the DEV also requires data from one variable of the study system. Moreover, the BB method is designed for estimating the lag-1 autoregressive coefficient of one-dimensional system, whereas our DL algorithm is suitable for analyzing multidimensional system. The DEV can also deal with multidimensional system.
     
	 \section{Data availability}
	 All the simulated datasets and the empirical datasets used to test the deep learning algorithm have been deposited in Github (\url{https://github.com/zhugchzo/dl_occurrence_tipping}) and the training set used to train the deep learning algorithm have been deposited in Zenodo (\url{https://zenodo.org/records/13894933}).
	 \section{Code availability}
	 Code and workflow to reproduce the analysis are available at the GitHub (\url{https://github.com/zhugchzo/dl_occurrence_tipping}).
	 \section{Acknowledgments}
	 This work is supported by National Natural Science Foundation of China under Grant 62388101, the National Key Research and Development Program of China under Grant 2022YFF0902800. We thank Didier Sornette, Jianxi Gao for valuable comments on this work.
	 \section{Author contributions}
	 W.C. conceived the project.  All authors designed the project. C.Z. analyzed the empirical data and did the analytical and numerical calculations. All authors analyzed the results. C.Z. and W.C. wrote the manuscript.
	 \section{Competing interests}
	 The authors declare no competing interests.

	 \bibliographystyle{unsrtnat}
     \bibliography{reference.bib}

\begin{thebibliography}{72}
\providecommand{\natexlab}[1]{#1}
\providecommand{\url}[1]{\texttt{#1}}
\expandafter\ifx\csname urlstyle\endcsname\relax
  \providecommand{\doi}[1]{doi: #1}\else
  \providecommand{\doi}{doi: \begingroup \urlstyle{rm}\Url}\fi

\bibitem[McSharry et~al.(2003)McSharry, Smith, and Tarassenko]{RN111}
Patrick~E McSharry, Leonard~A Smith, and Lionel Tarassenko.
\newblock Prediction of epileptic seizures: are nonlinear methods relevant?
\newblock \emph{Nature medicine}, 9\penalty0 (3):\penalty0 241--242, 2003.
\newblock ISSN 1078-8956.

\bibitem[Clark et~al.(2002)Clark, Pisias, Stocker, and Weaver]{RN75}
Peter~U Clark, Nicklas~G Pisias, Thomas~F Stocker, and Andrew~J Weaver.
\newblock The role of the thermohaline circulation in abrupt climate change.
\newblock \emph{Nature}, 415\penalty0 (6874):\penalty0 863--869, 2002.
\newblock ISSN 0028-0836.

\bibitem[Armstrong~McKay et~al.(2022)Armstrong~McKay, Staal, Abrams, Winkelmann, Sakschewski, Loriani, Fetzer, Cornell, Rockström, and Lenton]{RN110}
David~I Armstrong~McKay, Arie Staal, Jesse~F Abrams, Ricarda Winkelmann, Boris Sakschewski, Sina Loriani, Ingo Fetzer, Sarah~E Cornell, Johan Rockström, and Timothy~M Lenton.
\newblock Exceeding 1.5 c global warming could trigger multiple climate tipping points.
\newblock \emph{Science}, 377\penalty0 (6611):\penalty0 eabn7950, 2022.
\newblock ISSN 0036-8075.

\bibitem[Whaley(2000)]{RN118}
Robert~E Whaley.
\newblock The investor fear gauge.
\newblock \emph{Journal of portfolio management}, 26\penalty0 (3):\penalty0 12, 2000.
\newblock ISSN 0095-4918.

\bibitem[Scheffer et~al.(2009)Scheffer, Bascompte, Brock, Brovkin, Carpenter, Dakos, Held, van Nes, Rietkerk, and Sugihara]{RN24}
M.~Scheffer, J.~Bascompte, W.~A. Brock, V.~Brovkin, S.~R. Carpenter, V.~Dakos, H.~Held, E.~H. van Nes, M.~Rietkerk, and G.~Sugihara.
\newblock Early-warning signals for critical transitions.
\newblock \emph{Nature}, 461\penalty0 (7260):\penalty0 53--9, 2009.
\newblock ISSN 1476-4687 (Electronic) 0028-0836 (Linking).

\bibitem[Scheffer et~al.(2012)Scheffer, Carpenter, Lenton, Bascompte, Brock, Dakos, Van~de Koppel, Van~de Leemput, Levin, and Van~Nes]{RN56}
Marten Scheffer, Stephen~R Carpenter, Timothy~M Lenton, Jordi Bascompte, William Brock, Vasilis Dakos, Johan Van~de Koppel, Ingrid~A Van~de Leemput, Simon~A Levin, and Egbert~H Van~Nes.
\newblock Anticipating critical transitions.
\newblock \emph{science}, 338\penalty0 (6105):\penalty0 344--348, 2012.
\newblock ISSN 0036-8075.

\bibitem[Kuznetsov et~al.(1998)Kuznetsov, Kuznetsov, and Kuznetsov]{RN66}
Yuri~A Kuznetsov, Iu~A Kuznetsov, and Y~Kuznetsov.
\newblock \emph{Elements of applied bifurcation theory}, volume 112.
\newblock Springer, 1998.

\bibitem[Wissel(1984)]{RN53}
C~Wissel.
\newblock A universal law of the characteristic return time near thresholds.
\newblock \emph{Oecologia}, 65:\penalty0 101--107, 1984.
\newblock ISSN 0029-8549.

\bibitem[Held and Kleinen(2004)]{RN21}
H.~Held and T.~Kleinen.
\newblock Detection of climate system bifurcations by degenerate fingerprinting.
\newblock \emph{Geophysical Research Letters}, 31\penalty0 (23), 2004.
\newblock ISSN 0094-8276.

\bibitem[Boettner and Boers(2022)]{RN16}
C.~Boettner and N.~Boers.
\newblock Critical slowing down in dynamical systems driven by nonstationary correlated noise.
\newblock \emph{Physical Review Research}, 4\penalty0 (1), 2022.
\newblock ISSN 2643-1564.

\bibitem[Clarke et~al.(2023)Clarke, Huntingford, Ritchie, and Cox]{RN39}
J.~J. Clarke, C.~Huntingford, P.~D.~L. Ritchie, and P.~M. Cox.
\newblock Seeking more robust early warning signals for climate tipping points: the ratio of spectra method (rosa).
\newblock \emph{Environmental Research Letters}, 18\penalty0 (3), 2023.
\newblock ISSN 1748-9326.

\bibitem[Grziwotz et~al.(2023)Grziwotz, Chang, Dakos, van Nes, Schwarzländer, Kamps, Heßler, Tokuda, Telschow, and Hsieh]{RN58}
Florian Grziwotz, Chun-Wei Chang, Vasilis Dakos, Egbert~H van Nes, Markus Schwarzländer, Oliver Kamps, Martin Heßler, Isao~T Tokuda, Arndt Telschow, and Chih-hao Hsieh.
\newblock Anticipating the occurrence and type of critical transitions.
\newblock \emph{Science Advances}, 9\penalty0 (1):\penalty0 eabq4558, 2023.
\newblock ISSN 2375-2548.

\bibitem[Takens(1981)]{RN125}
Floris Takens.
\newblock Dynamical systems and turbulence.
\newblock \emph{Warwick, 1980}, pages 366--381, 1981.

\bibitem[Schulz and Stattegger(1997)]{RN103}
Michael Schulz and Karl Stattegger.
\newblock Spectrum: Spectral analysis of unevenly spaced paleoclimatic time series.
\newblock \emph{Computers \& Geosciences}, 23\penalty0 (9):\penalty0 929--945, 1997.
\newblock ISSN 0098-3004.

\bibitem[Rehfeld et~al.(2011)Rehfeld, Marwan, Heitzig, and Kurths]{RN101}
Kira Rehfeld, Norbert Marwan, Jobst Heitzig, and Jürgen Kurths.
\newblock Comparison of correlation analysis techniques for irregularly sampled time series.
\newblock \emph{Nonlinear Processes in Geophysics}, 18\penalty0 (3):\penalty0 389--404, 2011.
\newblock ISSN 1023-5809.

\bibitem[Piroddi and Petrou(2004)]{RN100}
Roberta Piroddi and Maria Petrou.
\newblock Analysis of irregularly sampled data: A review.
\newblock \emph{Advances in Imaging and Electron Physics}, 132:\penalty0 109--167, 2004.
\newblock ISSN 1076-5670.

\bibitem[Liew et~al.(2007)Liew, Xian, Wu, Smith, and Yan]{RN102}
Alan Wee-Chung Liew, Jun Xian, Shuanhu Wu, David Smith, and Hong Yan.
\newblock Spectral estimation in unevenly sampled space of periodically expressed microarray time series data.
\newblock \emph{BMC bioinformatics}, 8:\penalty0 1--19, 2007.

\bibitem[Bury et~al.(2021)Bury, Sujith, Pavithran, Scheffer, Lenton, Anand, and Bauch]{RN18}
T.~M. Bury, R.~I. Sujith, I.~Pavithran, M.~Scheffer, T.~M. Lenton, M.~Anand, and C.~T. Bauch.
\newblock Deep learning for early warning signals of tipping points.
\newblock \emph{Proc Natl Acad Sci U S A}, 118\penalty0 (39), 2021.
\newblock ISSN 1091-6490 (Electronic) 0027-8424 (Print) 0027-8424 (Linking).

\bibitem[Kuehn(2011)]{RN3}
C.~Kuehn.
\newblock A mathematical framework for critical transitions: Bifurcations, fast-slow systems and stochastic dynamics.
\newblock \emph{Physica D-Nonlinear Phenomena}, 240\penalty0 (12):\penalty0 1020--1035, 2011.
\newblock ISSN 0167-2789.

\bibitem[Pavithran and Sujith(2021)]{RN25}
Induja Pavithran and R.~I. Sujith.
\newblock Effect of rate of change of parameter on early warning signals for critical transitions.
\newblock \emph{Chaos: An Interdisciplinary Journal of Nonlinear Science}, 31\penalty0 (1), 2021.
\newblock ISSN 1054-1500.

\bibitem[Bonciolini et~al.(2018)Bonciolini, Ebi, Boujo, and Noiray]{RN97}
Giacomo Bonciolini, Dominik Ebi, Edouard Boujo, and Nicolas Noiray.
\newblock Experiments and modelling of rate-dependent transition delay in a stochastic subcritical bifurcation.
\newblock \emph{Royal Society Open Science}, 5\penalty0 (3):\penalty0 172078, 2018.

\bibitem[Bury et~al.(2023)Bury, Dylewsky, Bauch, Anand, Glass, Shrier, and Bub]{RN108}
Thomas~M Bury, Daniel Dylewsky, Chris~T Bauch, Madhur Anand, Leon Glass, Alvin Shrier, and Gil Bub.
\newblock Predicting discrete-time bifurcations with deep learning.
\newblock \emph{Nature Communications}, 14\penalty0 (1):\penalty0 6331, 2023.
\newblock ISSN 2041-1723.

\bibitem[Dylewsky et~al.(2023)Dylewsky, Lenton, Scheffer, Bury, Fletcher, Anand, and Bauch]{RN199}
Daniel Dylewsky, Timothy~M Lenton, Marten Scheffer, Thomas~M Bury, Christopher~G Fletcher, Madhur Anand, and Chris~T Bauch.
\newblock Universal early warning signals of phase transitions in climate systems.
\newblock \emph{Journal of the Royal Society Interface}, 20\penalty0 (201):\penalty0 20220562, 2023.
\newblock ISSN 1742-5662.

\bibitem[Deb et~al.(2022)Deb, Sidheekh, Clements, Krishnan, and Dutta]{RN28}
S.~Deb, S.~Sidheekh, C.~F. Clements, N.~C. Krishnan, and P.~S. Dutta.
\newblock Machine learning methods trained on simple models can predict critical transitions in complex natural systems.
\newblock \emph{R Soc Open Sci}, 9\penalty0 (2):\penalty0 211475, 2022.
\newblock ISSN 2054-5703 (Print) 2054-5703 (Electronic) 2054-5703 (Linking).

\bibitem[Kong et~al.(2021)Kong, Fan, Grebogi, and Lai]{RN29}
L.~W. Kong, H.~W. Fan, C.~Grebogi, and Y.~C. Lai.
\newblock Machine learning prediction of critical transition and system collapse.
\newblock \emph{Physical Review Research}, 3\penalty0 (1), 2021.
\newblock ISSN 2643-1564.

\bibitem[Patel et~al.(2021)Patel, Canaday, Girvan, Pomerance, and Ott]{RN200}
Dhruvit Patel, Daniel Canaday, Michelle Girvan, Andrew Pomerance, and Edward Ott.
\newblock Using machine learning to predict statistical properties of non-stationary dynamical processes: System climate, regime transitions, and the effect of stochasticity.
\newblock \emph{Chaos: An Interdisciplinary Journal of Nonlinear Science}, 31\penalty0 (3), 2021.
\newblock ISSN 1054-1500.

\bibitem[Patel and Ott(2023)]{RN201}
Dhruvit Patel and Edward Ott.
\newblock Using machine learning to anticipate tipping points and extrapolate to post-tipping dynamics of non-stationary dynamical systems.
\newblock \emph{Chaos: An Interdisciplinary Journal of Nonlinear Science}, 33\penalty0 (2), 2023.
\newblock ISSN 1054-1500.

\bibitem[Huke and Broomhead(2007)]{RN205}
Jeremy~P Huke and David~S Broomhead.
\newblock Embedding theorems for non-uniformly sampled dynamical systems.
\newblock \emph{Nonlinearity}, 20\penalty0 (9):\penalty0 2205, 2007.
\newblock ISSN 0951-7715.

\bibitem[May(1977)]{RN85}
Robert~M May.
\newblock Thresholds and breakpoints in ecosystems with a multiplicity of stable states.
\newblock \emph{Nature}, 269\penalty0 (5628):\penalty0 471--477, 1977.
\newblock ISSN 0028-0836.

\bibitem[McCann and Yodzis(1994)]{RN86}
Kevin McCann and Peter Yodzis.
\newblock Nonlinear dynamics and population disappearances.
\newblock \emph{The American Naturalist}, 144\penalty0 (5):\penalty0 873--879, 1994.
\newblock ISSN 0003-0147.

\bibitem[Rosenzweig and MacArthur(1963)]{RN84}
Michael~L Rosenzweig and Robert~H MacArthur.
\newblock Graphical representation and stability conditions of predator-prey interactions.
\newblock \emph{The American Naturalist}, 97\penalty0 (895):\penalty0 209--223, 1963.
\newblock ISSN 0003-0147.

\bibitem[Fraedrich(1978)]{RN87}
Klaus Fraedrich.
\newblock Structural and stochastic analysis of a zero-dimensional climate system.
\newblock \emph{Quarterly Journal of the Royal Meteorological Society}, 104\penalty0 (440):\penalty0 461--474, 1978.
\newblock ISSN 0035-9009.

\bibitem[Fraedrich(1979)]{RN88}
Klaus Fraedrich.
\newblock Catastrophes and resilience of a zero-dimensional climate system with ice-albedo and greenhouse feedback.
\newblock \emph{Quarterly Journal of the Royal Meteorological Society}, 105\penalty0 (443):\penalty0 147--167, 1979.
\newblock ISSN 0035-9009.

\bibitem[Maasch and Saltzman(1990)]{RN115}
Kirk~A Maasch and Barry Saltzman.
\newblock A low-order dynamical model of global climatic variability over the full pleistocene.
\newblock \emph{Journal of Geophysical Research: Atmospheres}, 95\penalty0 (D2):\penalty0 1955--1963, 1990.
\newblock ISSN 0148-0227.

\bibitem[Cox(2001)]{RN64}
Peter~M Cox.
\newblock Description of the" triffid" dynamic global vegetation model.
\newblock 2001.

\bibitem[Phillips and Robinson(2007)]{RN105}
AJK Phillips and Peter~A Robinson.
\newblock A quantitative model of sleep-wake dynamics based on the physiology of the brainstem ascending arousal system.
\newblock \emph{Journal of Biological Rhythms}, 22\penalty0 (2):\penalty0 167--179, 2007.
\newblock ISSN 0748-7304.

\bibitem[Zhou et~al.(2019)Zhou, Li, Xie, Ma, and Zhang]{RN107}
Chengyi Zhou, Zhijun Li, Fei Xie, Minglin Ma, and Yi~Zhang.
\newblock Bursting oscillations in sprott b system with multi-frequency slow excitations: two novel “hopf/hopf”-hysteresis-induced bursting and complex amb rhythms.
\newblock \emph{Nonlinear Dynamics}, 97:\penalty0 2799--2811, 2019.
\newblock ISSN 0924-090X.

\bibitem[Veraart et~al.(2011)Veraart, Faassen, Dakos, van Nes, Lurling, and Scheffer]{RN37}
A.~J. Veraart, E.~J. Faassen, V.~Dakos, E.~H. van Nes, M.~Lurling, and M.~Scheffer.
\newblock Recovery rates reflect distance to a tipping point in a living system.
\newblock \emph{Nature}, 481\penalty0 (7381):\penalty0 357--9, 2011.
\newblock ISSN 1476-4687 (Electronic) 0028-0836 (Linking).

\bibitem[Scheffer(2020)]{RN63}
Marten Scheffer.
\newblock \emph{Critical transitions in nature and society}, volume~16.
\newblock Princeton University Press, 2020.
\newblock ISBN 1400833272.

\bibitem[Pimm(1984)]{RN43}
Stuart~L. Pimm.
\newblock The complexity and stability of ecosystems.
\newblock \emph{Nature}, 307\penalty0 (5949):\penalty0 321--326, 1984.
\newblock ISSN 0028-0836 1476-4687.

\bibitem[Van~Nes and Scheffer(2007)]{RN73}
Egbert~H Van~Nes and Marten Scheffer.
\newblock Slow recovery from perturbations as a generic indicator of a nearby catastrophic shift.
\newblock \emph{The American Naturalist}, 169\penalty0 (6):\penalty0 738--747, 2007.
\newblock ISSN 0003-0147.

\bibitem[Dakos et~al.(2008)Dakos, Scheffer, Van~Nes, Brovkin, Petoukhov, and Held]{RN74}
Vasilis Dakos, Marten Scheffer, Egbert~H Van~Nes, Victor Brovkin, Vladimir Petoukhov, and Hermann Held.
\newblock Slowing down as an early warning signal for abrupt climate change.
\newblock \emph{Proceedings of the National Academy of Sciences}, 105\penalty0 (38):\penalty0 14308--14312, 2008.
\newblock ISSN 0027-8424.

\bibitem[Juniper(2011)]{RN91}
Matthew~P Juniper.
\newblock Triggering in the horizontal rijke tube: non-normality, transient growth and bypass transition.
\newblock \emph{Journal of Fluid Mechanics}, 667:\penalty0 272--308, 2011.
\newblock ISSN 1469-7645.

\bibitem[Gopalakrishnan and Sujith(2015)]{RN90}
EA~Gopalakrishnan and RI~Sujith.
\newblock Effect of external noise on the hysteresis characteristics of a thermoacoustic system.
\newblock \emph{Journal of Fluid Mechanics}, 776:\penalty0 334--353, 2015.
\newblock ISSN 0022-1120.

\bibitem[Matveev(2003)]{RN76}
Konstantin~Ivanovich Matveev.
\newblock \emph{Thermoacoustic instabilities in the Rijke tube: experiments and modeling}.
\newblock California Institute of Technology, 2003.
\newblock ISBN 0493982302.

\bibitem[Truong et~al.(2020)Truong, Oudre, and Vayatis]{RN207}
Charles Truong, Laurent Oudre, and Nicolas Vayatis.
\newblock Selective review of offline change point detection methods.
\newblock \emph{Signal Processing}, 167:\penalty0 107299, 2020.
\newblock ISSN 0165-1684.

\bibitem[Bauch et~al.(2016)Bauch, Sigdel, Pharaon, and Anand]{RN77}
Chris~T. Bauch, Ram Sigdel, Joe Pharaon, and Madhur Anand.
\newblock Early warning signals of regime shifts in coupled human–environment systems.
\newblock \emph{Proceedings of the National Academy of Sciences}, 113\penalty0 (51):\penalty0 14560--14567, 2016.

\bibitem[Ushio et~al.(2018)Ushio, Hsieh, Masuda, Deyle, Ye, Chang, Sugihara, and Kondoh]{RN204}
Masayuki Ushio, Chih-hao Hsieh, Reiji Masuda, Ethan~R Deyle, Hao Ye, Chun-Wei Chang, George Sugihara, and Michio Kondoh.
\newblock Fluctuating interaction network and time-varying stability of a natural fish community.
\newblock \emph{Nature}, 554\penalty0 (7692):\penalty0 360--363, 2018.
\newblock ISSN 0028-0836.

\bibitem[Tong et~al.(2023)Tong, Hong, Zhang, Aihara, Chen, Liu, and Chen]{RN197}
Yuyan Tong, Renhao Hong, Ze~Zhang, Kazuyuki Aihara, Pei Chen, Rui Liu, and Luonan Chen.
\newblock Earthquake alerting based on spatial geodetic data by spatiotemporal information transformation learning.
\newblock \emph{Proceedings of the National Academy of Sciences}, 120\penalty0 (37):\penalty0 e2302275120, 2023.
\newblock ISSN 0027-8424.

\bibitem[Aufinger et~al.(2022)Aufinger, Brenner, and Simmel]{RN128}
Lukas Aufinger, Johann Brenner, and Friedrich~C Simmel.
\newblock Complex dynamics in a synchronized cell-free genetic clock.
\newblock \emph{Nature communications}, 13\penalty0 (1):\penalty0 2852, 2022.
\newblock ISSN 2041-1723.

\bibitem[Guevara et~al.(1981)Guevara, Glass, and Shrier]{RN130}
Michael~R Guevara, Leon Glass, and Alvin Shrier.
\newblock Phase locking, period-doubling bifurcations, and irregular dynamics in periodically stimulated cardiac cells.
\newblock \emph{Science}, 214\penalty0 (4527):\penalty0 1350--1353, 1981.
\newblock ISSN 0036-8075.

\bibitem[Stone(1993)]{RN129}
Lewi Stone.
\newblock Period-doubling reversals and chaos in simple ecological models, 1993.

\bibitem[Pathak et~al.(2018)Pathak, Hunt, Girvan, Lu, and Ott]{RN119}
Jaideep Pathak, Brian Hunt, Michelle Girvan, Zhixin Lu, and Edward Ott.
\newblock Model-free prediction of large spatiotemporally chaotic systems from data: A reservoir computing approach.
\newblock \emph{Physical review letters}, 120\penalty0 (2):\penalty0 024102, 2018.

\bibitem[Li et~al.(2024)Li, Zhu, Zhao, Duan, Zhao, Zhang, Ma, Sun, and Lin]{RN120}
Xin Li, Qunxi Zhu, Chengli Zhao, Xiaojun Duan, Bolin Zhao, Xue Zhang, Huanfei Ma, Jie Sun, and Wei Lin.
\newblock Higher-order granger reservoir computing: simultaneously achieving scalable complex structures inference and accurate dynamics prediction.
\newblock \emph{Nature Communications}, 15\penalty0 (1):\penalty0 2506, 2024.
\newblock ISSN 2041-1723.

\bibitem[Hasani et~al.(2021)Hasani, Lechner, Amini, Rus, and Grosu]{RN121}
Ramin Hasani, Mathias Lechner, Alexander Amini, Daniela Rus, and Radu Grosu.
\newblock Liquid time-constant networks.
\newblock In \emph{Proceedings of the AAAI Conference on Artificial Intelligence}, volume~35, pages 7657--7666, 2021.
\newblock ISBN 2374-3468.

\bibitem[Lechner et~al.(2020)Lechner, Hasani, Amini, Henzinger, Rus, and Grosu]{RN122}
Mathias Lechner, Ramin Hasani, Alexander Amini, Thomas~A Henzinger, Daniela Rus, and Radu Grosu.
\newblock Neural circuit policies enabling auditable autonomy.
\newblock \emph{Nature Machine Intelligence}, 2\penalty0 (10):\penalty0 642--652, 2020.
\newblock ISSN 2522-5839.

\bibitem[Hasani et~al.(2022)Hasani, Lechner, Amini, Liebenwein, Ray, Tschaikowski, Teschl, and Rus]{RN123}
Ramin Hasani, Mathias Lechner, Alexander Amini, Lucas Liebenwein, Aaron Ray, Max Tschaikowski, Gerald Teschl, and Daniela Rus.
\newblock Closed-form continuous-time neural networks.
\newblock \emph{Nature Machine Intelligence}, 4\penalty0 (11):\penalty0 992--1003, 2022.
\newblock ISSN 2522-5839.

\bibitem[Rudin(2019)]{RN124}
Cynthia Rudin.
\newblock Stop explaining black box machine learning models for high stakes decisions and use interpretable models instead.
\newblock \emph{Nature machine intelligence}, 1\penalty0 (5):\penalty0 206--215, 2019.
\newblock ISSN 2522-5839.

\bibitem[Virtanen et~al.(2020)Virtanen, Gommers, Oliphant, Haberland, and Scipy]{RN92}
P.~Virtanen, R.~Gommers, T.~E. Oliphant, M.~Haberland, and Contributors Scipy.
\newblock Author correction: Scipy 1.0: fundamental algorithms for scientific computing in python (nature methods, (2020), 17, 3, (261-272), 10.1038/s41592-019-0686-2).
\newblock \emph{Nature methods}, 2020.

\bibitem[Doedel et~al.(2007)Doedel, Fairgrieve, Sandstede, Champneys, Kuznetsov, and Wang]{RN93}
Eusebius~J. Doedel, Thomas~F. Fairgrieve, Björn Sandstede, Alan~R. Champneys, Yuri~A. Kuznetsov, and Xianjun Wang.
\newblock Auto-07p: Continuation and bifurcation software for ordinary differential equations.
\newblock \emph{US}, 2007.

\bibitem[Kloeden et~al.(1992)Kloeden, Platen, Kloeden, and Platen]{RN98}
Peter~E Kloeden, Eckhard Platen, Peter~E Kloeden, and Eckhard Platen.
\newblock \emph{Stochastic differential equations}.
\newblock Springer, 1992.
\newblock ISBN 364208107X.

\bibitem[Boyce and DiPrima(2020)]{RN99}
William~E Boyce and Richard~C DiPrima.
\newblock \emph{Elementary differential equations and boundary value problems}.
\newblock Wiley, 2020.
\newblock ISBN 0470458313.

\bibitem[Horsthemke(2012)]{RN67}
Werner Horsthemke.
\newblock Noise induced transitions.
\newblock In \emph{Non-Equilibrium Dynamics in Chemical Systems: Proceedings of the International Symposium, Bordeaux, France, September 3-7, 1984}, pages 150--160. Springer, 2012.

\bibitem[Wang et~al.(2012)Wang, Dearing, Langdon, Zhang, Yang, Dakos, and Scheffer]{RN27}
R.~Wang, J.~A. Dearing, P.~G. Langdon, E.~Zhang, X.~Yang, V.~Dakos, and M.~Scheffer.
\newblock Flickering gives early warning signals of a critical transition to a eutrophic lake state.
\newblock \emph{Nature}, 492\penalty0 (7429):\penalty0 419--22, 2012.

\bibitem[Zhao et~al.(2019)Zhao, Mao, and Chen]{RN42}
J.~F. Zhao, X.~Mao, and L.~J. Chen.
\newblock Speech emotion recognition using deep 1d \& 2d cnn lstm networks.
\newblock \emph{Biomedical Signal Processing and Control}, 47:\penalty0 312--323, 2019.
\newblock ISSN 1746-8094.

\bibitem[Wang et~al.(2023)Wang, Cheng, Tian, and Gao]{RN1}
J.~L. Wang, S.~W. Cheng, J.~M. Tian, and Y.~F. Gao.
\newblock A 2d cnn-lstm hybrid algorithm using time series segments of eeg data for motor imagery classification.
\newblock \emph{Biomedical Signal Processing and Control}, 83, 2023.
\newblock ISSN 1746-8094.

\bibitem[Dablander and Bury(2022)]{RN19}
F.~Dablander and T.~M. Bury.
\newblock Deep learning for tipping points: Preprocessing matters.
\newblock \emph{Proc Natl Acad Sci U S A}, 119\penalty0 (37):\penalty0 e2207720119, 2022.
\newblock ISSN 1091-6490 (Electronic) 0027-8424 (Print) 0027-8424 (Linking).

\bibitem[Bury(2023)]{RN206}
Thomas~M Bury.
\newblock ewstools: a python package for early warning signals of bifurcations in time series data.
\newblock \emph{Journal of Open Source Software}, 8\penalty0 (82):\penalty0 5038, 2023.
\newblock ISSN 2475-9066.

\bibitem[Boulton et~al.(2013)Boulton, Good, and Lenton]{RN23}
C.~A. Boulton, P.~Good, and T.~M. Lenton.
\newblock Early warning signals of simulated amazon rainforest dieback.
\newblock \emph{Theoretical Ecology}, 6\penalty0 (3):\penalty0 373--384, 2013.
\newblock ISSN 1874-1738.

\bibitem[Jimenez-Munoz et~al.(2016)Jimenez-Munoz, Mattar, Barichivich, Santamaria-Artigas, Takahashi, Malhi, Sobrino, and Schrier]{RN36}
J.~C. Jimenez-Munoz, C.~Mattar, J.~Barichivich, A.~Santamaria-Artigas, K.~Takahashi, Y.~Malhi, J.~A. Sobrino, and Gv~Schrier.
\newblock Record-breaking warming and extreme drought in the amazon rainforest during the course of el nino 2015-2016.
\newblock \emph{Sci Rep}, 6:\penalty0 33130, 2016.
\newblock ISSN 2045-2322 (Electronic) 2045-2322 (Linking).

\bibitem[Sprott(1994)]{RN106}
J~Clint Sprott.
\newblock Some simple chaotic flows.
\newblock \emph{Physical review E}, 50\penalty0 (2):\penalty0 R647, 1994.

\bibitem[Boers(2021)]{RN33}
Niklas Boers.
\newblock Observation-based early-warning signals for a collapse of the atlantic meridional overturning circulation.
\newblock \emph{Nature Climate Change}, 11\penalty0 (8):\penalty0 680--688, 2021.
\newblock ISSN 1758-678X 1758-6798.

\end{thebibliography}


\begin{thebibliography}{9}
\providecommand{\natexlab}[1]{#1}
\providecommand{\url}[1]{\texttt{#1}}
\expandafter\ifx\csname urlstyle\endcsname\relax
  \providecommand{\doi}[1]{doi: #1}\else
  \providecommand{\doi}{doi: \begingroup \urlstyle{rm}\Url}\fi

\bibitem[Bury et~al.(2021)Bury, Sujith, Pavithran, Scheffer, Lenton, Anand, and Bauch]{RN18}
T.~M. Bury, R.~I. Sujith, I.~Pavithran, M.~Scheffer, T.~M. Lenton, M.~Anand, and C.~T. Bauch.
\newblock Deep learning for early warning signals of tipping points.
\newblock \emph{Proc Natl Acad Sci U S A}, 118\penalty0 (39), 2021.
\newblock ISSN 1091-6490 (Electronic) 0027-8424 (Print) 0027-8424 (Linking).

\bibitem[Huke and Broomhead(2007)]{RN205}
Jeremy~P Huke and David~S Broomhead.
\newblock Embedding theorems for non-uniformly sampled dynamical systems.
\newblock \emph{Nonlinearity}, 20\penalty0 (9):\penalty0 2205, 2007.
\newblock ISSN 0951-7715.

\bibitem[Strogatz(2018)]{RN203}
Steven~H Strogatz.
\newblock \emph{Nonlinear dynamics and chaos: with applications to physics, biology, chemistry, and engineering}.
\newblock CRC press, 2018.
\newblock ISBN 0429492561.

\bibitem[Held and Kleinen(2004)]{RN21}
H.~Held and T.~Kleinen.
\newblock Detection of climate system bifurcations by degenerate fingerprinting.
\newblock \emph{Geophysical Research Letters}, 31\penalty0 (23), 2004.
\newblock ISSN 0094-8276.

\bibitem[Boettner and Boers(2022)]{RN16}
C.~Boettner and N.~Boers.
\newblock Critical slowing down in dynamical systems driven by nonstationary correlated noise.
\newblock \emph{Physical Review Research}, 4\penalty0 (1), 2022.
\newblock ISSN 2643-1564.

\bibitem[Grziwotz et~al.(2023)Grziwotz, Chang, Dakos, van Nes, Schwarzländer, Kamps, Heßler, Tokuda, Telschow, and Hsieh]{RN58}
Florian Grziwotz, Chun-Wei Chang, Vasilis Dakos, Egbert~H van Nes, Markus Schwarzländer, Oliver Kamps, Martin Heßler, Isao~T Tokuda, Arndt Telschow, and Chih-hao Hsieh.
\newblock Anticipating the occurrence and type of critical transitions.
\newblock \emph{Science Advances}, 9\penalty0 (1):\penalty0 eabq4558, 2023.
\newblock ISSN 2375-2548.

\bibitem[Kuznetsov et~al.(1998)Kuznetsov, Kuznetsov, and Kuznetsov]{RN66}
Yuri~A Kuznetsov, Iu~A Kuznetsov, and Y~Kuznetsov.
\newblock \emph{Elements of applied bifurcation theory}, volume 112.
\newblock Springer, 1998.

\bibitem[Bury et~al.(2023)Bury, Dylewsky, Bauch, Anand, Glass, Shrier, and Bub]{RN108}
Thomas~M Bury, Daniel Dylewsky, Chris~T Bauch, Madhur Anand, Leon Glass, Alvin Shrier, and Gil Bub.
\newblock Predicting discrete-time bifurcations with deep learning.
\newblock \emph{Nature Communications}, 14\penalty0 (1):\penalty0 6331, 2023.
\newblock ISSN 2041-1723.

\bibitem[Deb et~al.(2022)Deb, Sidheekh, Clements, Krishnan, and Dutta]{RN28}
S.~Deb, S.~Sidheekh, C.~F. Clements, N.~C. Krishnan, and P.~S. Dutta.
\newblock Machine learning methods trained on simple models can predict critical transitions in complex natural systems.
\newblock \emph{R Soc Open Sci}, 9\penalty0 (2):\penalty0 211475, 2022.
\newblock ISSN 2054-5703 (Print) 2054-5703 (Electronic) 2054-5703 (Linking).

\end{thebibliography}
	
\end{document}


\maketitle
	\tableofcontents
	\fontsize{11.2pt}{15.85pt}\selectfont
	
	\clearpage
	\section{Supplementary Figures}
	\vspace{0.5cm}
    \begin{figure}[htbp]
	\centering
	\includegraphics[width=1\linewidth]{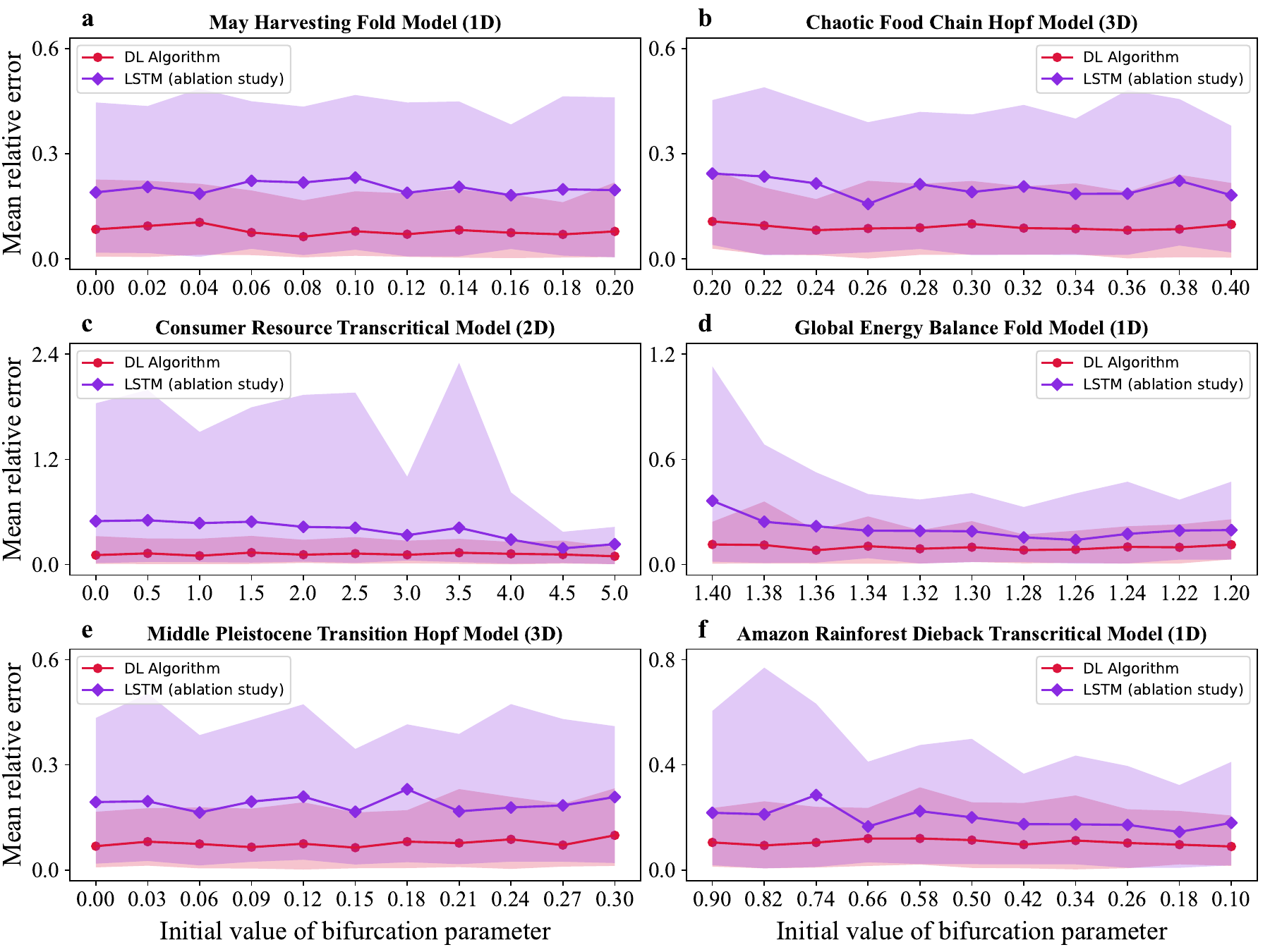}
	\caption*{\textbf{Figure S1.}\fontsize{10pt}{12pt}\selectfont{
			\textbf{The mean relative error of tipping points prediction between the DL algorithm and LSTM on regularly-sampled model time series in the ablation study.} The horizontal axis represents the initial values of the bifurcation parameter, and the vertical axis represents the mean relative error of prediction. The area covered by the polyline represents the 90\% confidence interval for the relative error of tipping points prediction. (a-c) Three ecological model time series with white noise, which undergo fold, Hopf, and transcritical bifurcation, respectively. (d-f) Three climate model time series with red noise, which undergo fold, Hopf, and transcritical bifurcation, respectively.}}
    \end{figure}
    
    \clearpage
    \begin{figure}[htbp]
    \centering
    \includegraphics[width=1\linewidth]{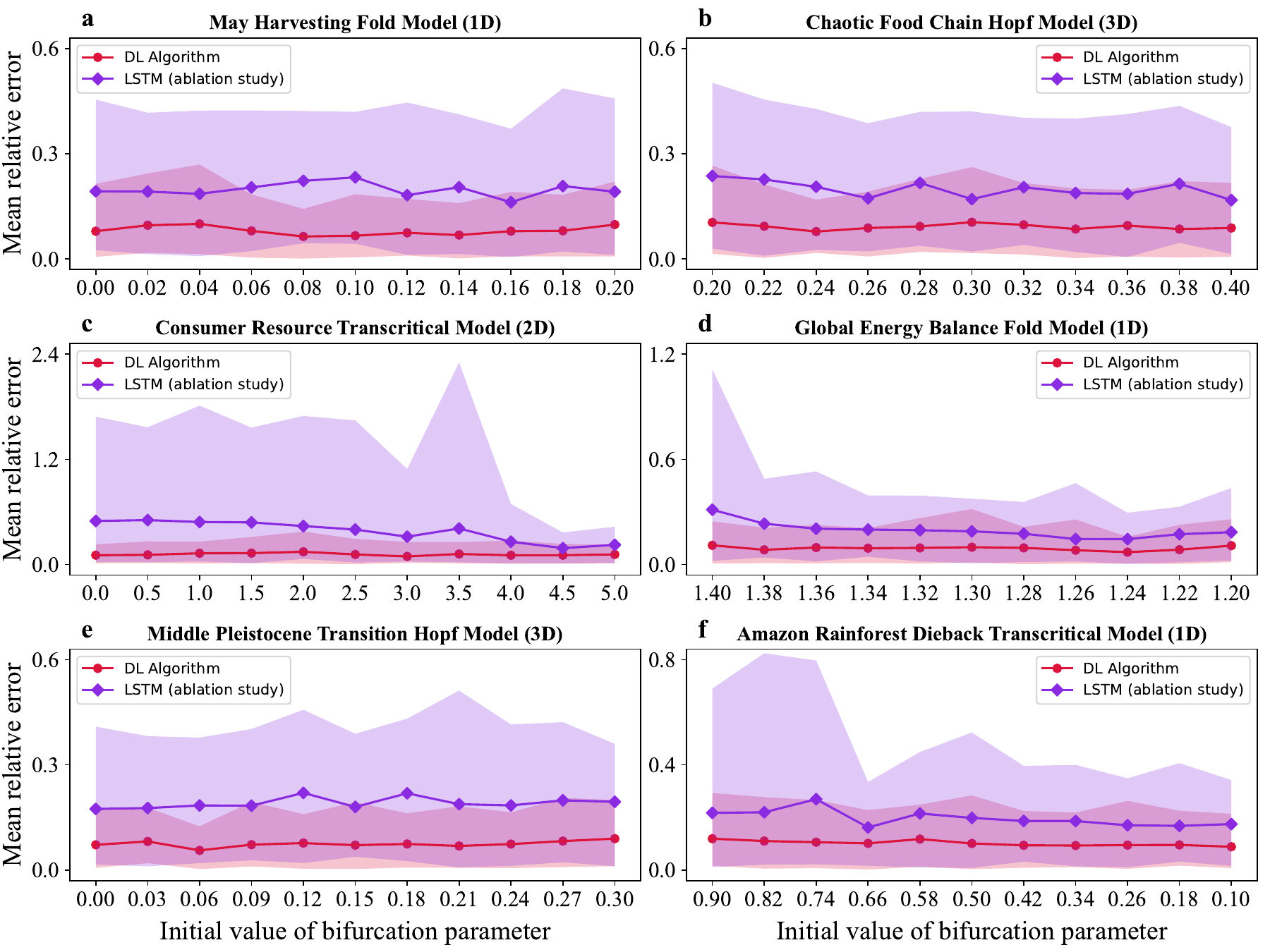}
    \caption*{\textbf{Figure S2.}\fontsize{10pt}{12pt}\selectfont{
    			\textbf{The mean relative error of tipping points prediction between the DL algorithm and LSTM on irregularly-sampled model time series in the ablation study.} The horizontal axis represents the initial values of the bifurcation parameter, and the vertical axis represents the mean relative error of prediction. The area covered by the polyline represents the 90\% confidence interval for the relative error of tipping points prediction. (a-c) Three ecological model time series with white noise, which undergo fold, Hopf, and transcritical bifurcation, respectively. (d-f) Three climate model time series with red noise, which undergo fold, Hopf, and transcritical bifurcation, respectively.}}
    \end{figure}
    
    \clearpage
    \begin{figure}[htbp]
	\centering
	\includegraphics[width=1\linewidth]{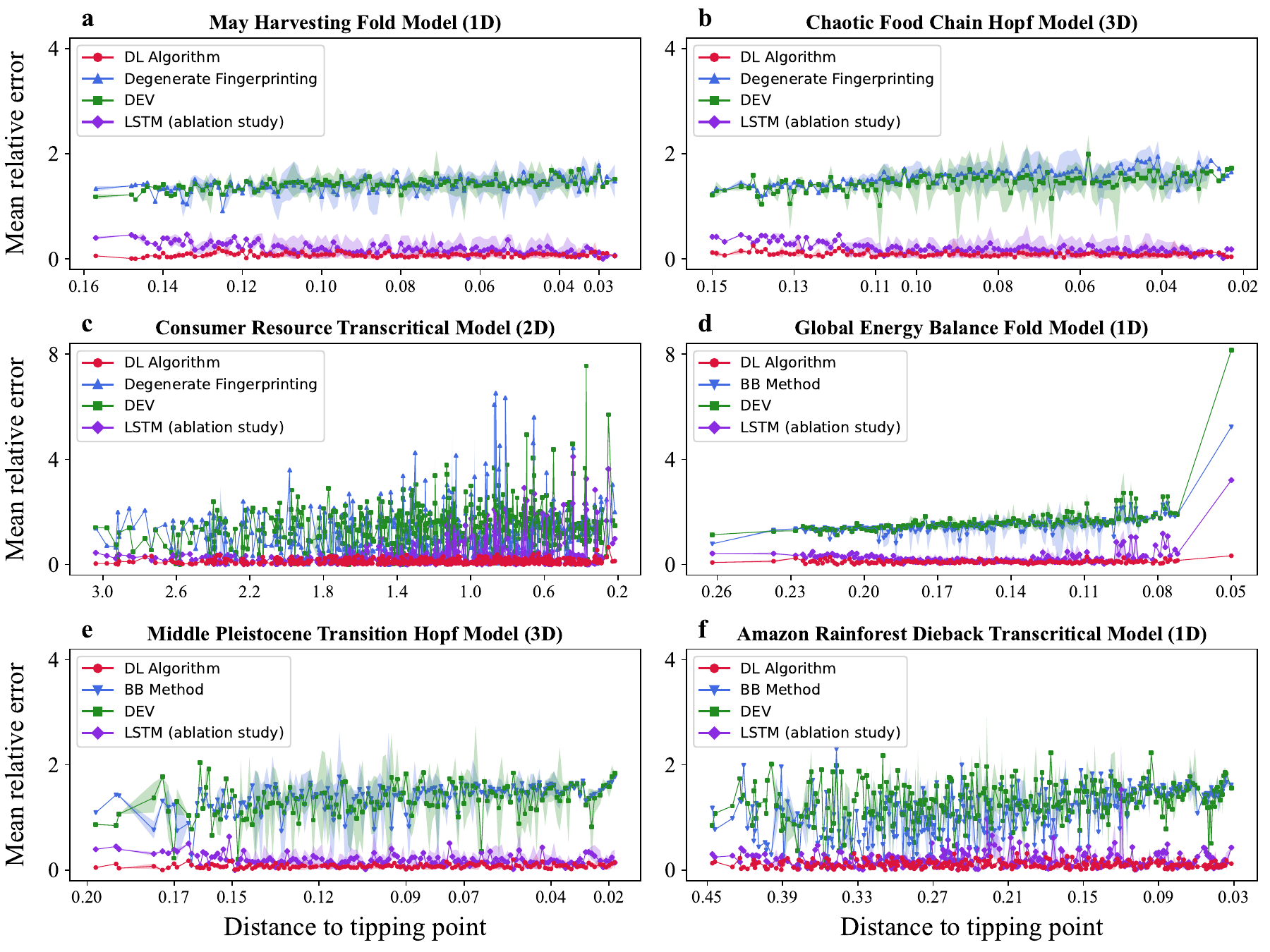}
	\caption*{\textbf{Figure S3.}\fontsize{10pt}{12pt}\selectfont{
			\textbf{The mean relative error of tipping points prediction between the DL algorithm and competing algorithms on regularly-sampled model time series.} The horizontal axis represents the distance between the final value of bifurcation parameter time series and the value of the tipping point, and the vertical axis represents the mean relative error of prediction. The area covered by the polyline represents the 90\% confidence interval for the relative error of tipping points prediction. (a-c) We compared the DL algorithm (red lines) with degenerate fingerprinting (blue lines), DEV (green lines) and LSTM (purple lines) on three ecological model time series with white noise. These model time series undergo fold, Hopf, and transcritical bifurcation, respectively. (d-f) The DL algorithm (red lines) is compared with BB method (blue lines), DEV (green lines) and LSTM (purple lines) on three climate model time series with red noise. These model time series undergo fold, Hopf, and transcritical bifurcation, respectively.}}
    \end{figure}
    
    \clearpage
    \begin{figure}[htbp]
	\centering
	\includegraphics[width=1\linewidth]{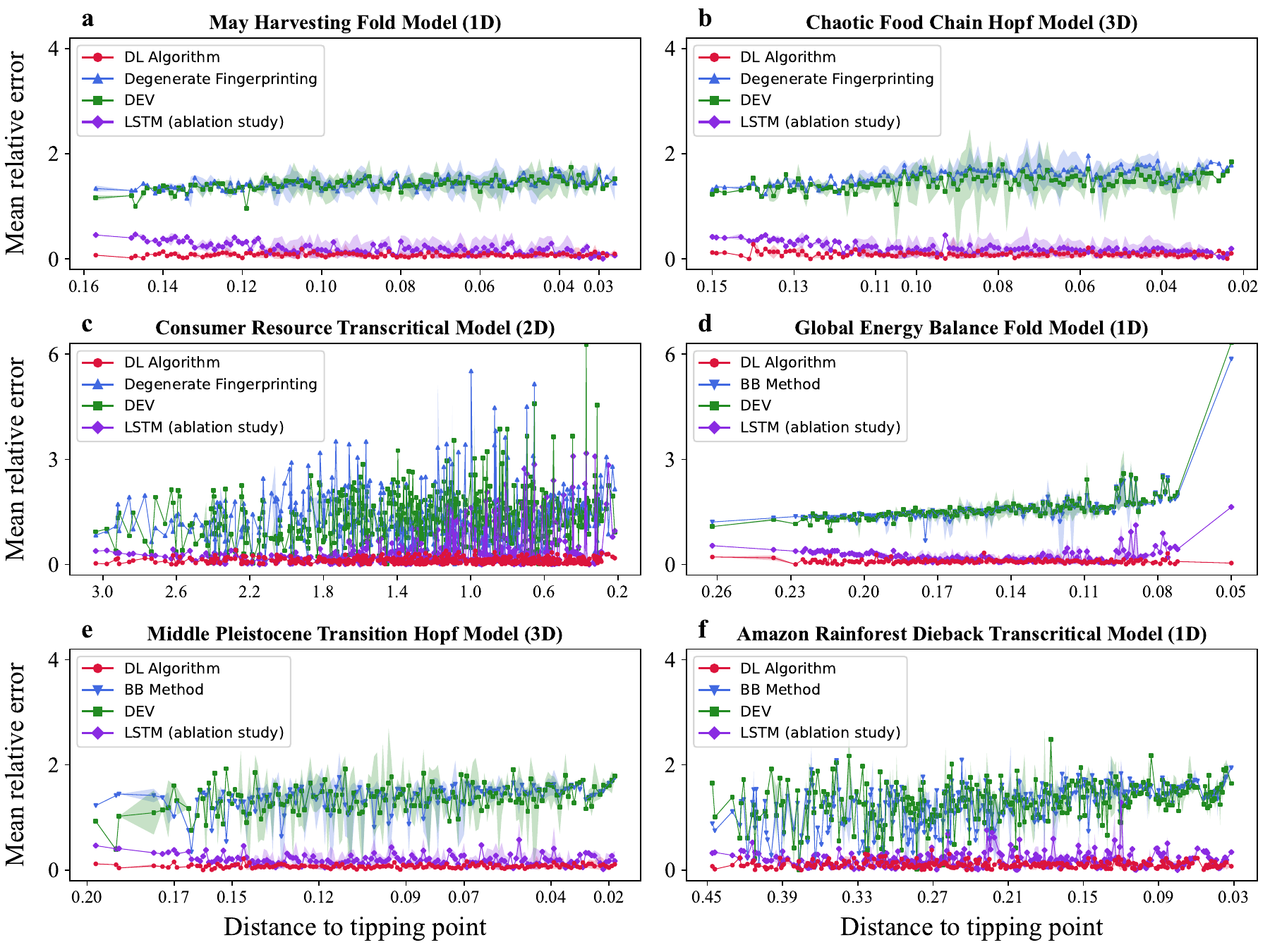}
	\caption*{\textbf{Figure S4.}\fontsize{10pt}{12pt}\selectfont{
			\textbf{The mean relative error of tipping points prediction between the DL algorithm and competing algorithms on irregularly-sampled model time series.} The horizontal axis represents the distance between the final value of bifurcation parameter time series and the value of the tipping point, and the vertical axis represents the mean relative error of prediction. The area covered by the polyline represents the 90\% confidence interval for the relative error of tipping points prediction. (a-c) We compared the DL algorithm (red lines) with degenerate fingerprinting (blue lines), DEV (green lines) and LSTM (purple lines) on three ecological model time series with white noise. These model time series undergo fold, Hopf, and transcritical bifurcation, respectively. (d-f) The DL algorithm (red lines) is compared with BB method (blue lines), DEV (green lines) and LSTM (purple lines) on three climate model time series with red noise. These model time series undergo fold, Hopf, and transcritical bifurcation, respectively.}}
    \end{figure}
    
    \clearpage
    \begin{figure}[htbp]
	\centering
	\includegraphics[width=1\linewidth]{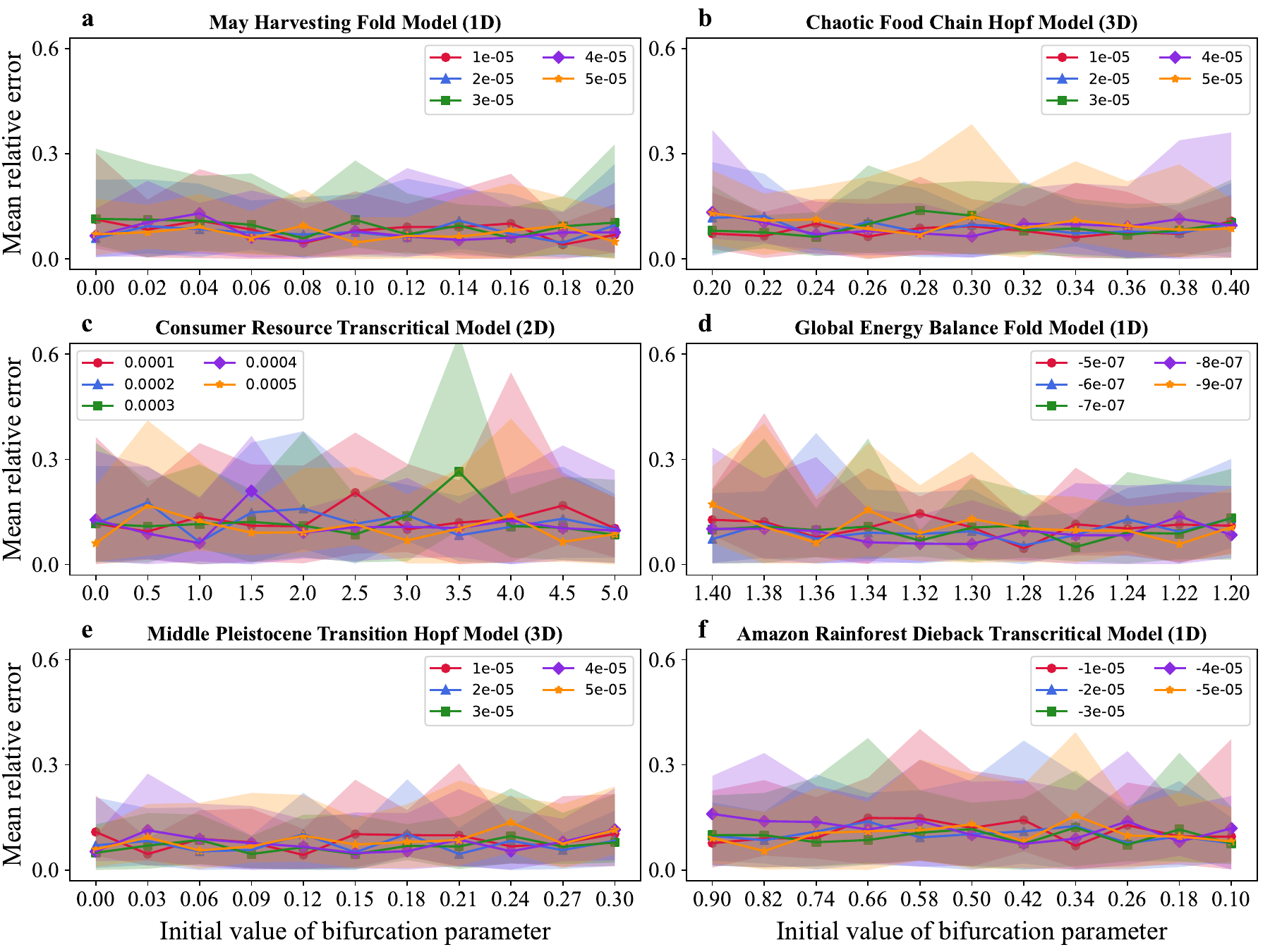}
	\caption*{\textbf{Figure S5.}\fontsize{10pt}{12pt}\selectfont{
			\textbf{The mean relative error of predicted tipping points by the DL algorithm on regularly-sampled model time series of different changing rates of bifurcation parameter.} The horizontal axis represents the initial values of the bifurcation parameter, and the vertical axis represents the mean relative error of prediction. The area covered by the polyline represents the 90\% confidence interval for the relative error of tipping points prediction. (a-c) Three ecological model time series with white noise, which undergo fold, Hopf, and transcritical bifurcation, respectively. (d-f) Three climate model time series with red noise, which undergo fold, Hopf, and transcritical bifurcation, respectively.}}
    \end{figure}
    
    \clearpage
    \begin{figure}[htbp]
	\centering
	\includegraphics[width=1\linewidth]{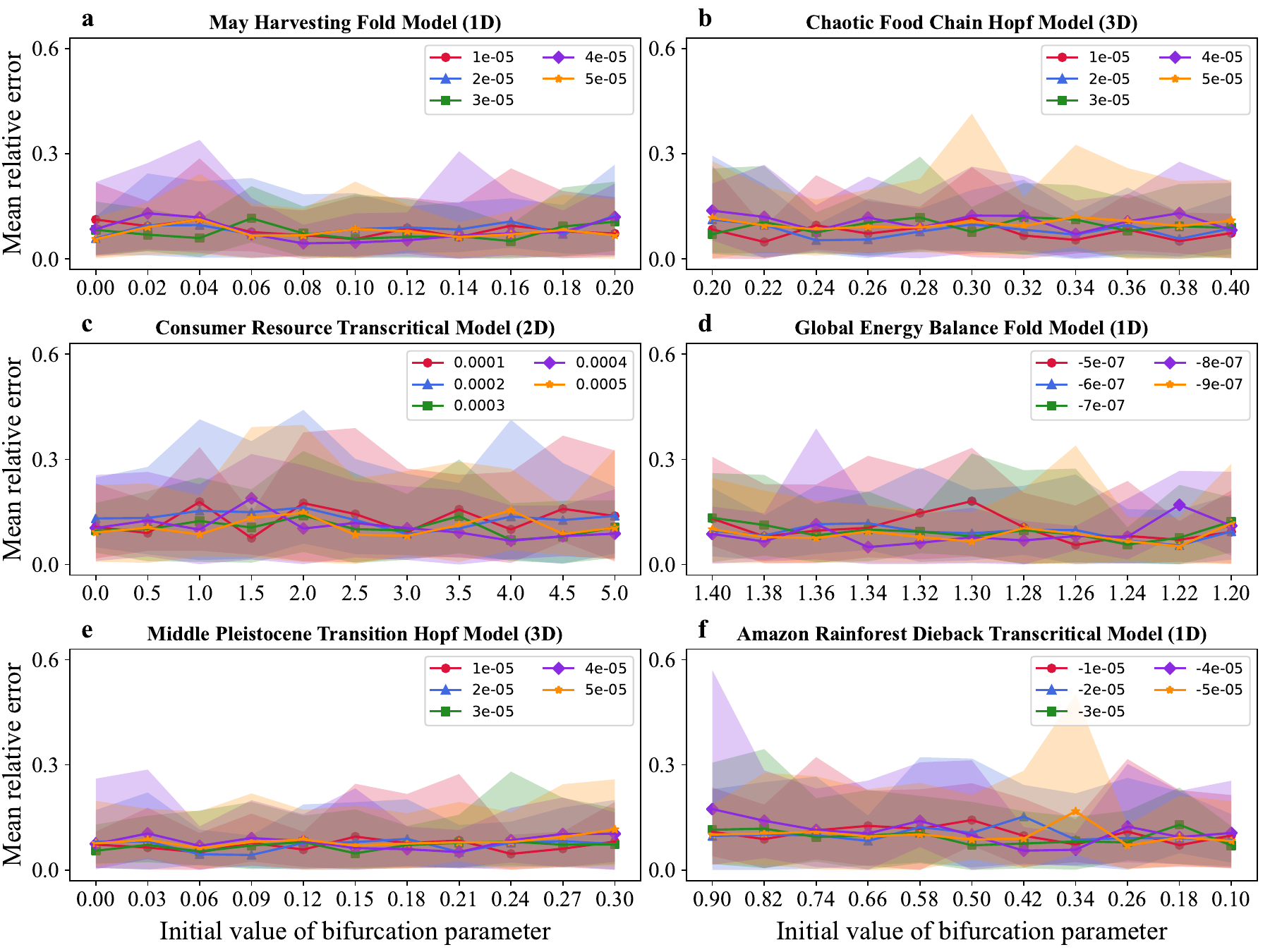}
	\caption*{\textbf{Figure S6.}\fontsize{10pt}{12pt}\selectfont{
			\textbf{The mean relative error of predicted tipping points by the DL algorithm on irregularly-sampled model time series of different changing rates of bifurcation parameter.} The horizontal axis represents the initial values of the bifurcation parameter, and the vertical axis represents the mean relative error of prediction. The area covered by the polyline represents the 90\% confidence interval for the relative error of tipping points prediction. (a-c) Three ecological model time series with white noise, which undergo fold, Hopf, and transcritical bifurcation, respectively. (d-f) Three climate model time series with red noise, which undergo fold, Hopf, and transcritical bifurcation, respectively.}}
    \end{figure}
    
    \clearpage
    \begin{figure}[htbp]
	\centering
	\includegraphics[width=1.0\linewidth]{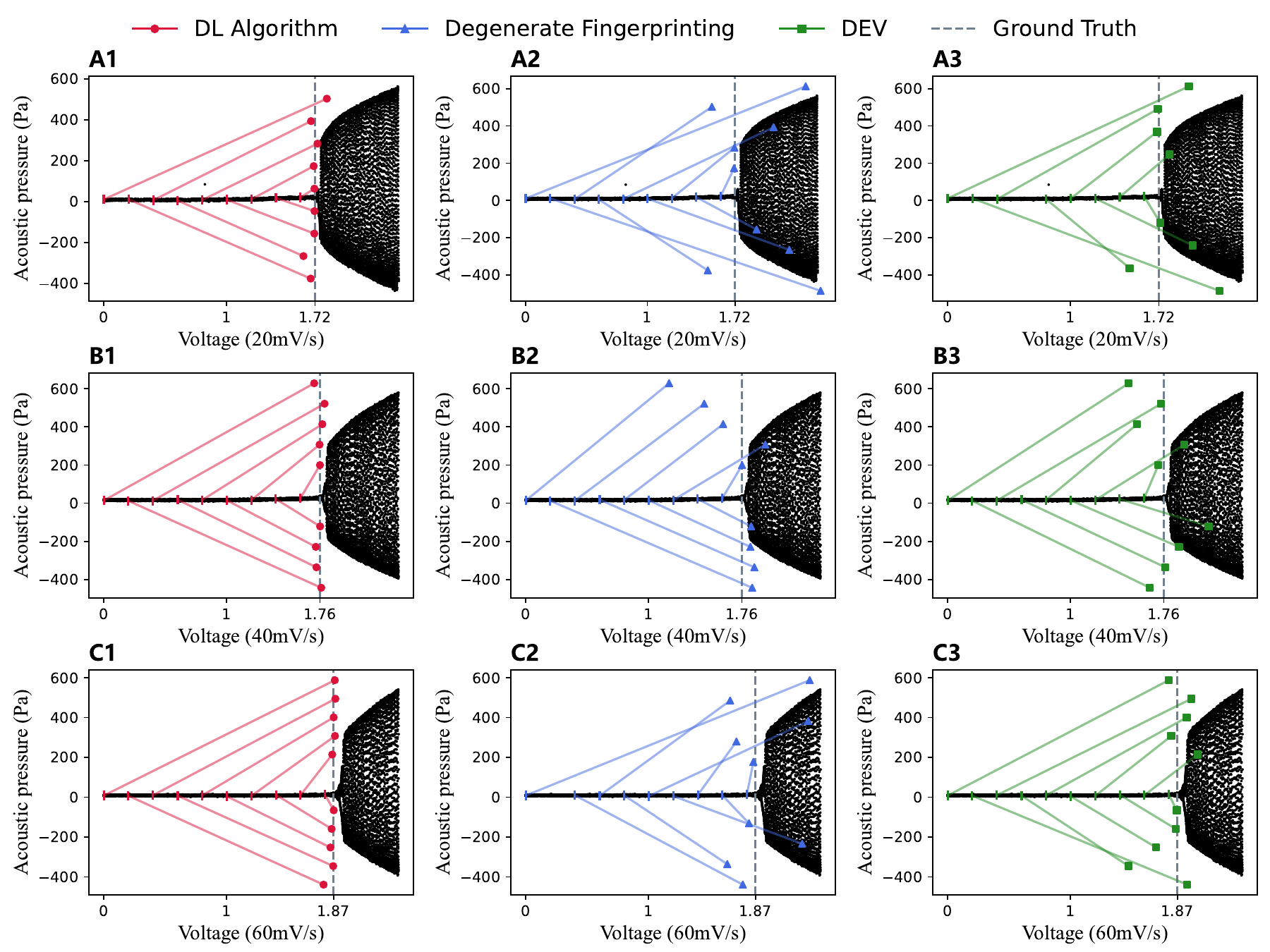}
	\caption*{\textbf{Figure S7.}\fontsize{10pt}{12pt}\selectfont {
			\textbf{The performance of DL algorithm in predicting tipping points on irregularly-sampled thermoacoustic time series under different changing rates of the voltage.}  Different rows of this figure (A1-A3, B1-B3, C1-C3) represent thermoacoustic systems under different changing rates of the voltage (20mV/s, 40mV/s, 60mV/s). The dots denote the DL predictions, and the short vertical lines denote the initial points of the time series data used for prediction. We connect them with lines. We compared the performance of the DL algorithm (red) with degenerate fingerprinting (blue) and DEV (green). We used linear interpolation to transform these irregularly-sampled time series into equidistant data so that they are suitable for degenerate fingerprinting and DEV. In A3 and C2, there is an extreme outlier identified by each competing algorithm. Here we have excluded these two predictions from the figure.}}
    \end{figure}
    
    \clearpage
    \begin{figure}[htbp]
    \centering
    \includegraphics[width=1\linewidth]{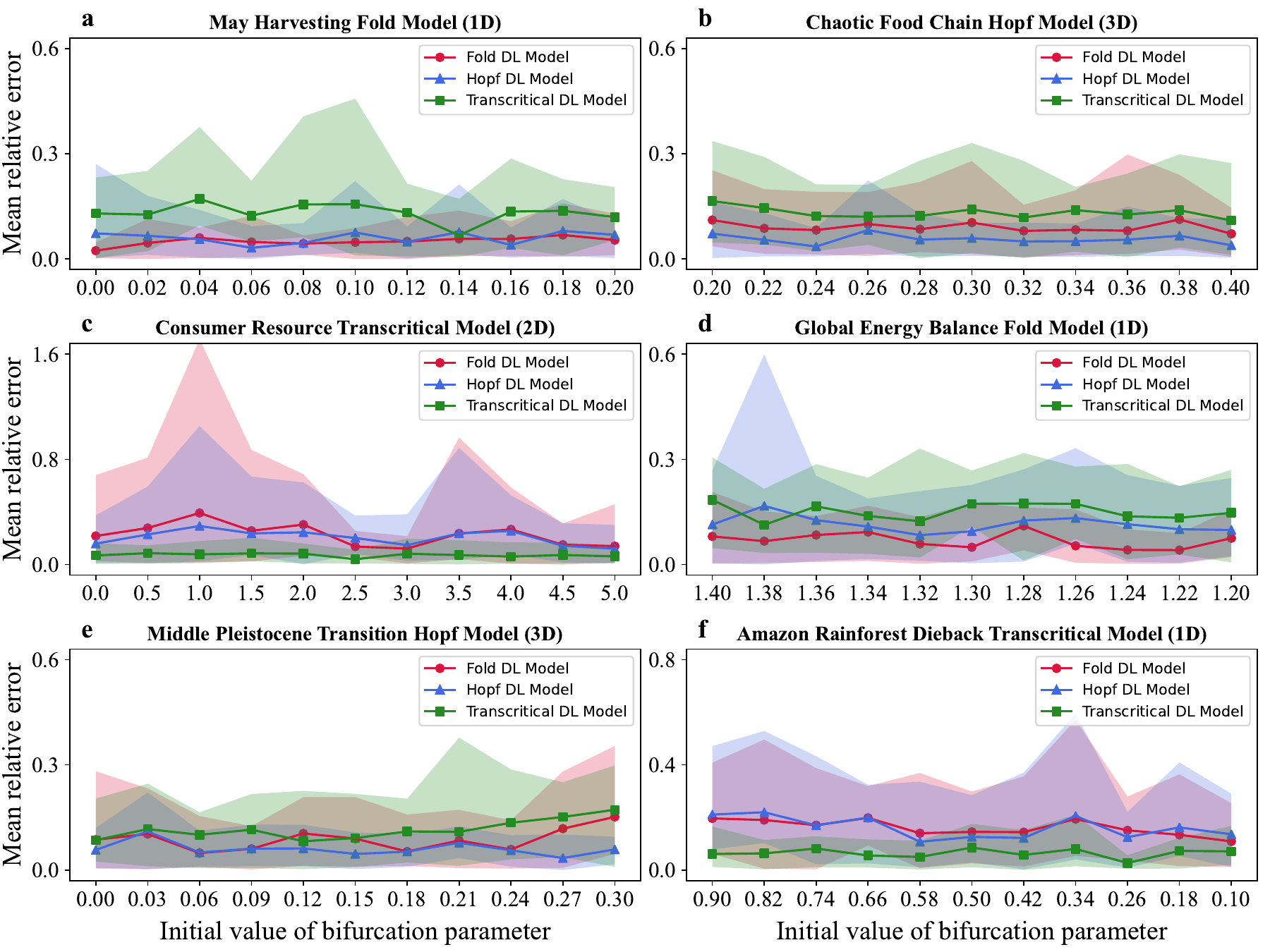}
    \caption*{\textbf{Figure S8.}\fontsize{10pt}{12pt}\selectfont{
    	\textbf{The mean relative error of tipping points prediction among three DL models on irregularly-sampled model time series in the first control experiment studied in the main manuscript.} The horizontal axis represents the initial values of the bifurcation parameter, and the vertical axis represents the mean relative error of prediction. The red, blue and green lines represent the DL model trained on the dataset consisting solely of time series with fold, Hopf, or transcritical bifurcation, respectively. The area covered by the polyline represents the 90\% confidence interval for the relative error of tipping points prediction. (a-c) Three ecological model time series with white noise, which undergo fold, Hopf, and transcritical bifurcation, respectively. (d-f) Three climate model time series with red noise, which undergo fold, Hopf, and transcritical bifurcation, respectively.}}
    \end{figure}
    
    \clearpage
    \begin{figure}[htbp]
	\centering
	\includegraphics[width=1\linewidth]{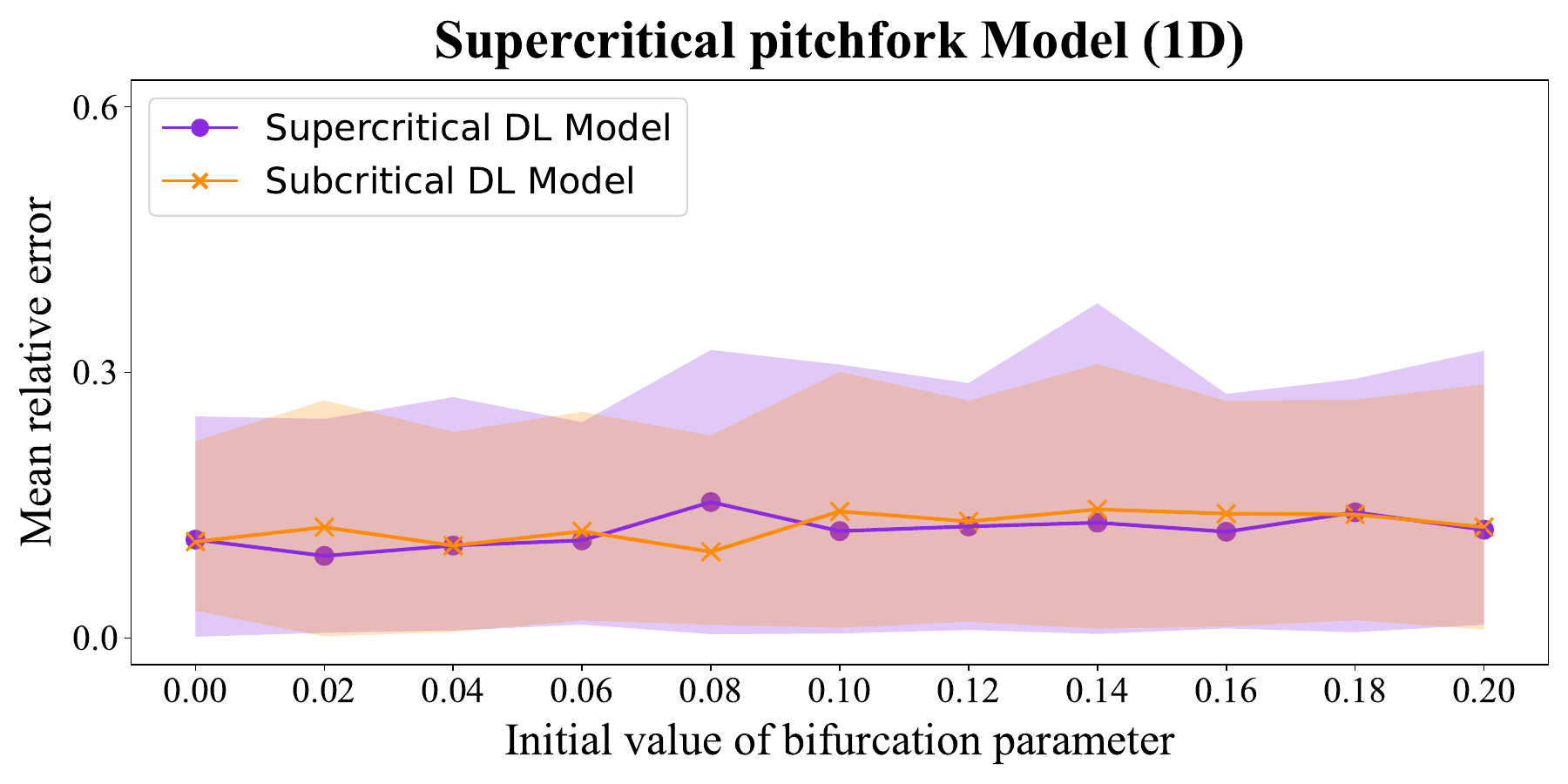}
	\caption*{\textbf{Figure S9.}\fontsize{10pt}{12pt}\selectfont{
		\textbf{The mean relative error of tipping points prediction between two DL models on irregularly-sampled model time series in the second control experiment studied in the main manuscript.} The horizontal axis represents the initial values of the bifurcation parameter, and the vertical axis represents the mean relative error of prediction. The purple and orange lines represent the DL model trained on the dataset consisting solely of time series with supercritical pirchfork and subcritical pirchfork bifurcation, respectively. The area covered by the polyline represents the 90\% confidence interval for the relative error of tipping points prediction. We tested these two DL models on irregularly-sampled model time series with supercritical pirchfork bifurcation.}}
    \end{figure}
    
	\clearpage
    \begin{figure}[htbp]
	\centering
	\includegraphics[width=1\linewidth]{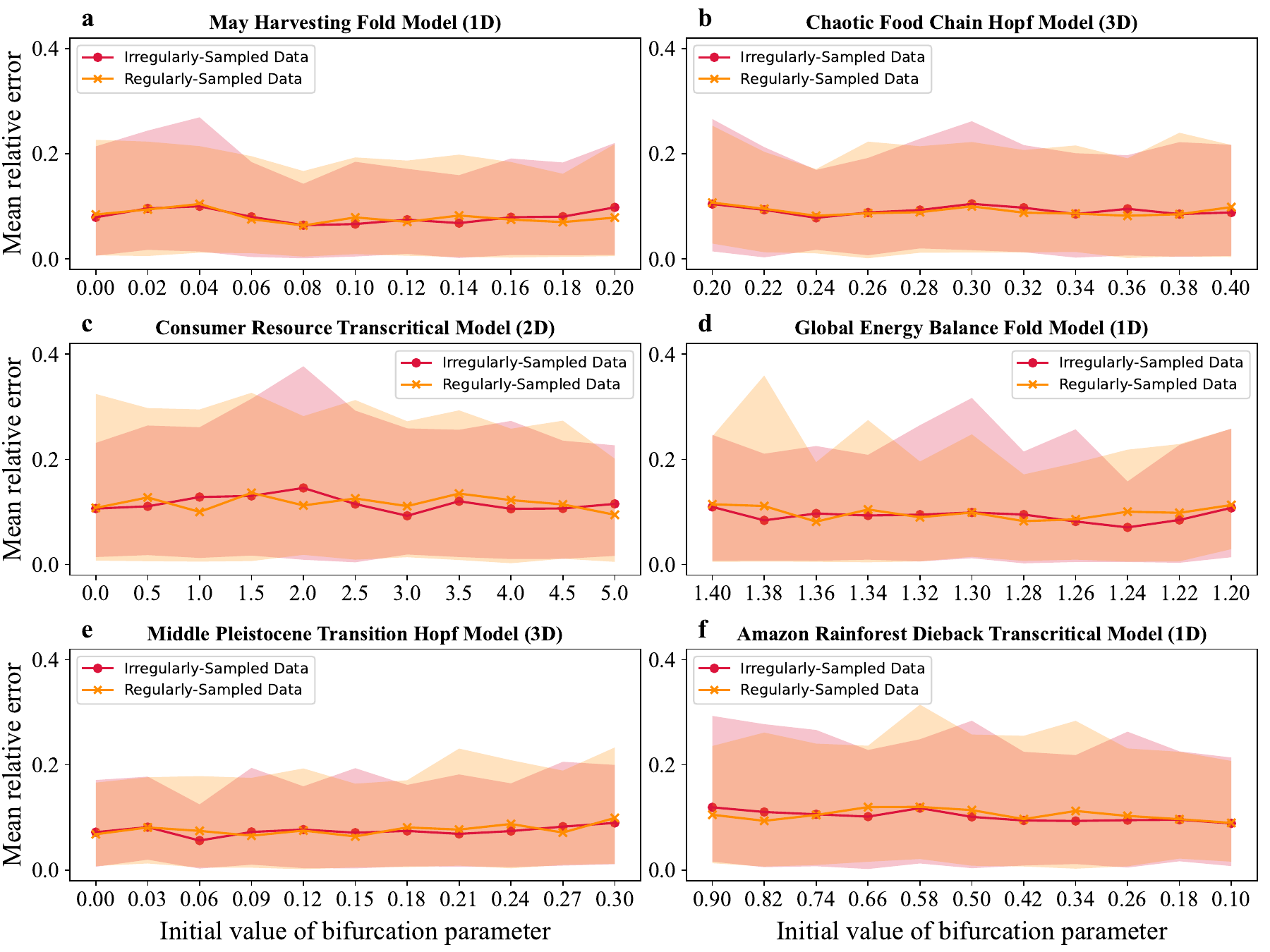}
	\caption*{\textbf{Figure S10.}\fontsize{10pt}{12pt}\selectfont{
			\textbf{The performance of the DL algorithm on irregularly-sampled (red lines) and regularly-sampled model time series (orange lines), as illustrated in Fig. 3 and Fig. 4 of the main manuscript respectively.} The horizontal axis represents the initial values of the bifurcation parameter, and the vertical axis represents the mean relative error of prediction. The area covered by the polyline represents the 90\% confidence interval for the relative error of tipping points prediction. (a-c) Three ecological model time series with white noise, which undergo fold, Hopf, and transcritical bifurcation, respectively. (d-f) Three climate model time series with red noise, which undergo fold, Hopf, and transcritical bifurcation, respectively.}}
    \end{figure}
    
    \clearpage
    \section{Supplementary Notes}
    \vspace{0.5cm}
    \subsection{Supplementary Note 1. Critical slowing down}
    The local behavior of a dynamical system about an equilibrium can be well characterized by a linear approximation of the equations that dominate its dynamics. We illustrate this for a one-dimensional system $dx/dt = f(x)$ with equilibrium $x^*$, which means $f(x^*)=0$. The dynamics about equilibrium following a perturbation by $\varepsilon$ satisfy
    \begin{equation}
    	\begin{aligned}
    		\frac{d(x^*+\varepsilon)}{dt}=f(x^*+\varepsilon)&=f(x^*)+\frac{\partial f}{\partial x}\Big |_{x=x^*}\varepsilon+\frac{1}{2}\frac{\partial^2 f}{\partial x^2}\Big |_{x=x^*}\varepsilon^2+\dots \\
    		&=\lambda_1\varepsilon+\lambda_2\varepsilon^2+\dots, 
    	\end{aligned}
    	\tag{S1}
    \end{equation}
    where $\lambda_1,\lambda_2,\dots$ are coefficients of the Taylor expansion, and $\lambda_1$ is the dominant eigenvalue. The potential landscape of this system centered on $x^*$ is given by
    \begin{equation}
    	V(\varepsilon)=-\int f(x^*+\varepsilon)d\varepsilon=-\frac{1}{2}\lambda_1\varepsilon^2-\frac{1}{3}\lambda_2\varepsilon^3-\dots, \notag
    \end{equation}
    where the arbitrary integration constant has been dropped. For the slight perturbation $\varepsilon$, displacement from equilibrium $x^*$ is small, we can linearize equation (S1) by using a first-order Taylor expansion
    \begin{equation}
    	\frac{d(x^*+\varepsilon)}{dt}=f(x^*+\varepsilon)\approx f(x^*)+\frac{\partial f}{\partial x}\Big |_{x=x^*}\varepsilon. \notag
    \end{equation}
    Therefore we have
    \begin{equation}
    	f(x^*)+\frac{d\varepsilon}{dt}=f(x^*)+\lambda_1\varepsilon \Rightarrow \frac{d\varepsilon}{dt}=\lambda_1\varepsilon. \tag{S2}
    \end{equation}
    
    Equation (S2) reflects the system's ability to return to the equilibrium after being perturbed, where $\lambda_1$ is the recovery rate. The system has high resilience when it is far from the bifurcation point, and its state will rapidly return to equilibrium after deviating from it (Fig. S11. a). As a local bifurcation is approached, $\lambda_1$ approaches 0 and the resilience of the system decreases, which is denoted as critical slowing down (Fig. S11. b).
    \begin{figure}[htbp]
	\centering
	\begin{minipage}[b]{0.4\linewidth}
		\centering
		\includegraphics[width=\linewidth]{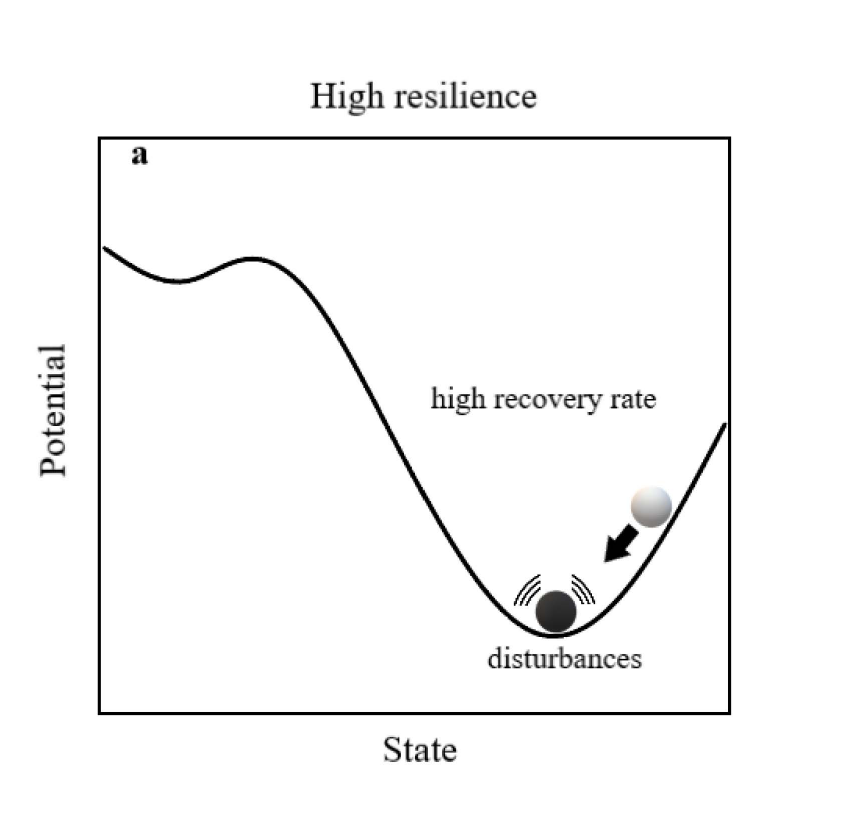}
	\end{minipage}
	\begin{minipage}[b]{0.4\linewidth}
		\centering
		\includegraphics[width=\linewidth]{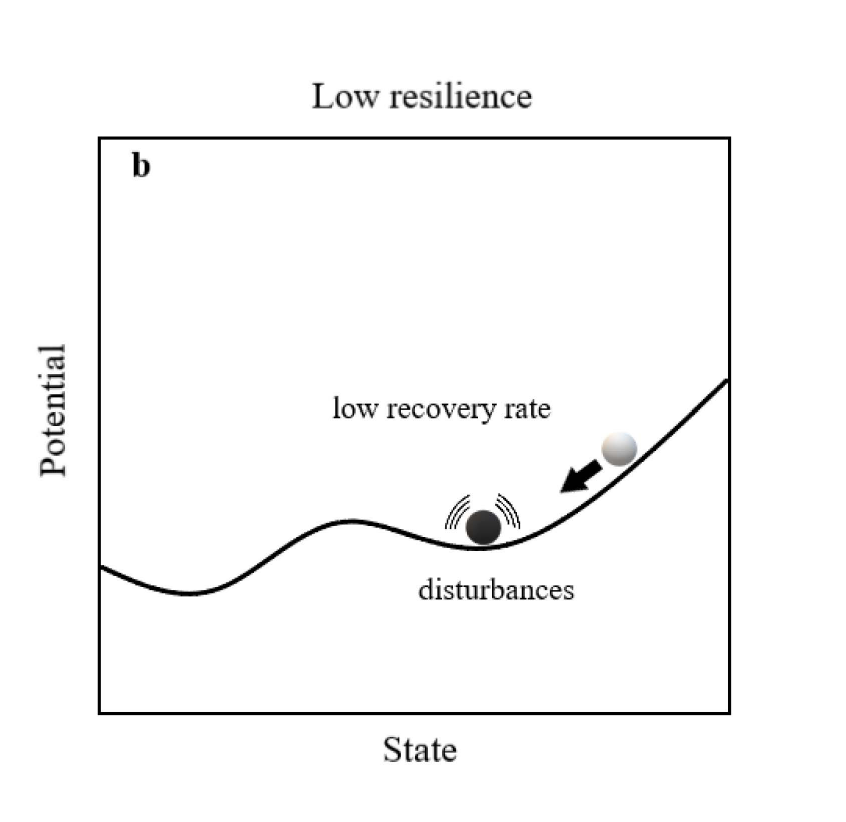}
	\end{minipage}
	\caption*{\textbf{Figure S11.}\fontsize{10pt}{12pt}\selectfont{
		The potential landscape of a system centered on its equilibrium. (a) The system has high resilience when it is far from the bifurcation point. (b) The system has low resilience when it is near the bifurcation point.
		}}
   \end{figure}
    
    We will next illustrate how critical slowing down can lead to an increase in the lag-1 autocorrelation. From equation (S2), we can solve the displacement $\varepsilon_t$ from equilibrium $x^*$,
    \begin{equation}
    	\varepsilon_t = e^{\lambda_1t}\Rightarrow x_t-x^*=e^{\lambda_1t}. \notag
    \end{equation}   
    Thus, the evolution of the system state over time can be described in a simple lag-1 autoregressive model:
    \begin{equation}
    	\begin{aligned}
    		\centering
    		x_{t+1}-x^*&=e^{\lambda_1(t+1)}+\sigma \xi_t=e^{\lambda_1}(x_t-x^*)+\sigma \xi_t  \\
    		\Leftrightarrow y_{t+1}&=e^{\lambda_1}y_t+\sigma \xi_t=\alpha y_t+\sigma \xi_t, \notag		
    	\end{aligned}
    \end{equation}
    where $y_t= x_t-x^*$, is the deviation of the state variable $x_t$ from the equilibrium $x^*$, $\xi_t$ is a random noise and $\sigma$ is the standard deviation. As $\lambda_1$ approaches zero, $\alpha=e^{\lambda_1}$ approaches 1, leading to an increase in the lag-1 autocorrelation. When the lag-1 autoregressive coefficient $\alpha$ reaches 1, a bifurcation occurs. Therefore, many methods based on the lag-1 autocorrelation of the system states, have been developed for tipping points prediction. For example, the degenerate fingerprinting and BB method are designed for predicting tipping points for systems driven by white noise and red noise respectively. Note that if the additive noise $\xi_t$ is red noise with lag-1 autocorrelation, the lag-1 autocorrelation of the system state will be partially from the lag-1 autocorrelation of the red noise, which make the lag-1 autoregressive coefficient $\alpha$ more difficult to estimate.
    
    Then we will show the significance of high-order terms in dynamical systems close to a bifurcation point. As a system is far from a bifurcation point in a regime of small noise, the displacement $\varepsilon$ from equilibrium is small. Thus in the simplified equation 
    \begin{equation}
    	\frac{d(x^*+\varepsilon)}{dt}=f(x^*+\varepsilon)\approx f(x^*)+\frac{\partial f}{\partial x}\Big |_{x=x^*}\varepsilon, \notag
    \end{equation}
    we can omit the higher-order terms and then simplify this dynamical system to equation (S2). However, as a local bifurcation is approached, the recovery rate $\lambda_1$ approaches 0 when critical slowing down occurs. This allows the noise to push the system farther from equilibrium which means $\varepsilon$ increases. Therefore high-order terms become significant for the dynamics of the system\cite{RN18}.

    \subsection{Supplementary Note 2. Fast-slow systems and rate-delayed tipping}
    \subsubsection{Fast-slow systems in critical transition}
    Consider a family of ordinary differential equations:
    \begin{equation}
    	\frac{dx}{dt}=f(x,\mu), \tag{S3}
    \end{equation}
    where $x\in R^m$ are phase space variables and $\mu\in R^m$ are parameters. Since the critical transitions occur when a parameter evolves slowly until a tipping point, it is a natural way to include the parameters in the original differential equation. So the equation (S3) can be written as
    \begin{equation}
    	\begin{aligned}
    		\frac{dx}{dt}&=f(x,\mu)\\
    		\frac{d\mu}{dt}&=0.
    	\end{aligned}
    	\tag{S4}
    \end{equation}
    
    Then we add a slow evolution to $\mu$ in equation (S4)
    \begin{equation}
    	\begin{aligned}
    		\frac{dx}{dt}&=f(x,\mu) \\
    		\frac{d\mu}{dt}&=\epsilon g(x,\mu), 
    	\end{aligned}
    	\tag{S5}
    \end{equation}
    where $0<\epsilon \ll 1$ is a small parameter and $g$ is sufficiently smooth. In many cases, we can assume that the parameter dynamics is decoupled from phase space dynamics and $g\equiv 1$. The equations (S5) form a fast–slow system where $x\in R^m$ are the fast variables, $\mu \in R^m$ are the slow variables and the parameter $\epsilon$ characterizes the time scale separation.
    
    When $\epsilon g(x,\mu)$ is small, system (S5) can be simplified to one-dimensional dynamical system (S3) based on the theory of fast-slow systems. Therefore the effect of the changing rate of the parameter on the location of the critical transition will not be taken into account.
    \subsubsection{Rate-delayed tipping}
    For dynamical system (S5), if $\epsilon g(x,\mu)$ is not sufficiently small, the changing rate of the bifurcation parameter $\mu$ will cause a delay on the Bifurcation-tipping. Here is an example of the normal form of fold bifurcation where the bifurcation parameter $\mu$ approaches the bifurcation point at a rate of $r$:
    \begin{align}
        \frac{dx}{dt}&=\mu +x^2 \tag{S6} \\
    	\frac{d\mu}{dt}&=r,   \tag{S7}
    \end{align}
    this dynamical system has a stable equilibrium branch $x^2=-\mu$ $(x<0)$ if we only consider the equation (S6). However, due to the nonzero changing rate of the bifurcation parameter described in equation (S7), the expression of this equilibrium branch is not entirely precise. If we take the derivative of $x^2=-\mu$ with respect to $t$ on both sides, we will obtain the following:
    \begin{equation}
    	\begin{aligned}
    		\frac{dx^2}{dt}&=2x\frac{dx}{dt}=-\frac{d\mu}{dt}=-r \\
    		 \Rightarrow \frac{dx}{dt}&=\frac{-r}{2x}=\mu +x^2 \\
    		  \Rightarrow -\mu&=x^2+\frac{r}{2x}. \notag
    	\end{aligned}
    \end{equation}
    
    This implies that the trajectory of the quasi-static attractor $x$ as it moves with the changing bifurcation parameter $\mu$ is approximately governed by $x^2+\frac{r}{2x}=-\mu$. Moreover, as the changing rate of $\mu$ increases (larger $r$ values), the trajectory of the quasi-static attractor $x$ deviates farther from $x^2=-\mu$. Here we set $r$ to be 0, 0.5 and 1 respectively, and plot the graph of $x^2+\frac{r}{2x}=-\mu$, as shown in Fig. S12. It can be observed that as $r$ increases, the occurrence of the tipping points is progressively delayed comparing with the bifurcation point $\mu=0$.
    \begin{figure}[htbp]
    	\centering
    	\includegraphics[trim=1.5cm 0.5cm 3cm 2.7cm, clip, width=1\linewidth]{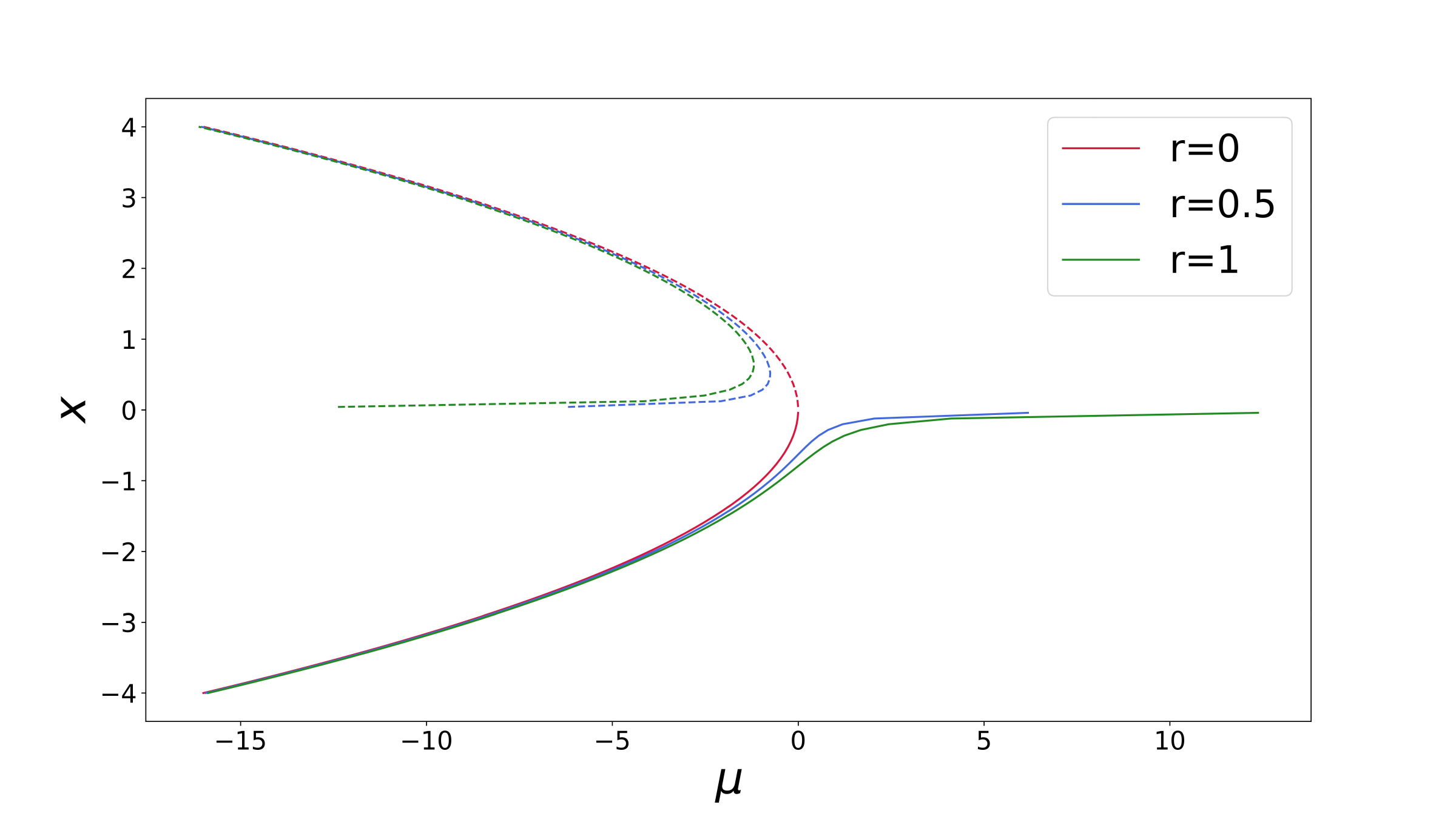}
    	\caption*{\textbf{Figure S12.}\fontsize{10pt}{12pt}\selectfont{
    	A rate-delayed tipping example of the normal form of fold bifurcation. The bifurcation occurs at $\mu=0$ in the normal form of fold bifurcation $dx/dt=\mu+x^2$. Solid lines represent stable equilibrium branches, while dashed lines represent unstable equilibrium branches.}}
    \end{figure}
    \subsubsection{Training labels of tipping points}
    Due to the effect of rate-delayed tipping, the real tipping points of the simulated training time series have been delayed. As a result, we have not utilized the bifurcation points given by AUTO-07P as the training labels. Instead, we identify the location of the tipping points where the recovery rate changes from negative value to positive value and use them as training labels for tipping points. This location is where the quasi-static attractor losses stability. Here, we provide some examples in the training set to show the location of our training labels. For each bifurcation type, we plot figures of training time series that undergo tipping points, generated by 50 different systems with white noise. Half of the figures represent systems with an increasing parameter, while the other half represent systems with a decreasing parameter. Then we compare the tipping points given by AUTO-07P with those identified by the recovery rate. It can be observed that due to the effect of rate-delayed tipping, the tipping points identified by the recovery rate are more accurate, as shown in Fig. S13-S18.
    \begin{figure}[htbp]
    	\centering
    	\includegraphics[trim=0 0 0 0.5cm, clip, width=0.8\linewidth]{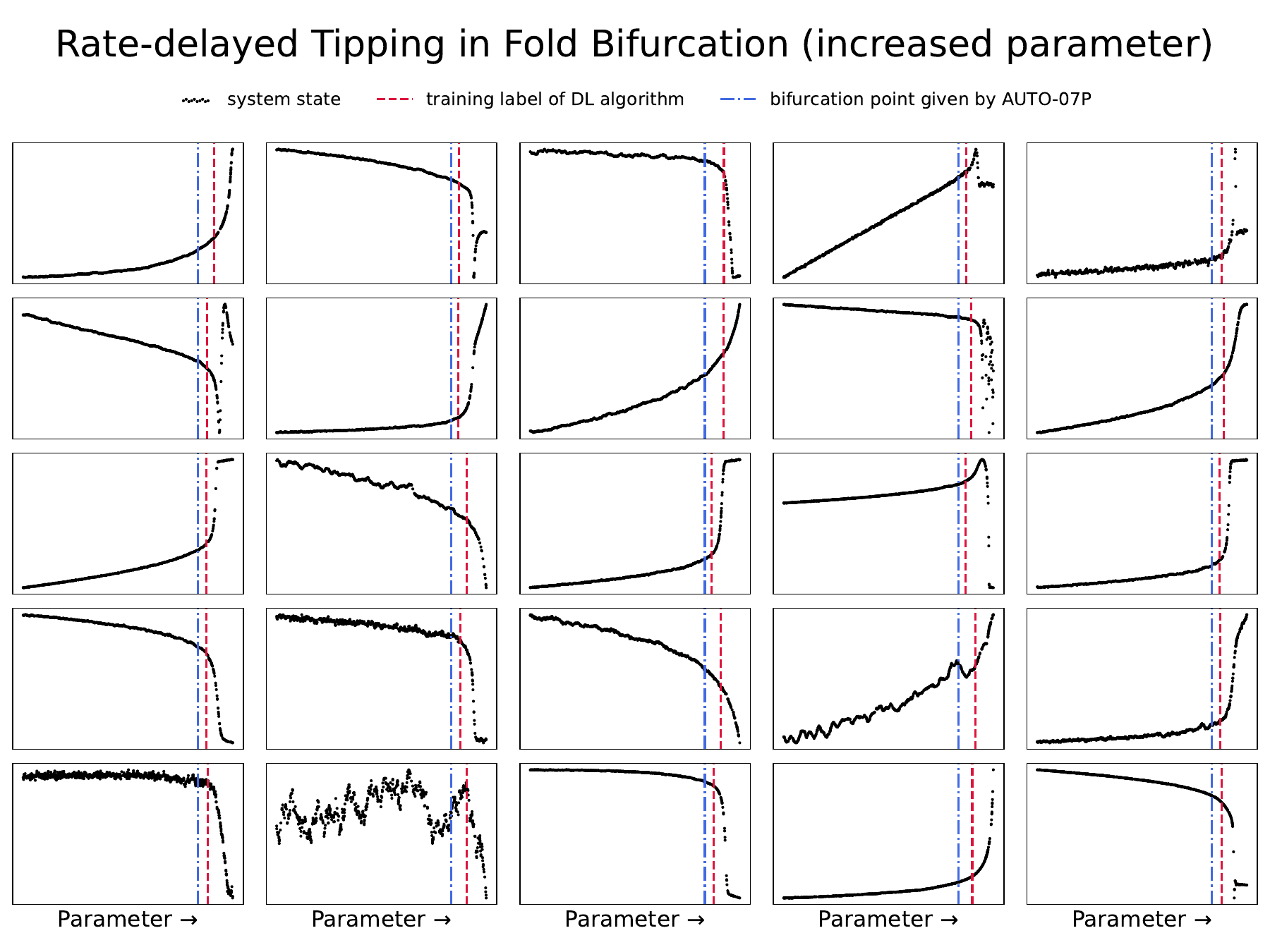}
    	\caption*{\textbf{Figure S13.}\fontsize{10pt}{12pt}\selectfont {
        An example of 25 different time series with white noise in our training set, each going through a fold bifurcation, where the bifurcation parameter is increasing. The red dashed lines are the labels used to train our DL algorithm identified by the recovery rate, while the blue dash-dot lines are the bifurcation points given by AUTO-07P.}}
   \end{figure}
   
   \clearpage
   \begin{figure}[htbp]
    	\centering
    	\includegraphics[trim=0 0 0 0.5cm, clip, width=0.8\linewidth]{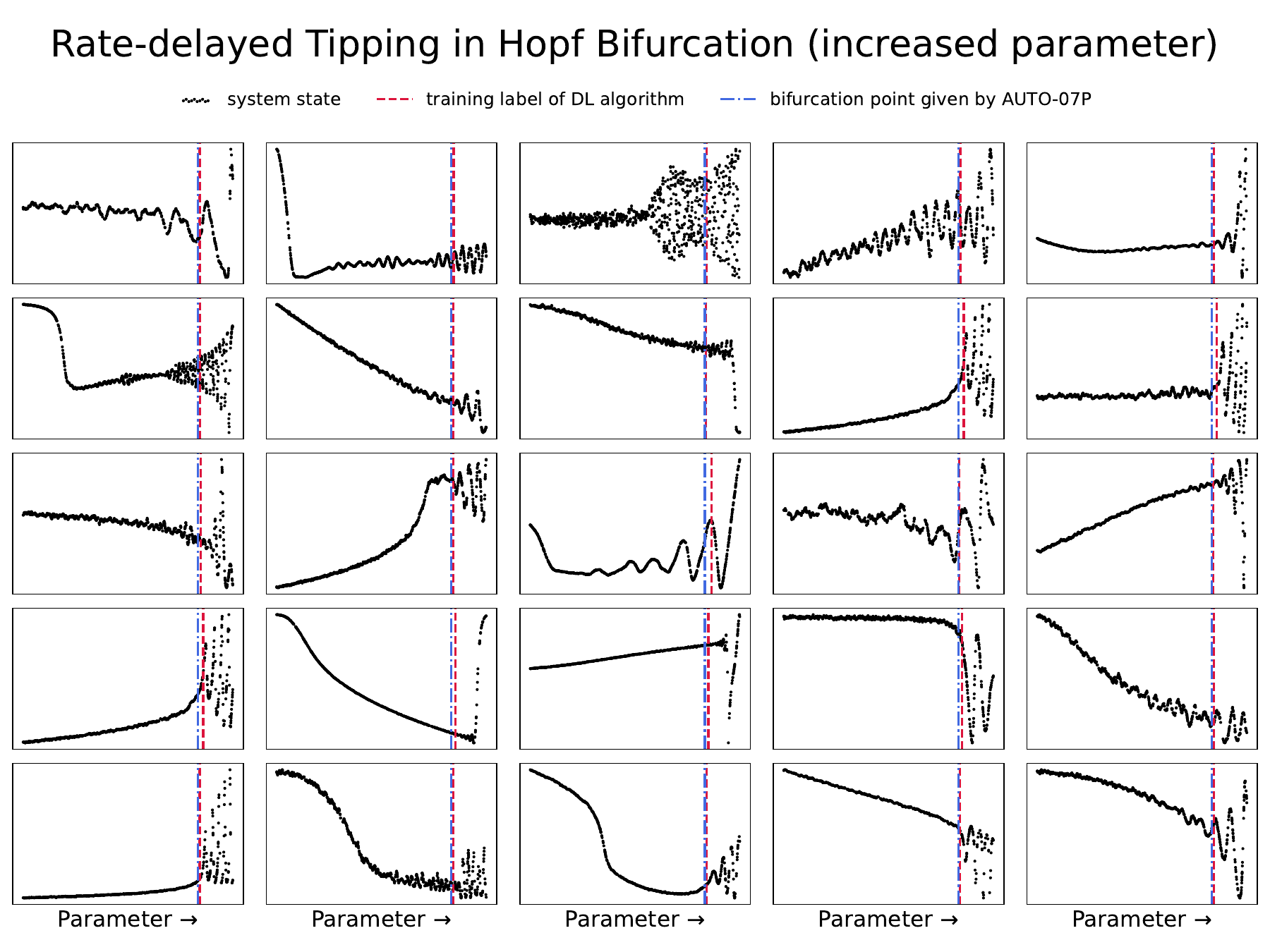}
        \caption*{\textbf{Figure S14.}\fontsize{10pt}{12pt}\selectfont{
        An example of 25 different time series with white noise in our training set, each going through a Hopf bifurcation, where the bifurcation parameter is increasing. The red dashed lines are the labels used to train our DL algorithm identified by the recovery rate, while the blue dash-dot lines are the bifurcation points given by AUTO-07P.}}
        \vspace{0.3cm}
    	\centering	
    	\includegraphics[trim=0 0 0 0.5cm, clip, width=0.8\linewidth]{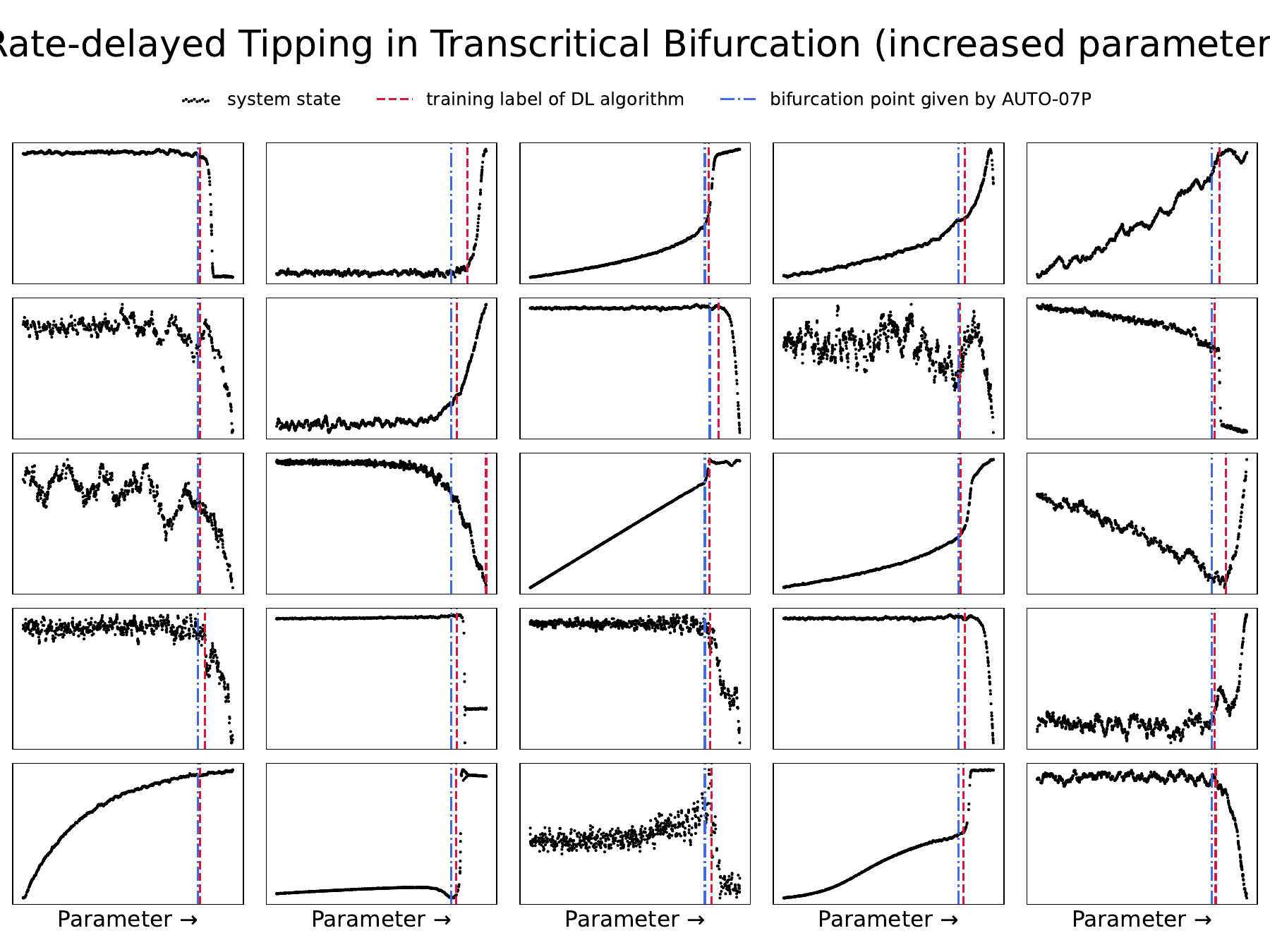}
        \caption*{\textbf{Figure S15.}\fontsize{10pt}{12pt}\selectfont{
        An example of 25 different time series with white noise in our training set, each going through a transcritical bifurcation, where the bifurcation parameter is increasing. The red dashed lines are the labels used to train our DL algorithm identified by the recovery rate, while the blue dash-dot lines are the bifurcation points given by AUTO-07P.}}
   \end{figure}
   
    \clearpage
   \begin{figure}[htbp]
	    \centering
    	\includegraphics[trim=0 0 0 0.5cm, clip, width=0.8\linewidth]{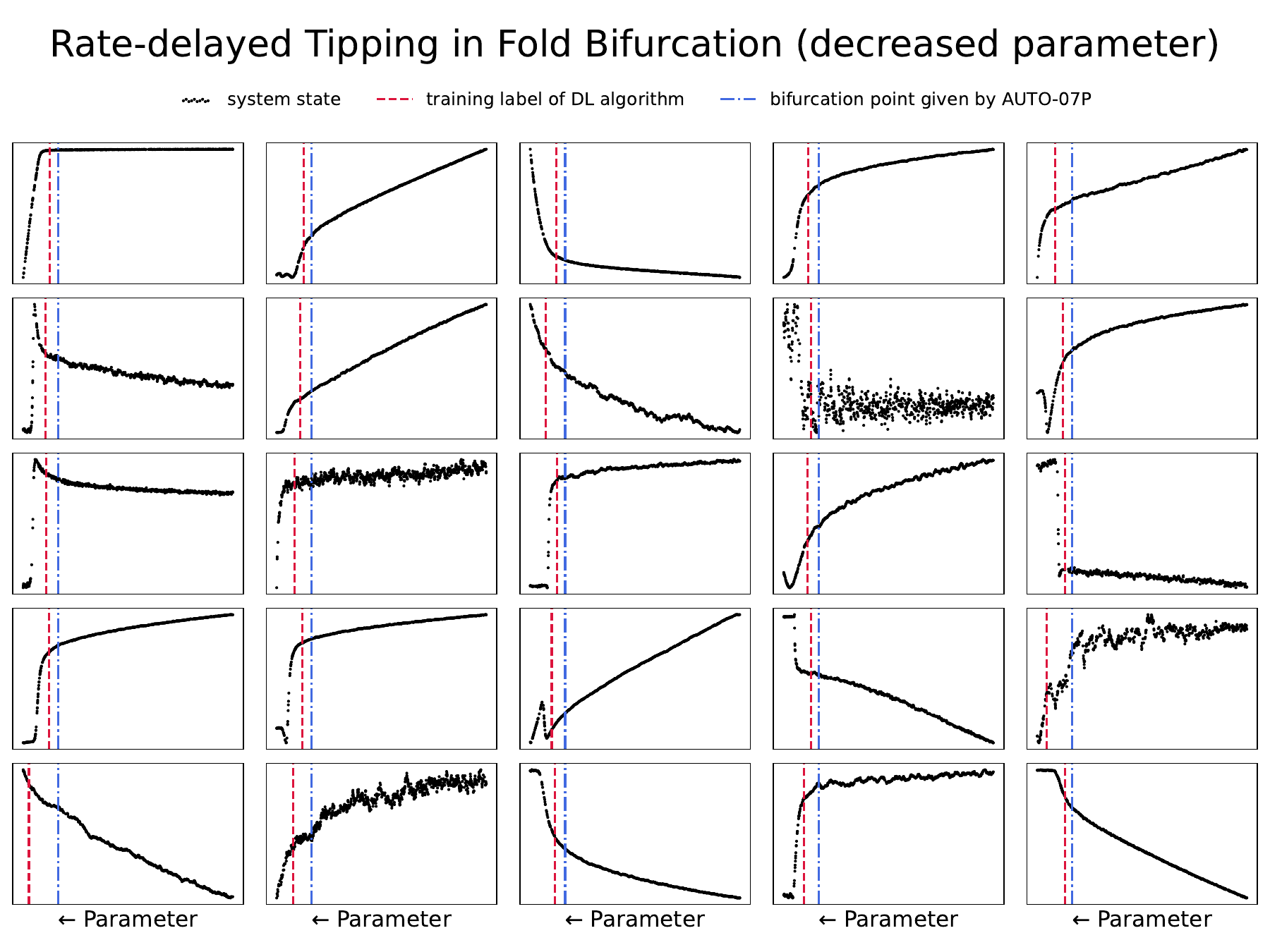}
    	\caption*{\textbf{Figure S16.}\fontsize{10pt}{12pt}\selectfont{
        An example of 25 different time series with white noise in our training set, each going through a fold bifurcation, where the bifurcation parameter is decreasing. The red dashed lines are the labels used to train our DL algorithm identified by the recovery rate, while the blue dash-dot lines are the bifurcation points given by AUTO-07P.}}
        \vspace{0.3cm}
    	\includegraphics[trim=0 0 0 0.5cm, clip, width=0.8\linewidth]{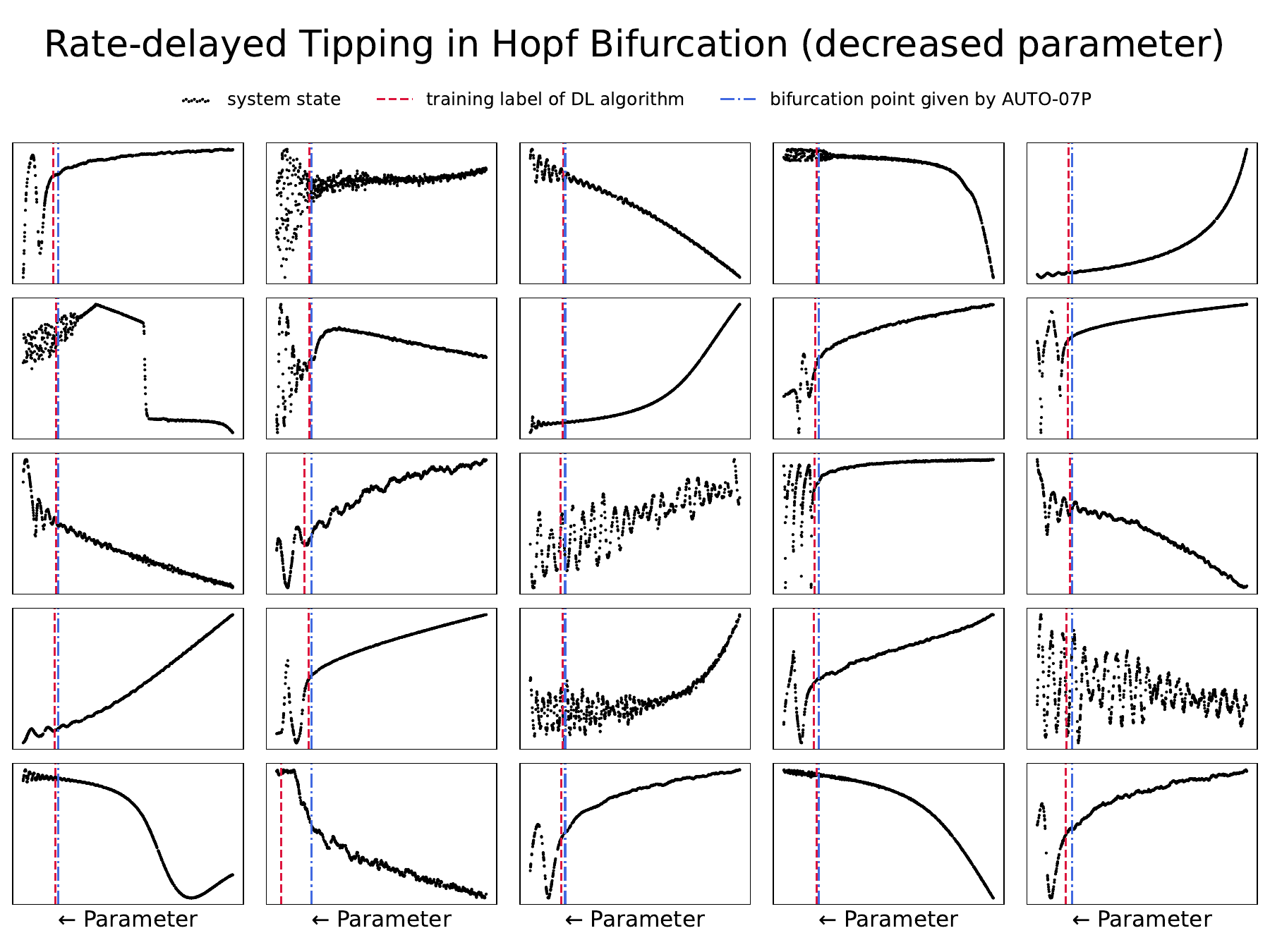}
        \caption*{\textbf{Figure S17.}\fontsize{10pt}{12pt}\selectfont{
        An example of 25 different time series with white noise in our training set, each going through a Hopf bifurcation, where the bifurcation parameter is decreasing. The red dashed lines are the labels used to train our DL algorithm identified by the recovery rate, while the blue dash-dot lines are the bifurcation points given by AUTO-07P.}}
    \end{figure}
    
    \clearpage
   \begin{figure}[htbp]
   	    \centering
    	\includegraphics[trim=0 0 0 0.5cm, clip, width=0.8\linewidth]{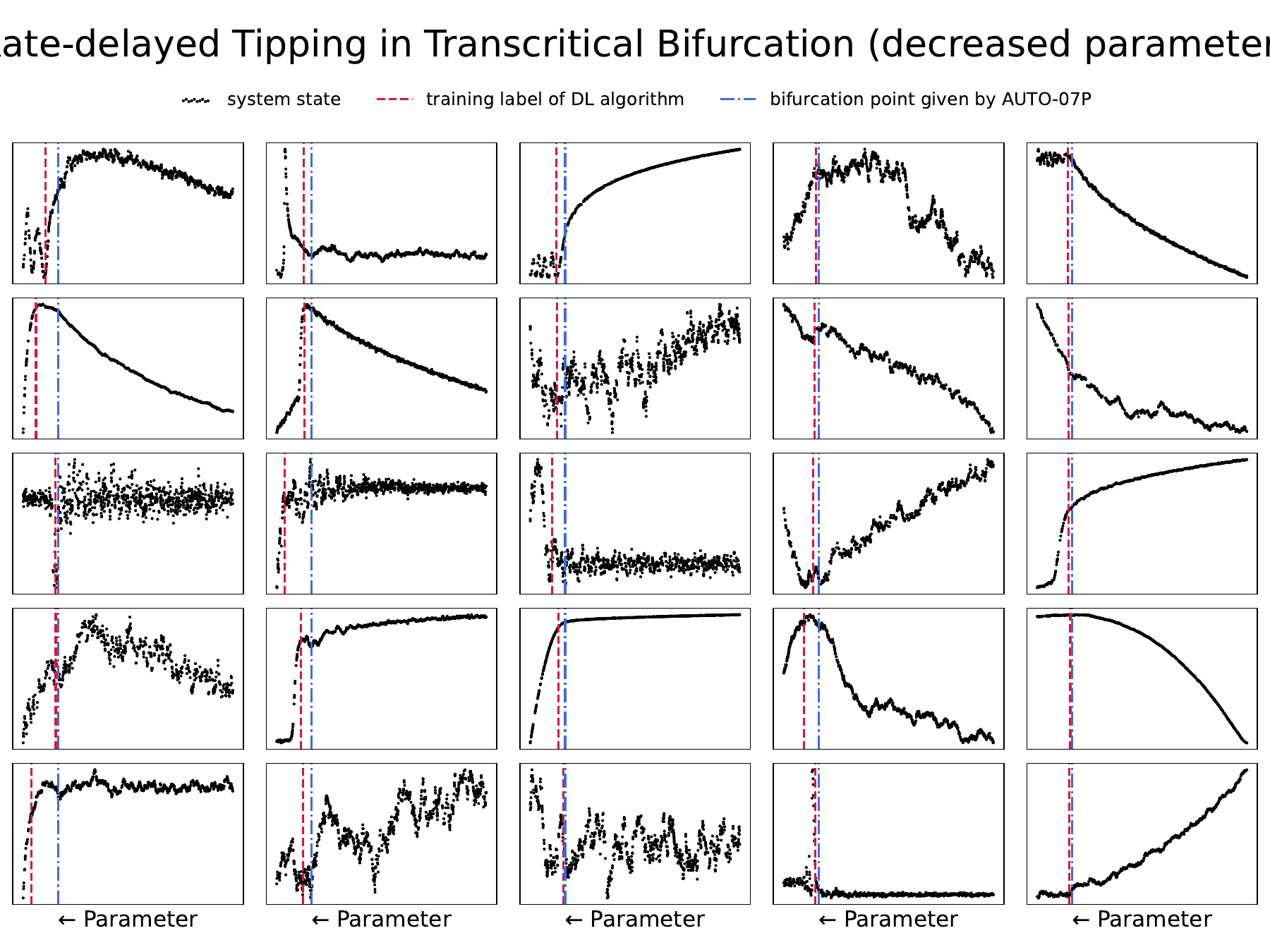}
    	\caption*{\textbf{Figure S18.}\fontsize{10pt}{12pt}\selectfont{
        An example of 25 different time series with white noise in our training set, each going through a transcritical bifurcation, where the bifurcation parameter is decreasing. The red dashed lines are the labels used to train our DL algorithm identified by the recovery rate, while the blue dash-dot lines are the bifurcation points given by AUTO-07P.}}
    \end{figure}
	
	\subsection{Supplementary Note 3. Embedding theorem for irregular sampling}
	Consider a $m$-dimensional continuous-time dynamical system
	\begin{equation}
		\frac{dx}{dt}=f(x), \notag
	\end{equation}
	whose state space is a differential manifold $M$, of finite dimension $m$. The dynamics can be described by a smooth flow
	\begin{equation}
		\phi:M \times \mathbb{R} \rightarrow M, \notag
	\end{equation}
	the initial state $x\in M$ evolves, after time $t$, to state $\phi(x,t)$. If we sample at $x$, the next sample will be taken at the state which $x$ evolves to after a time $\tau(x)$ ($\tau$ is a positive valued function where $\tau:M \rightarrow \mathbb{R}^+$), namely $\phi(x,\tau(x))$. Let us define the sampling state map $g:M \rightarrow M$ by $g(x)=\phi(x,\tau(x))$. Then sampling at $x$ means the next sample will be taken at $g(x)$, the sample after that at $g(g(x))=g^2(x)$, and so on. The sequence of sampled states is $\{x,g(x),\dots,g^n(x),\dots\}$ and the sampling intervals are $\{\tau(x),\tau(g(x)),\dots,\tau(g^n(x)),\dots\}$.
	
	If we want to measure the value of some property of the system, such as the temperature, the voltage at some point in an electrical system, and the velocity at some point in a fluid system, we can suppose there is a function $y:M \rightarrow \mathbb{R}$ such that, if the system
	is in state $x$, the result of the measurement is $y(x)$. Thus the sequence of measured values of sampled states is $\{y(x),y(g(x)),\dots,y(g^n(x)),\dots\}$. Based on these definitions, for each state $x$ of the system, its corresponding delay vector based on a map $\Phi_{y,g}$ is
	\begin{equation}
	\Phi_{y,g}(x)=(y(x),y\circ g(x),\dots,y\circ g^{d-1}(x)). \tag{S8}
    \end{equation}
    
    Prior to presenting the embedding theorem for irregular sampling\cite{RN205}, we first introduce the following lemma\cite{RN205}.
    \begin{lemma}
    	Let $X$ be a $C^r$ $(r\geq 1)$ vector field on a compact connected manifold $M$, and $\tau:M\rightarrow \mathbb{R}^+$ a $C^r$ function. The mapping $g:M \rightarrow M$ defined by $g(x)=\phi(x,\tau(x))$, (where $\phi:M \times \mathbb{R}^+ \rightarrow M$ is the flow arising from $X$), is a $C^r$ diffeomorphism if and only if $D\tau(x)X(x)>-1$ for all $x\in M$.
    \end{lemma}

    The following embedding theorem for irregular sampling is based on the preceding definitions and the \textbf{Lemma 1}. 
    \begin{theorem}
    	If $d>2m$ then the triples $(y,\tau,X)$, for which the map $\Phi_{y,g}$ defined in (S8) is an embedding, form an open and dense subset of $C^2(M,\mathbb{R}) \times \mathcal{T}$ where $\mathcal{T} \subset C^2(M,\mathbb{R}^+) \times \mathcal{X}^2(M)$ is $\{(\tau,x):D\tau(x)X(x)>-1,\forall x \in M\}$.
    \end{theorem}
    
    Based on the \textbf{Lemma 1} and the \textbf{Theorem 1}, we find that if $X$ is a $C^r$ $(r\geq 2)$ vector field on a compact connected manifold $M$, $\tau:M\rightarrow \mathbb{R}^+$ is a $C^r$ function and the sampling function $g:M \rightarrow M$ is a $C^2$ diffeomorphism, the map $\Phi_{y,g}(x)=(y(x),y\circ g(x),\dots,y\circ g^{d-1}(x))$ is an embedding ($d>2m$). Here, we explain the irregular sampling on a dynamical system defined by a $C^r$ $(r\geq 2)$ vector field approaching a bifurcation and $\tau:M\rightarrow \mathbb{R}^+$ is a $C^r$ function, where the employed sampling function $g=\phi(x,\tau(x))$ is a $C^2$ diffeomorphism. 
    
    First, we note that as a dynamical system approaches a bifurcation, its dynamics can be simplify to those of the normal form of that bifurcation. Thus each component of the dynamical function $f$ is either always positive or always negative, without loss of generality, we can assume that $f_i$ which is the $i$-th component of the function $f$ is always negative. Suppose the domain of the function $g$ is $M_g$, $g_i$ is the $i$-th component of the sampling function $g$, $g_i=\phi_i(x,\tau(x))$, we prove that the sampling function $g$ is a $C^2$ diffeomorphism which has three properties:
    
    (i). Bijectivity: $\forall g_k$ ($k=1,2,\dots,m$) and $x^{(1)},x^{(2)} \in M_g$, $x_k^{(i)}$ is the $k$-th component of $x^{(i)}$, if $x_k^{(1)}>x_k^{(2)}$, due to the function $f_k$ is always negative, we note that $g_k(x^{(1)})$ is the largest one among numbers less than $x_k^{(1)}$ in the $k$-th component of $M_g$. Thus we have
    \begin{equation}
    	x_k^{(1)}>g_k(x^{(1)})\geq x_k^{(2)}>g_k(x^{(2)}) \Rightarrow g_k(x^{(1)})>g_k(x^{(2)}), \notag
    \end{equation}
    which means that $g_k$ is a monotonically increasing function. Thus the sampling function $g$ satisfies bijectivity.
    
    (ii). Differentiability: Assume that when the system evolves to state $x=(x_1,x_2,\dots,x_m)$, the time is $t_x$. $\forall g_k$ ($k=1,2,\dots,m$), $g_k(x)$ can be given by
    \begin{equation}
    \begin{aligned}
    	g_k(x)&=x_k+\int_{t_x}^{t_x+\tau(x)}f_k(x(s))ds \\
    	&=x_k+\int_{0}^{\tau(x)}f_k(\phi(x,s))ds,
    \end{aligned}
    \tag{S9}
    \end{equation}
    we differentiate both sides of equation (S9) with respect to $x_i$, we have
    \begin{equation}
     \frac{\partial g_k(x)}{\partial x_i}=\frac{\partial x_k}{\partial x_i}+f_k(\phi(x,\tau(x)))\frac{\partial \tau(x)}{\partial x_i}+\int_{0}^{\tau(x)}\sum_{p=1}^{m}\frac{\partial \phi_p}{\partial x_i}(x,s)\frac{\partial f_k}{\partial \phi_p}(\phi(x,s))ds, \notag
    \end{equation}
    where we note that $\frac{\partial f_k}{\partial \phi_p}=\frac{\partial f_k}{\partial x_p}$, we have   
    \begin{equation}
     \frac{\partial g_k(x)}{\partial x_i}=\frac{\partial x_k}{\partial x_i}+f_k(\phi(x,\tau(x)))\frac{\partial \tau(x)}{\partial x_i}+\int_{0}^{\tau(x)}\sum_{p=1}^{m}\frac{\partial \phi_p}{\partial x_i}(x,s)\frac{\partial f_k}{\partial x_p}(\phi(x,s))ds. \tag{S10}
    \end{equation}
    Then, we differentiate both sides of equation (S10) with respect to $x_j$, we have
    \begin{equation}
    \begin{aligned}
    \frac{\partial^2 g_k(x)}{\partial x_i\partial x_j}&=\sum_{p=1}^{m}[\frac{\partial \phi_p}{\partial x_j}(x,\tau(x))+\frac{\partial \phi_p}{\partial \tau}(x,\tau(x))\frac{\partial \tau(x)}{\partial x_j}]\frac{\partial f_k}{\partial \phi_p}(\phi(x,\tau(x)))\frac{\partial \tau(x)}{\partial x_i} \\
   	&+f_k(\phi(x,\tau(x)))\frac{\partial^2 \tau(x)}{\partial x_i\partial x_j}+\sum_{p=1}^{m}[\frac{\partial \phi_p}{\partial x_i}(x,\tau(x))\frac{\partial f_k}{\partial x_p}(\phi(x,\tau(x)))]\frac{\partial \tau(x)}{\partial x_j} \\
    &+\int_{0}^{\tau(x)}\sum_{p=1}^{m}\{\frac{\partial^2 \phi_p}{\partial x_i\partial x_j}(x,s)\frac{\partial f_k}{\partial x_p}(\phi(x,s))+\frac{\partial \phi_p}{\partial x_i}(x,s)\sum_{q=1}^{m}[\frac{\partial \phi_q}{\partial x_j}(x,s)\frac{\partial^2 f_k}{\partial x_p\partial \phi_q}(\phi(x,s))]\}ds, \notag
    \end{aligned}    
   \end{equation}
    where we note that $\frac{\partial \phi_p(x,t)}{\partial t}=f_p(\phi(x,t))$ and $\frac{\partial f_k}{\partial \phi_p}=\frac{\partial f_k}{\partial x_p}$, we have
    \begin{equation}
   	\begin{aligned}
   	\frac{\partial^2 g_k(x)}{\partial x_i\partial x_j}&=\sum_{p=1}^{m}[\frac{\partial \phi_p}{\partial x_j}(x,\tau(x))+f_p(\phi(x,\tau(x)))\frac{\partial \tau(x)}{\partial x_j}]\frac{\partial f_k}{\partial x_p}(\phi(x,\tau(x)))\frac{\partial \tau(x)}{\partial x_i} \\
   	&+f_k(\phi(x,\tau(x)))\frac{\partial^2 \tau(x)}{\partial x_i\partial x_j}+\sum_{p=1}^{m}[\frac{\partial \phi_p}{\partial x_i}(x,\tau(x))\frac{\partial f_k}{\partial x_p}(\phi(x,\tau(x)))]\frac{\partial \tau(x)}{\partial x_j} \\
   	&+\int_{0}^{\tau(x)}\sum_{p=1}^{m}\{\frac{\partial^2 \phi_p}{\partial x_i\partial x_j}(x,s)\frac{\partial f_k}{\partial x_p}(\phi(x,s))+\frac{\partial \phi_p}{\partial x_i}(x,s)\sum_{q=1}^{m}[\frac{\partial \phi_q}{\partial x_j}(x,s)\frac{\partial^2 f_k}{\partial x_p\partial x_q}(\phi(x,s))]\}ds.
   	\end{aligned}
   	\tag{S11}
   \end{equation}
    We note that $\tau(x) \in C^r$ and $X$ is a $C^r$ $(r\geq 2)$ vector field on a compact connected manifold $M$, thus $g_k(x)$ is first-order differentiable (S10) and second-order differentiable (S11). Thus the sampling function $g \in C^2$ satisfies differentiability.  
    
    (iii). Inverse Differentiability: Since $g_k$ is monotonically increasing and differentiable, and clearly $g_k \not\equiv 0$, according to the Inverse Function Theorem, $g_k^{-1}$ exists and is differentiable. Thus $g^{-1}$ exists and is differentiable, $g$ satisfies inverse differentiability.
    
    Thus based on the embedding theorem for irregular sampling, as long as the length $d$ of the convolutional kernels in the CNN layer is more than twice the dimension $m$ of the study or training system, these kernels can extract dynamical features of the entire system from segments of the irregularly-sampled time series from a single variable. Then these extracted features serve as input for the LSTM layer for predicting tipping points.
    
    Additionally, it is important to note that due to the time-varying nonstationarity of dynamical system approaching bifurcation, the features extracted from shorter convolutional kernel should contain much more dynamical information of the system, compared to those extracted from longer convolutional kernel. Therefore, although the length $d$ of the convolutional kernel is required to be more than twice the dimension $m$ of the system, it should not be excessively long. In our DL algorithm training, the length of the convolutional kernel is tuned through hyperparameter random search, and the length used is ten.
	
	\subsection{Supplementary Note 4. Normal form}
	\subsubsection{The formula for the recovery rate in the normal form}
	The normal form of the fold bifurcation is
	\begin{equation}
		\frac{dx}{dt}=\mu+x^2, \notag
	\end{equation}
	which exhibits a stable equilibrium branch $x^*(\mu)=-\sqrt{-\mu}$. Therefore, we have the following relation between the recovery rate $\lambda$ and the bifurcation parameter $\mu$,
	\begin{equation}
		\begin{aligned}
			\lambda&=\frac{\partial (\mu+x^2)}{\partial x}\Big |_{x=x^*(\mu)} \\
			&=2x\Big |_{x=-\sqrt{-\mu}} \\
			&=-2\sqrt{-\mu}. \notag
		\end{aligned}
	\end{equation}
	
	The normal form of the supercritical Hopf bifurcation is
	\begin{equation}
		\frac{dx}{dt}=\mu x-y-x(x^2+y^2),\, \frac{dy}{dt}=x+\mu y-y(x^2+y^2), \notag
	\end{equation}
	which exhibits a stable equilibrium branch $(x^*(\mu),y^*(\mu))=(0,0)$. Therefore, we have the following relation between the recovery rate $\lambda$ and the bifurcation parameter $\mu$,
	\begin{equation}
		\begin{aligned}
		\lambda&=max(Re(eigvals(
			\begin{bmatrix}
				\frac{\partial (\mu x-y-x(x^2+y^2))}{\partial x} & \frac{\partial (\mu x-y-x(x^2+y^2))}{\partial y} \\
				\frac{\partial (x+\mu y-y(x^2+y^2))}{\partial x} & \frac{\partial (x+\mu y-y(x^2+y^2))}{\partial y}
			\end{bmatrix}_{(x,y)=(x^*(\mu),y^*(\mu))})))\\
			&=max(Re(eigvals(
			 \begin{bmatrix}
				\mu-3x^2-y^2 & -1-2xy \\
				1-2xy & \mu-x^2-3y^2
			\end{bmatrix}_{(x,y)=(0,0)})))\\
			&=max(Re(eigvals(
			 \begin{bmatrix}
				\mu & -1 \\
				1 & \mu
			 \end{bmatrix}))) \\
			 &=max(Re(\mu \pm i)) \\
			 &=\mu. \notag
		\end{aligned}
	\end{equation}
	
	The normal form of the subcritical Hopf bifurcation is
   \begin{equation}
	\frac{dx}{dt}=\mu x-y+x(x^2+y^2),\, \frac{dy}{dt}=x+\mu y+y(x^2+y^2), \notag
    \end{equation}
    which exhibits a stable equilibrium branch $(x^*(\mu),y^*(\mu))=(0,0)$. Therefore, we have the following relation between the recovery rate $\lambda$ and the bifurcation parameter $\mu$,
    \begin{equation}
	\begin{aligned}
		\lambda&=max(Re(eigvals(
		\begin{bmatrix}
			\frac{\partial (\mu x-y+x(x^2+y^2))}{\partial x} & \frac{\partial (\mu x-y+x(x^2+y^2))}{\partial y} \\
			\frac{\partial (x+\mu y+y(x^2+y^2))}{\partial x} & \frac{\partial (x+\mu y+y(x^2+y^2))}{\partial y}
		\end{bmatrix}_{(x,y)=(x^*(\mu),y^*(\mu))})))\\
		&=max(Re(eigvals(
		\begin{bmatrix}
			\mu+3x^2+y^2 & -1+2xy \\
			1+2xy & \mu+x^2+3y^2
		\end{bmatrix}_{(x,y)=(0,0)})))\\
		&=max(Re(eigvals(
		\begin{bmatrix}
			\mu & -1 \\
			1 & \mu
		\end{bmatrix}))) \\
		&=max(Re(\mu \pm i)) \\
		&=\mu. \notag
	\end{aligned}
    \end{equation}
	
	The normal form of the transcritical bifurcation is
	\begin{equation}
		\frac{dx}{dt}=\mu x-x^2, \notag
	\end{equation}
	which exhibits a stable equilibrium branch $x^*(\mu)=0$ when $\mu$ increases from negative value to positive value or a stable equilibrium branch $x^*(\mu)=\mu$ when $\mu$ decreases from positive value to negative value. Therefore, the relation between the recovery rate $\lambda$ and the bifurcation parameter $\mu$ when $\mu$ increases from negative value to positive value is:
	\begin{equation}
		\begin{aligned}
			\lambda&=\frac{\partial (\mu x-x^2)}{\partial x}\Big |_{x=x^*(\mu)} \\
			&=\mu-2x\Big |_{x=0} \\
			&=\mu \quad (\mu<0). \notag
		\end{aligned}
	\end{equation}
	The relation between the recovery rate $\lambda$ and the bifurcation parameter $\mu$ when $\mu$ decreases from positive value to negative value is:
	\begin{equation}
		\begin{aligned}
			\lambda&=\frac{\partial (\mu x-x^2)}{\partial x}\Big |_{x=x^*(\mu)} \\
			&=\mu-2x\Big |_{x=\mu} \\
			&=-\mu \quad (\mu>0), \notag
		\end{aligned}
	\end{equation}
	which is equivalent to $\lambda=\mu \quad (\mu<0)$ when $\mu$ increases from negative value to positive value.
	\subsubsection{Dimension reduction near a bifurcation}
	Here we demonstrate how to simplify the dynamics of an arbitrary $n$-dimensional dynamical system $dX^n/dt = F(X^n,\mu)$ exhibiting a fold bifurcation to its normal form and derive the relation between the recovery rate $\lambda$ and the bifurcation parameter $\mu$ \cite{RN203}.
	
	As an $n$-dimensional dynamical system $dX^n/dt=F(X^n,\mu)$ approaches a bifurcation, its dynamics simplify to a one-dimensional dynamical system $dx/dt=f(x,\mu)$ according to center manifold theorem. We examine the behavior of $dx/dt=f(x,\mu)$ near the bifurcation at $x=x^*$ and $\mu=\mu_c$. Taylor's expansion yields
	\begin{equation}
		\begin{aligned}
		\frac{dx}{dt}&=f(x,\mu) \\
		&=f(x^*,\mu_c)+(x-x^*)\frac{\partial f}{\partial x}\Big |_{(x^*,\mu_c)}+(\mu-\mu_c)\frac{\partial f}{\partial \mu}\Big |_{(x^*,\mu_c)}+\frac{1}{2}(x-x^*)^2\frac{\partial^2 f}{\partial x^2}\Big |_{(x^*,\mu_c)}+\dots,  \notag			
		\end{aligned}
	\end{equation}
	where the quadratic terms in $(\mu-\mu_c)$ and cubic terms in $(x-x^*)$ are neglected. Two terms in this equation vanish: $f(x^*,\mu_c)=0$ since $x^*$ is a fixed point, and $\frac{\partial f}{\partial x}\Big |_{(x^*,\mu_c)}=0$ by the non-hyperbolicity of the fold bifurcation. Thus, we have
	\begin{equation}
		\frac{dx}{dt}=a(\mu-\mu_c)+b(x-x^*)^2, \tag{S12}
	\end{equation}
    where $a=\frac{\partial f}{\partial \mu}\Big |_{(x^*,\mu_c)}$, $b=\frac{1}{2}\frac{\partial^2 f}{\partial x^2}\Big |_{(x^*,\mu_c)}$, and for the transversality and non-degeneracy of fold bifurcation, $\frac{\partial f}{\partial \mu}\Big |_{(x^*,\mu_c)}\neq 0, \frac{\partial^2 f}{\partial x^2}\Big |_{(x^*,\mu_c)}\neq 0$ are satisfied. Thus, equation (S12) agrees with the normal form of fold bifurcation $dx/dt=\mu+x^2$.
    
    Since the equilibrium of the system (S12) is $x=x^*+\frac{1}{b}\sqrt{\lvert a(\mu_c-\mu) \rvert}$, we can derive the relation between the recovery rate $\lambda$ and the bifurcation parameter $\mu$ in the following 
    \begin{equation}
    	\begin{aligned}
    		 \lambda&=\frac{\partial}{\partial x}(a(\mu-\mu_c)+b(x-x^*)^2)\Big |_{x=x^*+\frac{1}{b}\sqrt{\lvert a(\mu_c-\mu) \rvert}} \\
    		 &=2b(x-x^*)\Big |_{x=x^*+\frac{1}{b}\sqrt{\lvert a(\mu_c-\mu) \rvert}} \\
    		 &=-2\sqrt{\lvert ab(\mu_c-\mu) \rvert}. \notag
    	\end{aligned}
    \end{equation}
    This function is a translation and scaling transformation of the function $\lambda=-2\sqrt{-\mu}$ which is the relation between the recovery rate $\lambda$ and the bifurcation parameter $\mu$ of the normal form of the fold bifurcation.
    
	\subsection{Supplementary Note 5. Competing algorithms}
	\subsubsection{Degenerate fingerprinting}
	The degenerate fingerprinting\cite{RN21} is applicable for tipping points prediction in high-dimensional dynamical systems with white noise. In the small-noise limit, the system's response to white noise can be approximated by the dynamics of linear modes. According to the theory of dynamical systems, one mode becomes unstable at any bifurcation which is called critical mode when the smallest decay rate $\kappa$ of perturbation vanishes. The critical mode produces diverging variance $\propto 1/\kappa$ as a bifurcation point is approached. Therefore, the critical mode can be approximated by leading principal component obtained by using principal component analysis (PCA) on high-dimensional time series data sampled from the system.
	
	The vicinity to a bifurcation allows for a simplification in the time-domain. Suppose that the other modes
	have much larger decay rate $\kappa_i$, their dynamics can be lumped into the noise. Then we can pre-aggregate the leading principal component time-series into a time-discrete dynamics of fixed time-step $\Delta t$ with $\Delta t \gg 1/\kappa_i$. If furthermore $1/\kappa \gg \Delta t$, the fluctuations of the critical mode can be modeled by a AR(1) process $y_{t+\Delta t}=\phi y_t+\xi_t=e^{-\kappa \Delta t}y_t+\xi_t$, where $\xi_t$ is Gaussian white noise. Thus we can estimate the lag-1 autoregressive coefficient $\phi$ from leading principal component time-series $y_t$. $\phi$ is used to be indicator for predicting the occurrence of tipping points, when $\phi$ reaches 1, we can predict that a bifurcation occurs.
	
	\subsubsection{BB method}
	The BB method\cite{RN16} is designed for estimating the lag-1 autoregressive coefficient of a time series sampled from one-dimensional systems with red noise. Due to the Takens embedding theorem, it can be applicable to a time series from one dimension of a high-dimensional system. Thus we compare the BB method with our DL algorithm on a time series from one dimension of high-dimensional systems in the main manuscript. The evolution of a one-dimensional time series of state $x_t$ under the disturbance of red noise $v_t$ over time can be modeled by
	\begin{equation}
		x_{t+1}=\varphi x_t+v_t,\quad v_{t+1}=\rho v_t+\epsilon_t,\quad \epsilon_t \sim N(0,1), \notag
	\end{equation}
	which has the following statistical property:
	\begin{equation}
		AC(x_{t+1},x_t)=\varphi_b=\frac{\varphi+\rho}{1+\varphi \rho}. \tag{S13}
	\end{equation}
	where $AC(x_{t+1},x_t)$ is the lag-1 autoregressive coefficient of time series $x_t$. The unbiased least-squares estimator for $\varphi_b$ is
	\begin{equation}
		\widehat{\varphi}_b=\frac{\sum_{i=1}^{n}(x_i-\overline{x})(x_{i-1}-\overline{x})}{\sum_{i=1}^{n}(x_{i-1}-\overline{x})^2}, \notag
	\end{equation}
	which only coincides with $\varphi$ for the white-noise case $\rho=0$. Since increasing $\varphi$ is an early warning signal for critical transition but the increase in $\rho$ leads to an increase in $\varphi_b$, thus $\varphi_b$ is not an effective early warning signal for the critical transition under red noise. For the least-squares estimator for $\rho$,
	\begin{equation}
		\widehat{\rho}_b=\frac{\sum_{i=1}^{n}\widehat{v}_i\widehat{v}_{i-1}}{\sum_{i=1}^{n}\widehat{v}_{i-1}^2},\quad \widehat{v}_i=x_{i+1}-\widehat{\varphi}_b x_i, \notag
	\end{equation}
	we have the following convergence property
	\begin{equation}
		\rho_b=\varphi \rho \varphi_b. \tag{S14}
	\end{equation}
	
	Using the equations (S13) and (S14), we have 
	\begin{equation}
		\varphi^2-(\varphi_b+\rho_b)\varphi+\frac{\rho_b}{\varphi_b}=0. \notag
	\end{equation}
	Thus the unbiased estimator $\widehat{\varphi}$ of $\varphi$ for $\varphi>\rho$ is given by
	\begin{equation}
		\widehat{\varphi}=\frac{(\widehat{\varphi}_b+\widehat{\rho}_b)+\sqrt{(\widehat{\varphi}_b+\widehat{\rho}_b)^2-4\frac{\widehat{\rho}_b}{\widehat{\varphi}_b}}}{2}, \notag
	\end{equation}
	and for $\rho>\varphi$ is given by
	\begin{equation}
		\widehat{\varphi}=\frac{(\widehat{\varphi}_b+\widehat{\rho}_b)-\sqrt{(\widehat{\varphi}_b+\widehat{\rho}_b)^2-4\frac{\widehat{\rho}_b}{\widehat{\varphi}_b}}}{2}, \notag
	\end{equation}
	where
	\begin{equation}
		\widehat{\varphi}_b=\frac{\sum_{i=1}^{n}(x_i-\overline{x})(x_{i-1}-\overline{x})}{\sum_{i=1}^{n}(x_{i-1}-\overline{x})^2},\quad 
		\widehat{\rho}_b=\frac{\sum_{i=1}^{n}\widehat{v}_i\widehat{v}_{i-1}}{\sum_{i=1}^{n}\widehat{v}_{i-1}^2},\quad \widehat{v}_i=x_{i+1}-\widehat{\varphi}_b x_i. \notag
	\end{equation}
	
	The estimator $\widehat{\varphi}$ of $\varphi$ is used to be an indicator for predicting the occurrence of tipping points. When $\widehat{\varphi}$ reaches 1, we can predict that a bifurcation occurs.
	
	\subsubsection{Dynamical eigenvalue}
	Based on the Takens embedding theorem, for a one-dimensional state time series $\left \{ x(t)\mid 1\leq t \leq n \right \}$ from a $d$-dimensional dynamical system $dy/dt=f(y)$, one can employ state space reconstruction to find an embedding space $X_t=[x(t),x(t-\tau),\dots,x(t-(E-1)\tau)]$ of dimension $E$ which is topologically equivalent to original state space. Here, $\tau$ represents the time delay, and $E>2d+1$. Therefore, the dominant eigenvalue of the Jacobian matrix $J$ of the reconstructed state space $X_t$ can be utilized as an early warning signal for critical transition in high-dimensional systems. It only requires time series sampled from one dimension of the system to estimate this indicator, which is named dynamical eigenvalue (DEV)\cite{RN58}.
	
	We set $X_t=[x(t),x(t-\tau),\dots,x(t-(E-1)\tau)]^T$. Then at time $t_a$, we have $X_{t_a+\tau}=JX_{t_a}+v$ where
	\begin{equation}
		X_{t_a+\tau}=JX_{t_a}+v \Rightarrow
		\begin{bmatrix}
			x(t_a+\tau) \\
			x(t_a) \\
			x(t_a-\tau) \\
			\vdots \\
			x(t_a-(E-2)\tau)
		\end{bmatrix}
		=
		\begin{bmatrix}
			j_{11} & j_{12} & j_{13} & \cdots & j_{1E} \\
			1 & 0 & 0 & \cdots & 0 \\
			0 & 1 & 0 & \cdots & 0 \\
			\vdots & \vdots & \ddots & \ddots & \vdots \\
			0 & 0 & \cdots & 1 & 0	
		\end{bmatrix}
		\begin{bmatrix}
			x(t_a) \\
			x(t_a-\tau) \\
			x(t_a-2\tau) \\
			\vdots \\
			x(t_a-(E-1)\tau)
		\end{bmatrix}
		+
		\begin{bmatrix}
			v_1 \\
			0 \\
			0 \\
			\vdots \\
			0
		\end{bmatrix}. \notag
	\end{equation}
	Based on the S-map algorithm, matrix $J$ of the parameters $j_{11},j_{12},j_{13},\cdots,j_{1E}$ can be estimated as the associated S-map coefficients. Then we have
	\begin{equation}
		[j_{11},j_{12},j_{13},\cdots,j_{1E}]^T=A^{-1}B, \notag
	\end{equation}
	where $A$ is an $n\times E$ dimensional matrix ($n$ is the number of observations in $\left \{ x(t)\mid 1\leq t \leq n \right \}$), given by 
	\begin{equation}
		A_{ij}=\omega(||X_{t_i}-X_{t_a}||^2)x(t_i-(j-1)\tau),\quad (1\leq i \leq n,\,1\leq j \leq E), \notag
	\end{equation}
	and $B$ is an $n$-dimensional vector, given by
	\begin{equation}
		B_i=\omega(||X_{t_i}-X_{t_a}||^2)x(t_i+\tau),\quad (1\leq i \leq n). \notag
	\end{equation}
	The weighting function $\omega$ is defined by
	\begin{equation}
		\omega(u)=exp(-\frac{\theta u}{\overline{u}}), \notag
	\end{equation}
	where $||\cdot ||$ denotes the Euclidean distance and $\overline{u}$ is the average distance between $X_{t_a}$ and all other vectors on the attractor. The weight is tuned by the nonlinear parameter $\theta \geq 0$.
	
	Then we use the dominant eigenvalue $\lambda$ of the Jacobian matrix $J$ as the indicator for tipping points. When $|\lambda|$ reaches 1, we can predict that a bifurcation occurs.
	
	\subsection{Supplementary Note 6. Noise-induced premature bifurcation}
	In this section, we will introduce several bifurcation-related definitions and theorem. Based on this preliminary mathematical knowledge, we explain how slight noise induces premature bifurcation and the stochasticity of this phenomenon\cite{RN66}.
	\begin{definition}[bifurcation]
		The appearance of a topologically nonequivalent phase portrait under variation of parameters is called a bifurcation.
	\end{definition}
	\begin{definition}[$n_-,n_0,n_+$]
		Consider a continuous-time dynamical system defined by
		\begin{equation}
			\frac{dx}{dt}=f(x),\quad x\in R^n, \notag
		\end{equation}
		where $f$ is smooth. Let $x_0$ be an equilibrium of the system and let $A$ denote the Jacobian matrix $\frac{df}{dx}$ evaluated at $x_0$. Let $n_-$, $n_0$ and $n_+$ be the numbers of eigenvalues of $A$ (counting multiplicities) with negative, zero and positive real part, respectively. 
	\end{definition}
	\begin{definition}[hyperbolic equilibrium]
		An equilibrium is called hyperbolic if $n_0=0$.
	\end{definition}
	\begin{theorem}
		The phase portraits of system $\frac{dx}{dt}=f(x)$ near two hyperbolic equilibria, $x_0$ and $y_0$, are locally topologically equivalent if and only if these equilibria have the same number $n_-$, $n_0$ and $n_+$. 
	\end{theorem}
	
	Here we mathematically investigate how the noise induces the premature bifurcation. We consider a system
	\begin{equation}
		\frac{dx}{dt}=f(x,\mu), \tag{S15}
	\end{equation}
	 where $f$ is a smooth function. We add stochastic perturbation $\varepsilon g(x)$ ($g$ is also smooth) to the system (S15) and have the following system
	 \begin{equation}
	 	\frac{dx}{dt}=f(x,\mu)+\varepsilon g(x). \tag{S16}
	 \end{equation}
     As a system approaches a bifurcation, all the real parts of the eigenvalues of the Jacobian matrix at the equilibrium are less than zero. Thus, the equilibrium is hyperbolic. We assume $x_0$ is a hyperbolic equilibrium of system (S15) at $\mu=\mu_0$, and system (S16) has an equilibrium $x(\varepsilon)$ at $\mu=\mu_0$, such that $x(0)=x_0$. The equation defining equilibria of system (S16) at $\mu=\mu_0$ can be written as
	 \begin{equation}
	 	F(x,\varepsilon)=f(x,\mu_0)+\varepsilon g(x)=0, \notag
	 \end{equation}
	 with $F(x_0,0)=0$. We also have $F_x(x_0,0)=A_0$, where $A_0$ is the Jacobian matrix of system (S15) at the equilibrium $x_0$, and because $x_0$ is hyperbolic, $\left | A_0 \right | \neq 0$. Thus, the Implicit Function Theorem guarantees the existence of a smooth function $x=x(\varepsilon)$, $x(0)=x_0$, satisfying
	 \begin{equation}
	 	F(x(\varepsilon),\varepsilon)=0, \notag
	 \end{equation}
	 for $\varepsilon \in (-\alpha(x_0),\alpha(x_0))$ (for small values of $\left | \varepsilon \right |$). The Jacobian matrix of $x(\varepsilon)$ in system (S16),
	 \begin{equation}
	 	A_\varepsilon=\Big (\frac{df(x)}{dx}+\varepsilon \frac{dg(x)}{dx}\Big )\Big |_{x=x(\varepsilon)}, \notag
	 \end{equation}
	 which depends smoothly on $\varepsilon$ and coincides with $A_0$ in system (S15) at $\varepsilon=0$. Therefore, the $n_-$, $n_0$ and $n_+$ of $A_\varepsilon$ equal that of $A_0$ for all sufficiently small $\left | \varepsilon \right |$. Since $x_0$ is a hyperbolic equilibrium, it follows that $x_\varepsilon$ is also a hyperbolic equilibrium. According to \textbf{Theorem 2}, as $\mu_0$ is far from the bifurcation point $\mu_c$, the phase portraits near the equilibria of system (S16) are locally topologically equivalent under variation of $\mu$ near $\mu_0$. However, as $\mu_0$ is approaching $\mu_c$, the eigenvalue of $A_0$ with the largest real part increasingly approaches the imaginary axis (the condition for bifurcation occurs). This leads to the threshold of $\left | \varepsilon \right |$ for $A_\varepsilon$ and $A_0$ to have the same $n_-$, $n_0$ and $n_+$ is becoming increasingly smaller. Therefore, there may be a moment when the $\varepsilon$ crosses the threshold, the $n_-$, $n_0$ and $n_+$ of $A_\varepsilon$ change. According to \textbf{Theorem 2}, a topologically nonequivalent phase portrait appears near the equilibrium of system (S16), according to \textbf{Definition 1}, a bifurcation occurs in system (S16). But all the real parts of the eigenvalues of the Jacobian matrix at the equilibrium of system (S15) are still less than zero, i.e., system (S15) is before bifurcation. Thus the bifurcation may occurs earlier in the system (S16) with stochastic perturbation $\varepsilon g(x)$ than in the system (S15).
	 
	 \subsection{Supplementary Note 7. The second control experiment studied in the main manuscript}
	 We note that the relation between the recovery rate $\lambda$ and the bifurcation parameter $\mu$ is the same in the normal forms of supercritical and subcritical pitchfork bifurcations, which is $\lambda=\mu$. Moreover, their normal forms only differ in the cubic term. Therefore, if the DL model trained on time series with subcritical (supercritical) pitchfork bifurcation can be used to predict tipping points of time series with supercritical (subcritical) pitchfork bifurcation, it indicates that the DL algorithm utilizes the features of recovery rate in the normal form to predict tipping points rather than other features in the data, such as those generated by higher-order terms.
	 
	 Based on the ideas above, we can design a control experiment. We train two DL models on two datasets, each consisting solely of time series with supercritical and subcritical pitchfork bifurcation, respectively. Then we apply these two DL models on irregularly-sampled model time series sampled from a dynamical system with supercritical pitchfork bifurcation. We compare whether there are differences in the performance of these two DL models in predicting tipping points on these time series.
	 \subsubsection{Generation of training data with pitchfork bifurcation}
	 Each training set with supercritical or subcritical pitchfork bifurcation consists of simulation data from a library of 50,000 models. The models are composed of the normal form of the supercritical or subcritical pitchfork bifurcation and higher order polynomial terms up to degree 10 with coefficients drawn from a normal distribution\cite{RN108}.
	 
	 The model for the supercritical pitchfork bifurcation is
	 \begin{equation}
	 	\frac{dx}{dt}=\mu x-x^3+\sum_{i=4}^{10}\alpha_ix^i, \notag
	 \end{equation}
	 and the model for the subcritical pitchfork bifurcation is
	 \begin{equation}
	 	\frac{dx}{dt}=\mu x+x^3+\sum_{i=4}^{10}\alpha_ix^i, \notag
	 \end{equation}
	 where $\alpha_i \sim N(0,1)$.
	 \subsubsection{Tested model with supercritical pitchfork bifurcation}
	 We use an ecological supercritical pitchfork bifurcation model\cite{RN28} for testing, which is given by
	 \begin{equation}
	 	\frac{dx}{dt}=rx(1 - \frac{x}{k})(x - x_c) - cx + I + \sigma \xi(t), \notag
	 \end{equation}
	 where $x$ represents biomass of some population, $k$ is its carrying capacity, $r$ is the maximum growth rate, $c$ is the maximum grazing rate, $x_c$ is the Allee threshold, $I$ is the immigration rate, and $\xi(t)$ is a Gaussian white noise process. We use parameter values $k=10$, $c=0.8$, $x_c=5$, $I=4$ and $r$ increases at rates of $1\times 10^{-5}$. We generate model time series by this equation from eleven initial values of $r$, which are $0,0.02,0.04,\dots,0.18,0.2$. In this configuration, the supercritical pitchfork bifurcation occurs at $r=0.32$.
	 
	 The performance of tipping points prediction between these two DL models on irregularly-sampled model time series with supercritical pitchfork bifurcation is presented in Fig. S9.
	 
    \clearpage
    \section{Supplementary Tables}
    \vspace{0.5cm}
	  \begin{table}[htbp]
	 	\centering
	 	\caption*{\textbf{Supplementary Table 1} \hspace{1mm} The hyperparameters of the DL model and the LSTM (ablation study).}
	 	\begin{tabular}{c c c c c c c}
	 		Model Name & Learning Rate & CNN Filters & CNN Kernel Size & Max Pooling Size & LSTM1 Cells & LSTM2 Cells \\
	 		\hline
	 		DL model & 0.01 & 60 & (10,2) & (4,1) & 40 & 60 \\
	 		LSTM & 0.01 & - & - & - & 40 & 60 \\
	 		\hline
	 	\end{tabular}
	 \end{table}
    \vspace{1cm}
    \begin{table}[htbp]
	\centering
	\caption*{\textbf{Supplementary Table 2} \hspace{1mm} The hyperparameters of the Fold DL model, the Hopf DL model and the Transcritical DL model.}
	\begin{tabular}{c c c c c c c}
		Model Name & Learning Rate & CNN Filters & CNN Kernel Size & Max Pooling Size & LSTM1 Cells & LSTM2 Cells \\
		\hline
		Fold DL model & 0.01 & 30 & (12,2) & (2,1) & 30 & 30 \\
		Hopf DL model & 0.01 & 30 & (14,2) & (4,1) & 60 & 50 \\
		Transcritical DL model & 0.01 & 30 & (8,2) & (3,1) & 40 & 60 \\
		\hline
	\end{tabular}
    \end{table}
    \vspace{1cm}
    \begin{table}[htbp]
	\centering
	\caption*{\textbf{Supplementary Table 3} \hspace{1mm} The hyperparameters of the Supercritical DL model and the Subcritical DL model.}
	\begin{tabular}{c c c c c c c}
		Model Name & Learning Rate & CNN Filters & CNN Kernel Size & Max Pooling Size & LSTM1 Cells & LSTM2 Cells \\
		\hline
		Supercritical DL model & 0.01 & 30 & (14,2) & (4,1) & 60 & 50 \\
		Subcritical DL model & 0.01 & 30 & (10,2) & (3,1) & 40 & 60 \\
		\hline
	\end{tabular}
   \end{table}
	
	 \vspace{2.5cm}
	 
	 \bibliographystyle{unsrtnat}
	 \bibliography{reference.bib}